\documentclass[10pt, a4paper, logo, twocolumn]{googledeepmind}

\usepackage[sort&compress, numbers]{natbib}
\usepackage[capitalise]{cleveref}
\usepackage{fancybox}
\usepackage{makecell}
\usepackage{appendix}
\usepackage[font=small,labelfont=bf]{caption}

\usepackage{amsfonts}
\usepackage{titlesec}
\usepackage{graphicx}
\usepackage{listings}
\usepackage{xcolor}
\usepackage{bbm}

\usepackage{paralist}
\usepackage{authblk}
\usepackage{pifont}
\usepackage{ifthen}
\usepackage{float}
\usepackage{subcaption}
\usepackage{booktabs}

\usepackage{amsmath}
\usepackage{amssymb}
\usepackage{amsthm}
\usepackage{comment}

\usepackage[normalem]{ulem}
\newtheorem{prop}{Proposition}
\usepackage{amsthm}
\usepackage{amsmath}
\makeatletter
\newtheorem*{rep@theorem}{\rep@title}
\newcommand{\newreptheorem}[2]{%
\newenvironment{rep#1}[1]{%
 \def\rep@title{#2 \ref{##1}}%
 \begin{rep@theorem}}%
 {\end{rep@theorem}}}
\makeatother

\newcommand{\todo}[1]{}
\renewcommand{\todo}[1]{{\color{blue} TODO: {#1}}}

\newreptheorem{theorem}{Theorem}

\newreptheorem{assumption}{Assumption}

\newreptheorem{lemma}{Lemma}

\newreptheorem{corollary}{Corollary}

\usepackage{amsmath,amsfonts,bm}

\def\eqref#1{equation~\ref{#1}}
\def\1{\bm{1}}

\DeclareMathAlphabet{\mathsfit}{\encodingdefault}{\sfdefault}{m}{sl}
\SetMathAlphabet{\mathsfit}{bold}{\encodingdefault}{\sfdefault}{bx}{n}

\newcommand{\softmax}{\mathrm{softmax}}

\DeclareMathOperator*{\argmax}{arg\,max}
\DeclareMathOperator*{\argmin}{arg\,min}

\PassOptionsToPackage{inline}{enumitem}
\newlist{inlinelist}{enumerate*}{1}
\setlist[inlinelist]{label=(\textit{\roman*})}

\begin{document}

\title{Uncovering mesa-optimization algorithms in Transformers}

\author[a,b,*]{Johannes von Oswald}
\author[a,b,*]{Maximilian Schlegel}
\author[a,b]{Alexander Meulemans}
\author[a,b]{Seijin Kobayashi}
\author[a]{Eyvind Niklasson}
\author[b]{Nicolas Zucchet}
\author[a]{Nino Scherrer}
\author[d]{Nolan Miller}
\author[d]{Mark Sandler}
\author[a]{Blaise Agüera y Arcas}
\author[d]{Max Vladymyrov}
\author[e]{Razvan Pascanu}
\author[a,b,*]{João Sacramento}

\affil[a]{Google, Paradigms of Intelligence Team}
\affil[b]{ETH Z\"{u}rich}
\affil[d]{Google Research}
\affil[e]{Google DeepMind}
\affil[*]{Contributed equally to this work.}

\correspondingauthor{ jvoswald@google.com, joaosacramento@google.com}

\begin{abstract}
Some autoregressive models exhibit in-context learning capabilities: being able to learn as an input sequence is processed, without undergoing any parameter changes, and without being explicitly trained to do so. The origins of this phenomenon are still poorly understood. Here we analyze a series of Transformer models trained to perform synthetic sequence prediction tasks, and discover that standard next-token prediction error minimization gives rise to a subsidiary learning algorithm that adjusts the model as new inputs are revealed. We show that this process corresponds to gradient-based optimization of a principled objective function, which leads to strong generalization performance on unseen sequences. Our findings explain in-context learning as a product of autoregressive loss minimization and inform the design of new optimization-based Transformer layers.
\end{abstract}

\maketitle
\noindent We are currently witnessing a paradigm shift in machine learning. Specialized models trained on large labeled data sets are being replaced by generalist foundation models trained with self-supervision~\citep{bommasani_opportunities_2022}. There is increasing evidence that these models can adapt to a wide range of tasks after a brief period of supervised learning (`fine-tuning'). Intriguingly, some foundation models are capable of learning directly from contextual input data, without having been explicitly designed or trained to do so. In this way, parameter fine-tuning can often be sidestepped altogether, making on-the-fly adaptation to new tasks possible simply by providing examples in context. This powerful yet puzzling phenomenon, known as in-context learning~\citep{brown_language_2020}, was first observed in autoregressive large language models (LLMs).

A number of recent theoretical studies have begun to shed light on how in-context learning works, and why it arises. A seminal analysis of Transformers, the backbone architecture~\citep{vaswani_attention_2017} of the majority of LLMs, identified a two-layer circuit mechanism called `induction head' responsible for in-context learning in shallow networks, and provided evidence for its likely involvement in deeper and more complex networks~\citep{elhage_mathematical_2021,olsson_-context_2022}. A complementary line of work has shown that in-context learning can emerge in small-scale models, as long as the data distribution displays certain properties~\citep{chan_data_2022,xie_explanation_2022}, and that it can vanish under long training times~\citep{singh_transient_2023}. Building on prior recurrent neural network studies~\citep{hochreiter_learning_2001,duan_rl2_2016,wang_learning_2016,rabinowitz_meta-learners_2019}, yet another line of investigation has studied the metalearning abilities of Transformers, explicitly training the models to solve supervised learning problems in-context~\citep{garg_what_2022,akyurek_what_2023,dai_why_2022,kirsch_general-purpose_2022}. In such a setup, in-context learning is no longer an emergent phenomenon, but is `forced' by the training regime, simplifying the analysis. Our previous work showing that Transformers solve linear regression tasks by gradient descent~\citep{von_oswald_transformers_2023}, later followed by a series of refined studies and mathematical analyses ~\citep{zhang_trained_2023,mahankali_one_2023,ahn_transformers_2023,li_transformers_2023,raventos_pretraining_2023,ding_causallm_2023,vladymyrov_linear_2024,fu_transformers_2023,giannou_how_2024}, falls under the same category as it also relies on explicit metalearning.

In this paper, we continue to analyze the in-context learning abilities of Transformers, but shift our focus to autoregressive sequence prediction tasks. Like LLMs---and most sequence models---we train Transformers in a self-supervised manner by minimizing a next-token prediction error objective. Based on our previous results on metalearned Transformers~\citep{von_oswald_transformers_2023}, we then investigate whether the prediction algorithm learned by autoregressive Transformers can be interpreted as gradient-based learning on a suitable contextual objective function. We find that this holds true for a range of synthetic sequence modeling tasks. In such a controlled synthetic data setting, we identify a gradient-based learning mechanism spanning multiple Transformer layers. We refer to this mechanism as a `mesa-optimizer' to emphasize that it is acquired through training, as opposed to being inherent to the model \citep[see][]{hubinger_risks_2019}. The mesa-optimizer adapts the model as new contextual information becomes available, enabling it to improve its predictions with near-optimal sample efficiency. Moreover, the same mechanism enables learning downstream tasks from contextual demonstrations only. Taken together, our results explain, at least in the settings we have considered, the emergence of in-context learning in Transformers trained only to predict the next token.

\section*{Results}

We study autoregressive sequence modeling tasks where the goal is to causally predict, at every time step $t=1,\ldots,T-1$, the next element $e_{t+1}$ in a sequence of tokens $e=(e_t)_{t=1}^T$, given the past $(e_{t^\prime})_{t^\prime=1}^t$ as context.

We examine a range of causally masked Transformer models \citep{vaswani_attention_2017} trained to solve such problems, from simple attention-only models to full-fledged deep Transformers comprising multiple attention layers, layer normalization \citep[LayerNorm;][]{ba_layer_2016}, and nonlinear multi-layer perceptron (MLP) blocks, cf.~\emph{Materials and Methods}. The objective of training is to find a set of parameters $\theta$ that minimize the cumulative next-token prediction error
\begin{equation}
\label{eq:autoregressive-loss}
\mathcal{L}(\theta) = \mathbb{E}_{e\sim p(e)}\!\left[\frac{1}{2}\sum_{t=1}^{T-1}\|e_{t+1} - f_{t}(e_{1:t}, \theta) \|^2\right],
\end{equation}
where $f_{t}(e_{1:t}, \theta)$ denotes the Transformer output conditioned on the context $e_{1:t}$, and the expectation is taken over the sequence distribution $p(e)$, which we describe next. We focus on continuous-state problems, with $e_t \in \mathbb{R}^{n_e}$, and take the squared error as the per-time-step loss, the standard objective for autoregressive problems with continuous outputs.

Our Transformers are trained on synthetic sequences $(s_t)_{t=1}^T$ of observations $s_t \in \mathbb{R}^{n_s}$ generated by discrete-time dynamical systems. As we detail in \emph{Materials and Methods}, we consider a range of sequence generators described in state-space representation: \begin{inlinelist} \item linear systems with full observability  ($s_t=h_t$) of the internal system state $h_t \in \mathbb{R}^{n_h}$; \item partially-observed linear systems, where we only allow access to a low-dimensional state projection, $s_t = C^* h_t$; \item nonlinear dynamics, with state transitions governed by nonlinear neural networks \end{inlinelist}. Typically, each token corresponds to one observation, $e_t = s_t$, but we also study tokenization schemes that aggregate several observations within one token. These aggregate token representations play an important role in the theory we develop below.

\subsection*{Next-token prediction by mesa-optimization}
In this paper, we hypothesize that training Transformers on next-token prediction tasks as described above installs a gradient-based, in-context optimization algorithm in the forward pass of the model. Following the terminology of Hubinger et al.~\cite{hubinger_risks_2019}, we refer to this hypothetical acquired process as mesa-optimization, to distinguish it from the base-optimization of Eq.~\ref{eq:autoregressive-loss}, over which we have explicit control.

More concretely, we hypothesize that generating the future-token prediction $f_{t}(e_{1:t}, \theta)$ involves using the current and past tokens $e_{1:t}$ to build a sequence-specific latent model on the fly. We focus on the case where this model is linear in its parameters, which we denote by $\Phi$.

According to our mesa-optimization hypothesis, trained Transformers successively \emph{learn} a sequence of such parameters $\Phi_t$ as input tokens are gradually revealed, by minimizing an in-context objective function $L_t(e_{1:t},\Phi)$ using gradient information $\nabla_\Phi L_t(e_{1:t},\Phi)$. The resulting in-context models are then used to generate the Transformer predictions. It is important to appreciate that these in-context latent models and their learning rules are not explicitly hardwired in the Transformer design, but are instead a by-product of base-optimization. The parameters $\Phi$ may thus be thought of as an implicit type of fast (i.e., sequence-specific) weights \cite{schmidhuber_learning_1992,ba_using_2016} which live in the short-term memory of a Transformer model, not in its learned parameters.

Before verifying whether our hypothesis holds for trained models, we first show that in theory, autoregressive linear Transformers are capable of optimizing quadratic loss functions in-context. We show this constructively, by providing a set of parameters $\theta$ such that a linear Transformer implements a mesa-optimizer. This construction will then guide our analyses of trained models.

\subsection*{Theory of self-attention mesa-optimizers}
Our first theoretical result concerns a single layer of causally-masked self-attention, the architectural component at the heart of an autoregressive Transformer; we will later consider deeper, more complex architectures. Given an input sequence $(e_t)_{t=1}^T$, one such layer with $H$ heads updates each token $e_t \gets e_t + \Delta e_t^{\text{sa}}$ following the rule
\begin{equation}
\label{eq:attention-dynamics}
\Delta e_{t}^{\text{sa}} = \sum_{h=1}^{H} P_h V_{h,t} \,\alpha(K_{h,t}^\top q_{h,t}),
\end{equation}

where $q_{h,t} = W_{h,q} e_t \in \mathbb{R}^{n_a}$ is referred to as a query, each column $k_{h,t^\prime} = W_{h,k} e_{t^\prime} \in \mathbb{R}^{n_a}$ of matrix $K_{h,t} \in \mathbb{R}^{n_a \times t}$ as a key, and each column $v_{h,t^\prime} = W_{h,v} e_{t^\prime} \in \mathbb{R}^{n_v}$ of matrix $V_{h,t} \in \mathbb{R}^{n_v \times t}$ as a value. The parameters of this layer are the projection matrices $\{(P_h, W_{h,q}, W_{h,k}, W_{h,v})\}_{h=1}^H$ for all heads; we absorb bias terms, and assume here for conciseness that all heads are equally sized. The function $\alpha$ applied to vector $a \in \mathbb{R}^t$ returns an attention weight vector. For the theoretical results presented below, we focus on the case where $\alpha$ is the identity function, which yields the linear self-attention layer, the main building block of linear Transformers~\citep[e.g.,][]{katharopoulos_transformers_2020,wang_linformer_2020,schlag_linear_2021,choromanski_rethinking_2021,ahn_linear_2024}. In our experimental analyses, we also study standard (softmax) self-attention layers, where $\alpha(a)_i = \softmax(a)_i := (\sum_{t^\prime=1}^t \exp(a_{t^\prime}))^{-1} \exp(a_i) $, present in the original and still most popular Transformer architecture~\citep{vaswani_attention_2017}.

\begin{figure}[t!]
    \centering
     \includegraphics[width=0.452\textwidth]{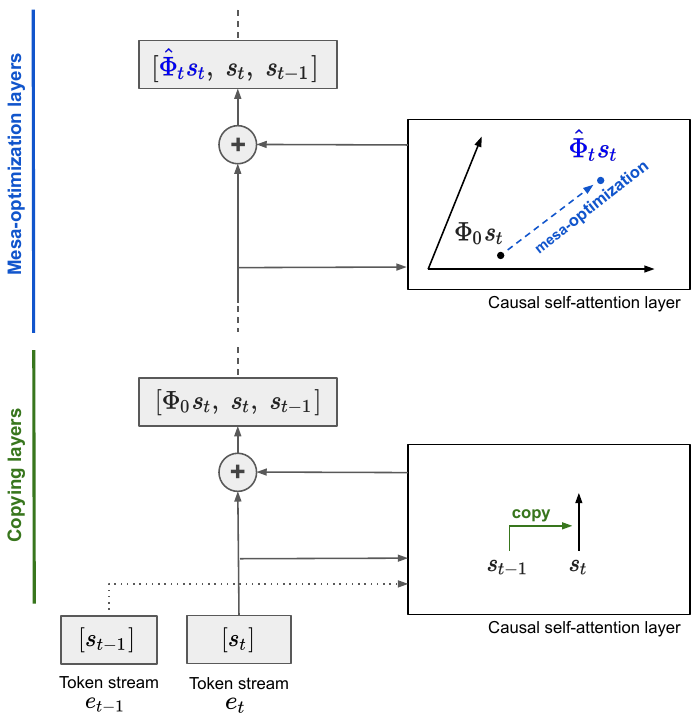}
    \vspace{-0.1cm}
    \caption{Illustration of mesa-optimization in autoregressive Transformers. The neural dynamics implements an optimization-based in-context learning algorithm, which optimizes the parameters $\Phi$ of a linear model over a series of causally-masked attention layers. Taking as inputs an initial set of parameters $\Phi_0$ and a training set of input-target pairs $\{(s_{t^\prime}, s_{t^\prime+1})\}_{t^\prime=1}^{t-1}$ constructed from context, this process returns a prediction $\hat{\Phi}_t s_t$ obtained by applying the learned model to the current input. Early layers implement a copy operation which binds multiple consecutive tokens together, in agreement with previous in-context learning analyses \citep{elhage_mathematical_2021,olsson_-context_2022}. This aggregate-token representation enables the implementation of gradient-based optimizers in subsequent attention layers, cf.~Propositions~\ref{prop:mesa-gradient-descent-construction} and \ref{prop:multi-layer-precond-gd}.}
    \label{fig:illus-prop-1}
     \vspace{-0.2cm}
\end{figure}

Consider the cumulative squared-error loss function
\begin{equation}
L_t(\Phi) = \sum_{t^\prime=1}^{t-1} \frac{1}{2} \|s_{t^\prime+1} - \Phi s_{t^\prime} \|^2,
\end{equation}
where $\Phi \in \mathbb{R}^{n_s \times n_s}$ parametrizes a first-order linear autoregressive model which predicts $s_{t+1}$ from $s_t$. We show that one linear self-attention layer can implicitly represent such a model in its activations, with mesa-parameters $\Phi$ learned by a step of gradient descent on the mesa-objective $L_t(\Phi)$.

\begin{prop}[1-step attention-based gradient descent]
\label{prop:mesa-gradient-descent-construction}
Given tokens of the form $e_{t} = [\Phi_0 s_t , s_t, s_{t-1}]$, for $t = 2,...,T$, if the projection matrices $W_k$, $W_q$, $W_v$, $P$ are such that
\begin{equation}
PW_v = \begin{bmatrix} 
 0 & \eta I_s
 & -\eta\Phi_0  \\
  0 & 0 & 0  \\
 0 & 0 & 0  \\
\end{bmatrix},\; W_k^\top W_q = \begin{bmatrix}
 0 & 0 & 0  \\
  0 & 0 & 0  \\
 0 & I_s & 0  \\
\end{bmatrix},\label{eq:3-channel-mesa-gradient-construction}\nonumber
\end{equation}
with $I_s$ the identity matrix of size $n_s \times n_s$, then the transformation of every token $e_t$ by one causally-masked linear self-attention head is identical to the gradient-induced update $e_t \leftarrow \begin{bmatrix} (\Phi_0 - \eta \nabla L_t(\Phi_0))s_t,s_t,s_{t-1}\end{bmatrix} $.
\end{prop}
Proposition~\ref{prop:mesa-gradient-descent-construction} (proven in \emph{Materials and Methods}) is an immediate extension of the main result of von Oswald et al.~\citep{von_oswald_transformers_2023} to the autoregressive sequence modeling setting, where $T$ loss functions $(L_t)_t$ must be optimized in sequence, see \emph{Materials and Methods} for details. Since $L_t$ is the cumulative squared error up to time $t$, Proposition~\ref{prop:mesa-gradient-descent-construction} implements a `full-batch' gradient step. Notably, the self-attention layer executes this step in all $T$ problems in parallel. We remark that our construction assumes a special three-channel tokenization, where a single token encodes the current input $s_t$, the previous input $s_{t-1}$, and an initial prediction $\Phi_0 s_t$. As illustrated in Fig.~\ref{fig:illus-prop-1}, we will later show that trained Transformers learn to internally produce such encodings when driven by a standard-format ($e_t=s_t$) sequence, but for now we proceed under the assumption that the tokens are structured in such a way.

We now turn to multi-layer, self-attention-only models. Here, we find that causally-masked autoregressive modeling complicates the problem, in the sense that stacking $k$ layers following~Proposition~\ref{prop:mesa-gradient-descent-construction} yields an unconventional biased algorithm that is expected to be slower than $k$-step gradient descent, as analyzed in \citep{ding_causallm_2023}. There exists, however, an alternative unbiased mesa-optimizer for multi-layer models, which introduces an additional layerwise operation for improving the preconditioning of mesa-optimization. This algorithm again makes use of self-attention layers, now employed to transform the input data. In the limit of many such layers, a single gradient descent step then yields the optimal (least-squares) mesa-optimization solution.

\begin{prop}[Multi-attention-layer mesa-optimizer]
\label{prop:multi-layer-precond-gd}
Assume we are given for every time step $t=2,\ldots,T$ a sequence of suitably-constructed input tokens $(e_{t^\prime})_{t^\prime=1}^t$, and a regularized mesa-objective we wish to minimize  $\bar{L}_t(\Phi) = \sum_{t^\prime=1}^{t-1} \frac{1}{2} \|s_{t^\prime+1} - \Phi s_{t^\prime} \|^2 + \frac{1}{2\lambda}||\Phi||_\mathrm{F}^2$  where $\lambda^{-1} \in \mathbb{R}$ is a regularization hyperparameter and $S_t$ is the data matrix whose columns are $(s_{t^\prime})_{t^\prime=1}^t$. Then, there exists a set of linear Transformer parameters $\theta$ that yield an approximation to the vectors $H_t^* s_t := (S_{t-1} S_{t-1}^\top + 1/\lambda I)^{-1}s_t$ in parallel for all $t$ in their forward pass, with approximation error decreasing with the number of linear self-attention layers $k$. As a consequence, in the many-layer limit the Transformer can minimize the regularized mesa-objective.
\end{prop}
A concrete parameter construction and proof are provided in the \emph{Materials and Methods}. 

Propositions \ref{prop:mesa-gradient-descent-construction} and \ref{prop:multi-layer-precond-gd} show that simplified Transformers can, at least in theory, minimize cumulative squared-error objectives in-context, without any actual parameter (`in-weights') learning taking place. As we shall see in our experimental section below, these ideal constructions yield solutions to our synthetic tasks, and they generate testable hypotheses that inform our experiments with trained models. Before proceeding to our empirical analyses, we present one last theoretical result motivated by the constructions above: a novel self-attention layer designed for efficient least-squares in-context learning.

\subsection*{An attention layer for optimal least-squares learning}
\label{sect:mesa-layer}
The mesa-optimizers discussed so far require in general many layers to reach a desired error. This observation leads us to develop the mesa-layer, a self-attention layer derived from in-context optimization first principles. More concretely, we show that an appropriately modified attention layer yields autoregressive least-squares solutions in sequence and in a single step, a computation that would otherwise require infinitely many linear self-attention layers under Proposition~\ref{prop:multi-layer-precond-gd}. Thus, if the mesa-optimization hypothesis advanced in this paper describes actual trained standard Transformers, it should be possible to improve their performance by introducing such a layer in their architecture. The mesa-layer therefore provides one additional way of verifying the mesa-optimization hypothesis in experiments.

The mesa-layer changes a sequence of input tokens according to the update
\begin{align}
\Delta e_{t}^\text{mesa} &= \sum_{h=1}^H P_h \hat{\Phi}_{h,t}^\text{mesa} q_{h, t},
\end{align}
with
\begin{equation}
\begin{aligned}
\label{eq:mesa-layer-objective}
     \hat{\Phi}_{h,t}^\text{mesa} = \argmin_\Phi \, \frac{1}{2}\sum_{t^\prime=1}^t || v_{h,t^\prime} - \Phi k_{h,t^\prime}||^2 +\frac{||\Phi||_\mathrm{F}^2}{2\lambda_h}.
\end{aligned}
\end{equation}
Above, the (learnable) scalar $\lambda_h^{-1} >0 $ controls the strength of a regularizer added to improve generalization, and key, value and query vectors are the usual learned head-specific affine transformations of the tokens, as in Eq.~\ref{eq:attention-dynamics}. However, through Eq.~\ref{eq:mesa-layer-objective} these vectors are now assigned a precise, interpretable role: value vectors specify targets to which an internal model with parameters $\Phi$ should map training and test inputs, represented by keys and queries, respectively. We note that the minimizer of a regularized squared-error objective can be mapped to Eq.~\ref{eq:mesa-layer-objective} under an appropriate tokenization (such as the one of Proposition~\ref{prop:mesa-gradient-descent-construction}) by appropriately setting the projection matrices $W_{h,v}$, $W_{h,k}$ and $W_{h,q}$.

At any given time step $t=1,\ldots,T$ computing $\Delta e_{t}^\text{mesa}$ requires solving a regularized least-squares problem per attention head. To efficiently solve this sequence of $T$ optimization problems, we leverage the recursive dependency of the $T$ solutions, which can be expressed in closed-form as 
\begin{align}
    \hat{\Phi}_{h,t}^\text{mesa} &= V_{h,t}K_{h,t}^\top R_{h,t}\nonumber\\
    &= \sum_{t^\prime=1}^t v_{h,t^\prime}k_{h,t^\prime}^\top\!\left(\sum_{t^\prime=1}^t k_{h,t^\prime}k_{h,t^\prime}^\top + 1/\lambda_h \, I\right)^{\!\!-1}.
    \label{eq:reg-optimal-W-hat}
\end{align}
As $\lambda_h \to 0$, we recover a standard linear self-attention layer. Thus, the mesa-layer strictly generalizes the latter.

We now use the Sherman-Morrison formula \citep{sherman_adjustment_1950} to obtain the inverse at time $t$ from the inverse at the previous time step $t-1$. This iterative update is possible because we only change the inverse by a rank-one update. The following solution scheme is known as recursive least squares \citep{gauss_theoria_1821}:
\begin{equation}
  R_{h,t} = R_{h,t-1} - \frac{R_{h,t-1}k_{h,t}k_{h,t}^\top R_{h,t-1}}{1 + k_{h,t}^\top R_{h,t-1}k_{h,t}}
\end{equation}
with $R_{h,0} = \lambda_h \, I$. We can then (causally in time) compute 
\begin{equation}
\Delta e_{t}^\text{mesa} = \sum_{h=1}^H P_h V_{h,t}K_{h,t}^\top R_{h,t} q_{h, t},
\end{equation}
which requires 2 additional vector-matrix and 2 vector-vector multiplications per step compared to standard self-attention.

Naive backward gradient computation requires storing matrices of dimension $n_a \times n_a$ in memory across time. However, this memory overhead can be avoided using the Sherman-Morrison formula in reverse during backpropagation, as we show in the \emph{SI Appendix}, enabling memory-efficient gradient computation of the output of the mesa-layer w.r.t.~its inputs. We note that while the implementation described here has a desirable $\mathcal{O}(1)$ inference memory cost, it is not parallelizable across time. This is a disadvantage for training on contemporary hardware shared with nonlinear recurrent neural networks, but not with standard self-attention layers.

The mesa-layer is closely related to the Delta-Net model of Schlag et al.~\citep{schlag_linear_2021}, which is hardwired to do one gradient descent step per time point. It can also be seen as an adaptation of the intention layer proposed by Garnelo \& Czarnecki \citep{garnelo_exploring_2023} to the sequential, autoregressive case. The latter corresponds exactly to a non-causally-masked version of Eq.~\ref{eq:reg-optimal-W-hat}. Here, we focus on the autoregressive setting, which leads us to develop recursive forward and backward updates, in order to achieve efficient sequential inference and training.

\subsection*{Aggregate internal token representations develop through training}

\begin{figure}[htbp!]
    \begin{center}      \includegraphics[width=0.47\textwidth]{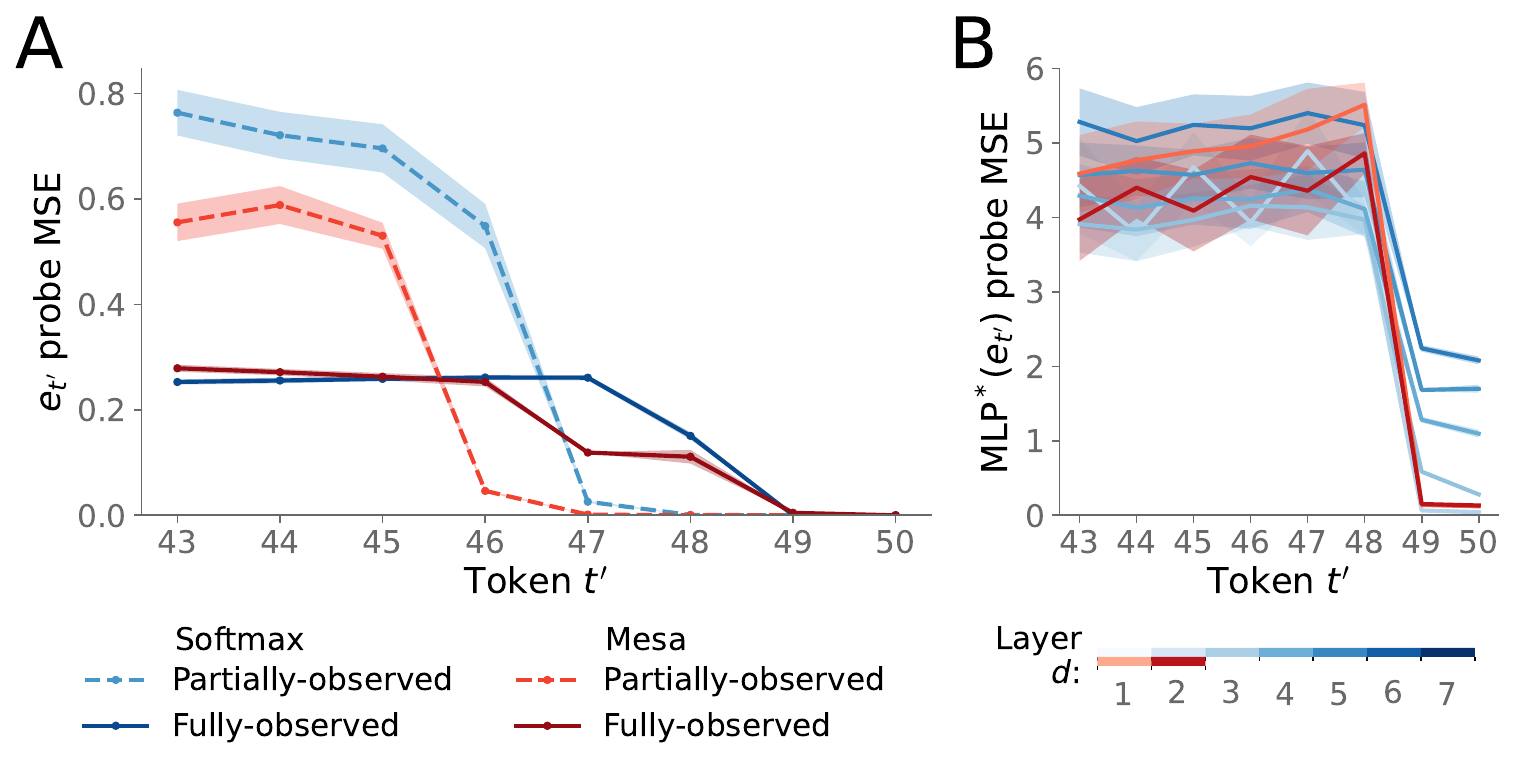}
    \caption{Early layers of trained autoregressive Transformers (blue lines) produce internal token representations that support mesa-optimization by subsequent layers. Similar results are obtained for standard deep Transformers and new compact, two-layer model variants which feature the mesa-layer (red lines). (\emph{A}) After training, the past token $e_{t^\prime}$ ($t'=49$) can be almost perfectly linearly decoded from the current ($t=50$) output of the first Transformer layer. The decoding horizon $t-t^\prime$ increases when the Transformer is trained to solve partially-observed tasks (dashed lines; notice low probing error for $t' \in \{49, 48, 47, 46\}$). (\emph{B}) Same analysis, now for the groundtruth hidden state $\text{MLP}^*(e_t^\prime)$ of a nonlinear sequence generator and for varying layer depth. Current ($t^\prime=t$) and preceding ($t^\prime = t-1$) states can be linearly decoded from  early Transformer layers (depicted with lighter color tones) after training on nonlinear tasks.}
    \label{fig:induction-head-copying}
    \end{center}
\end{figure}

We begin our empirical analysis of Transformer models trained by autoregressive loss (Eq.~\ref{eq:autoregressive-loss}) minimization by searching for evidence of an internal token binding mechanism. Recall that Propositions \ref{prop:mesa-gradient-descent-construction} and  \ref{prop:multi-layer-precond-gd} required a non-standard token format, in which consecutive observations were aggregated within a single token $e_t$. In our first set of experiments, we adopt a standard token format and provide only the current observation $s_t$ as the input $e_t$ to the model. The first prediction of our theory is that training should install a token binding mechanism, responsible for aggregating multi-time-step observation information within a single token. We now show that this indeed occurs in actual trained models.

In Fig.~\ref{fig:induction-head-copying}, we report the performance of linear decoders \citep[probes;][]{alain_understanding_2017} trained to predict previous tokens from the output of the first attention layer of a deep Transformer model. We consider both standard Transformer models featuring seven softmax self-attention layers, MLPs and LayerNorm, as well as a novel compact Transformer which combines one layer of softmax self-attention and one mesa-layer, described in detail in \emph{Materials and Methods}. We relegate architectures solely built out of mesa-layer models to the \emph{SI Appendix}, as we found that these were generally outperformed by hybrid softmax-mesa architectures. We see that after training it becomes possible to decode past tokens from the present token (see also Fig.~\ref{fig:illus-prop-1}A), with decoding horizon increasing for partially-observed problems, for both standard softmax Transformers and the novel hybrid softmax-mesa Transformers introduced in this paper. For the fully observed setting, the probe error increases quickly when predicting more than one step in the past, aligned with our token construction that binds together only consecutive tokens. Moreover, when the models are trained on systems with nonlinear dynamics, the performance of linear probes that decode the hidden state of the sequence generator system from the outputs of MLP layers improves, in particular for early MLP layers.

These results can be explained by analyzing the tasks the Transformers are trained on. When the input data is generated by a fully-observed linear dynamical system, the maximum likelihood estimator of the groundtruth parameters $W^*$ corresponds to the least-squares solution $\argmin_\Phi \sum_{t^\prime=1}^{t-1} \frac{1}{2} \|s_{t^\prime+1} - \Phi s_{t^\prime} \|^2$. The mesa-optimizers described in Propositions \ref{prop:mesa-gradient-descent-construction} and \ref{prop:multi-layer-precond-gd}, as well as the mesa-layer, can be readily applied to solve this problem (or a regularized variant, corresponding to maximum a posteriori estimation under a Gaussian prior on $\Phi$), as long as inputs and targets are both encoded within a single token $e_t$. This is what we observe in Fig.~\ref{fig:induction-head-copying}A.

The nonlinear case can be approached similarly, by performing least-squares estimation of $W^*$ after an appropriate nonlinear feature transformation. For a Transformer model, the MLP layers are perfectly placed to implement this transformation. In line with this, we find that early layers develop a set of basis functions that align with those of the nonlinear sequence generator, Fig.~\ref{fig:induction-head-copying}B, followed by a token binding step (cf.~\emph{SI Appendix}).

We find that there is a longer dependence on the past under partial observability (Fig.~\ref{fig:induction-head-copying}A), where next-token prediction is complicated due to the presence of latent variables. This behavior can again be explained in the light of our mesa-optimization hypothesis. First, we note that the task our Transformers face is harder than classical Kalman filtering \citep{kalman_new_1960}, where knowledge of groundtruth system parameters is assumed. Methods such as subspace identification \citep{tangirala_principles_2018} or variational expectation maximization \citep{marino_general_2018} are applicable to this harder setting, but we found these standard methods difficult to map to a Transformer. We identified however a less-orthodox algorithm, mathematically related to data-driven control techniques \citep{willems_note_2005,de_persis_formulas_2020}, which runs online as a sequence is unveiled, and that is based on least-squares estimation. This algorithm can therefore be implemented by a Transformer through Propositions \ref{prop:mesa-gradient-descent-construction} and \ref{prop:multi-layer-precond-gd}, or by a mesa-layer. The key step is to encode $k$ past observations $s_{t-k+1}, \ldots, s_{t}$ in a single augmented variable $z_t \in \mathbb{R}^{k n_s}$ of large enough dimensionality; it can then be shown that maximum likelihood estimation of the next token $s_{t+1}$ can be achieved by solving a least-squares problem involving the augmented variables $\{z_t\}$. We provide a full derivation and analysis in \emph{Materials and Methods}, where we show that the optimal value of $k$ depends on the compression ratio $n_h/n_s$. According to this theory, we would expect to see a higher-order ($k>1$) dependency on past inputs for the case of partially-observed dynamics. This corresponds to what we find in trained Transformers, cf.~Fig.~\ref{fig:induction-head-copying}.

\begin{figure}[htbp!]
    \centering
    \includegraphics[width=0.44\textwidth]{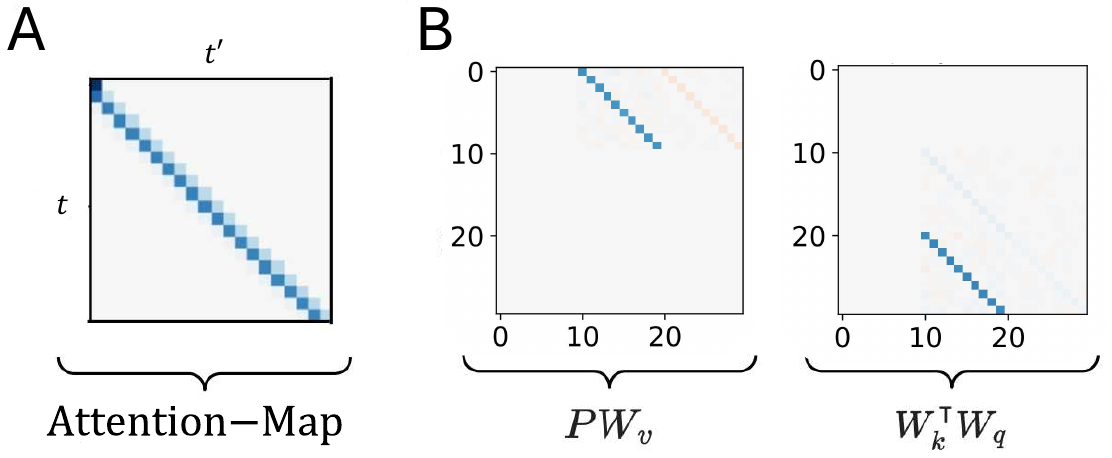}
    \caption{Visualization of activations and parameters for trained models. (\emph{A}): Additional evidence for a token binding mechanism on a 7-layer Transformer complementing Fig.~\ref{fig:induction-head-copying}, shown by plotting first-layer attention scores averaged over a batch of 2048 sequences. Clear data-independent attention on the previous and current token is shown resp.~by high sub-diagonal and main diagonal  attention, with zero everywhere else. (\emph{B}): One trained layer of linear self-attention  implements one step of gradient descent, compare with Proposition~\ref{prop:mesa-gradient-descent-construction}.}
    \label{fig:trained_weights}
\end{figure}

We thus conclude that training robustly installs a token binding mechanism in the first Transformer layers across a range of next-token prediction tasks and network architectures. Interestingly, this mechanism exactly coincides with the first layer of the induction head circuit \citep{elhage_mathematical_2021,olsson_-context_2022,singh_transient_2023}, which has inspired the design of new neural architectures \citep{fu_hungry_2023,arora_zoology_2023,poli_hyena_2023,de_griffin_2024}. Through their analysis of Transformers trained on natural language modeling, Olsson et al.~provide compelling evidence that the appearance of this mechanism during training is strongly correlated with improvements in in-context learning performance. Here, we interpret this phenomenon as part of a multi-layer circuit for mesa-optimization. In light of our theory, token binding can be understood as constructing an in-context training set of appropriate input-output associations. Once this step is concluded, the mesa-objective function is defined, and in-context optimization can take place.

\subsection*{Evidence for mesa-optimizers in linear attention-only models}
We proceed with our analysis of trained Transformers, focusing in this section on simplified linear-attention-only architectures, that can in theory be explained by Propositions~\ref{prop:mesa-gradient-descent-construction} and \ref{prop:multi-layer-precond-gd}. Having shown that a token binding mechanism can be learned, and aiming for the simplest deep Transformer setup, in this section we directly feed our models with aggregate token inputs, $e_t=\begin{bmatrix} 0,s_t,s_{t-1}\end{bmatrix}$,
as assumed by our theory. Moreover, we focus on fully-observable linear tasks.

\begin{figure}[t!]
    \begin{center}
    \includegraphics[width=0.47\textwidth]{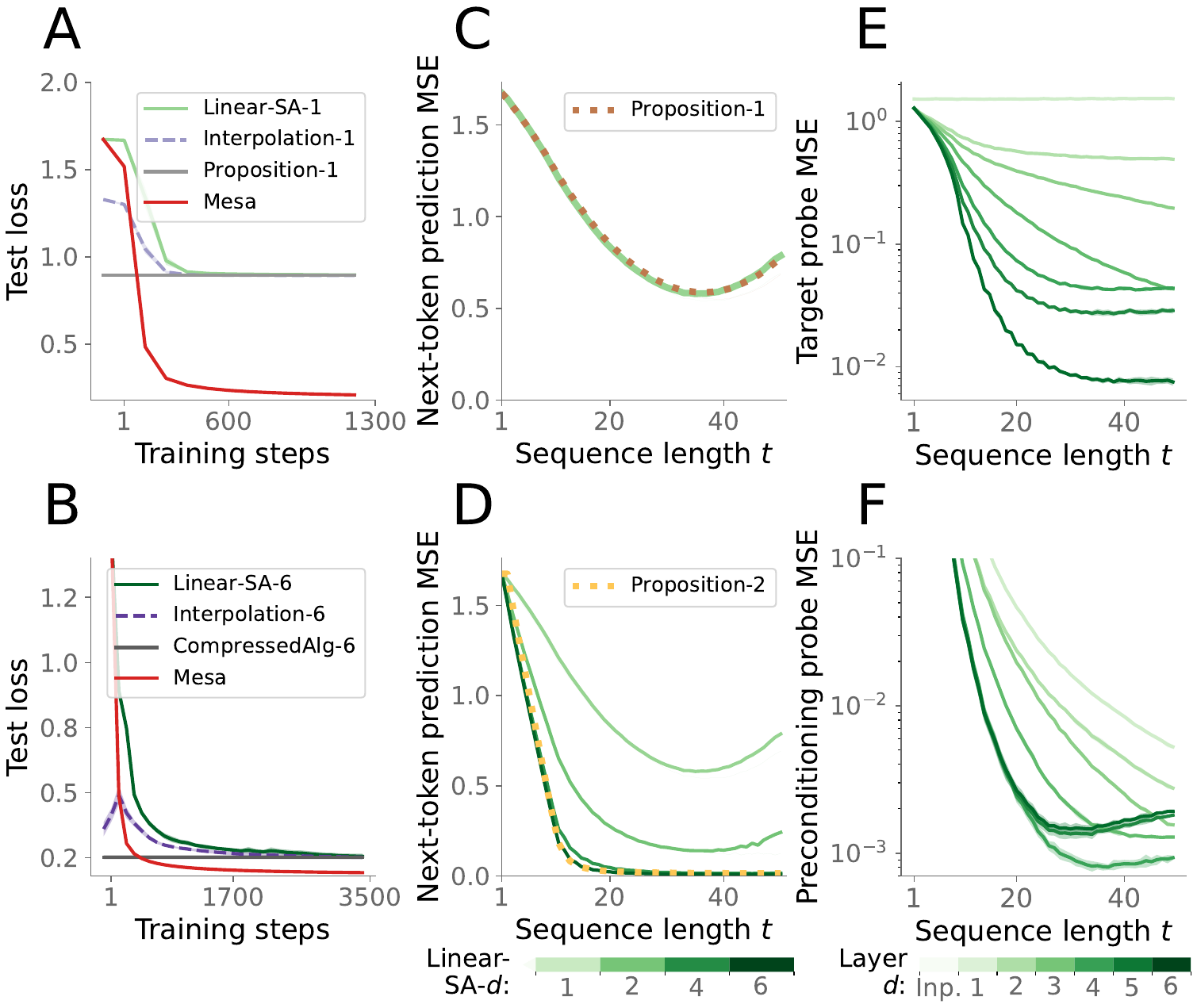}
    \caption{Evidence for mesa-optimization in linear self-attention networks. (\emph{A}) As training proceeds, the test loss of a single layer of linear self-attention (Linear-SA-1, green lines) converges to the loss achieved by 1-step gradient descent (Proposition-1, gray line) with optimized learning rate and initial parameters. A single mesa-layer (red lines) strongly outperforms a single  linear self-attention layer, consistent with the fact that it yields recursively the optimal (least-squares) solution at every time step. (\emph{B}) Same analysis, now for a 6-layer linear self-attention model. The increase in the number of attention layers reduces the gap towards the mesa-layer. The test loss of this model converges to that of the CompressedAlg-6 expression (black line), which comprises a small fraction (0.5\%) of parameters of the original model, reflecting the highly-structured parameters obtained after training. (\emph{C}) At convergence, trained models exhibit the same in-context learning performance (measured as the loss as a function of sequence length) as 1 step of gradient descent (dashed line). (\emph{D}) Similarly for 6-layer models, which can be almost perfectly described by the multi-layer mesa-optimizer of Proposition~\ref{prop:multi-layer-precond-gd} (dashed line). (\emph{E}) Linear probing of next-token targets $s_{t+1}$ from the internal Transformer activations improves with depth and context length, consistent with mesa-optimization for next-token prediction. (\emph{F}) Linear probing of preconditioned inputs $(S_{t-1}S_{t-1}^\top + 1/\lambda I)^{-1}s_t$ improves with depth and context length, consistent with the mesa-optimizer of Proposition~\ref{prop:multi-layer-precond-gd}.}
    \label{fig:reverse-engineering-construction-linear}
    \end{center}
\end{figure}

The results for single-layer networks are strikingly clear. After next-token prediction training, these networks implement the one-step gradient descent algorithm of Proposition~\ref{prop:mesa-gradient-descent-construction} in a near-exact fashion. This can be seen by visual inspection, Fig.~\ref{fig:trained_weights}, or quantitatively by comparing the loss reached by the trained layer with that of a linear autoregressive model learned through one step of gradient descent, cf.~Fig.~\ref{fig:reverse-engineering-construction-linear}A-C. We find that we can perfectly fit the outputs of our trained layer when using all degrees of freedom of our theory, including not only a learned learning rate $\eta$, but also a learned set of initial weights $\Phi_0$. Next-token prediction therefore installs in the Transformer an in-context variant of the model-agnostic metalearning algorithm due to Finn et al.~\citep{finn_model-agnostic_2017}.

Deep linear attention networks correspond to high-degree polynomial functions with a large number of terms. Despite their complexity, for such deep networks training once again leads to highly-structured sparse model parameters $\theta$; we provide visual examples in the \emph{SI Appendix}. This allows us to construct an expression (CompressedAlg-$d$, where $d$ denotes model depth) comprising only 16 parameters (instead of 3200) per layer head. We find that this compressed, albeit convoluted, expression can describe a trained deep linear Transformer. In particular, it allows interpolating between actual Transformer and CompressedAlg-$d$ weights (or Proposition~\ref{prop:mesa-gradient-descent-construction}, for the single-layer case) in an almost lossless fashion, cf.~Figure~\ref{fig:reverse-engineering-construction-linear}C. Further details can be found in \emph{Materials and Methods}.

While the CompressedAlg-$d$ expression explains a trained deep linear self-attention model with a small number of free parameters, it is difficult to interpret it from the lens of mesa-optimization and connect it exactly to the theoretical construction of Proposition \ref{prop:multi-layer-precond-gd}. We therefore resort to a linear probing analysis \citep{alain_understanding_2017} to look for signatures of our hypothesized mesa-optimization algorithms. Based on Propositions~\ref{prop:mesa-gradient-descent-construction} and \ref{prop:multi-layer-precond-gd} we design \begin{inlinelist} \item target probes measuring optimization progress, regressing $k$-th layer representations $e_t^{(k)}$ against the next token $s_{t+1}$ to be predicted, where we could expect multiple steps of gradient descent gradually approaching the target; and \item preconditioning probes regressing against preconditioned inputs $(S_{t-1}S_{t-1}^\top + 1/\lambda I)^{-1}s_t$\end{inlinelist}, cf.~\emph{Materials and Methods}. As shown in Fig.~\ref{fig:reverse-engineering-construction-linear}D-E we see that both probes succeed, with linear decoding performance increasing with sequence length and network depth. Base-optimization has therefore discovered a hybrid algorithm that descends over layers the mesa-objective $L_t(\Phi)$ while simultaneously improving the condition number of the mesa-optimization problem.
This leads to a fast descent of the mesa-objective.

Examining next-token prediction error, we find that it decreases quickly with depth, cf.~Figure~\ref{fig:reverse-engineering-construction-linear}C, with a 6-layer model coming close to but still not matching a single mesa-layer. The high performance of the mesa-layer in this setup can be explained by the fact that it yields the optimal (least-squares) predictor provided that the correct query, key and value inputs are fed to it. Moreover, prediction error decreases monotonically with sequence length both for the mesa-layer as well as for multi-layer linear Transformers. This improvement with context size matches the operational definition of in-context learning proposed by Kaplan et al.~\citep{kaplan_scaling_2020}; in this sense, the models are strong in-context learners, behaving similarly to regularized least-squares. Notably, we see in Fig.~\ref{fig:reverse-engineering-construction-linear} that performance-wise a deep model with $k$ linear attention layers can be almost perfectly explained by $k$ steps of the multi-layer mesa-optimizer described in Proposition~\ref{prop:multi-layer-precond-gd}, with appropriately tuned hyperparameters (cf.~\emph{Materials and Methods}). Importantly, these hyperparameters are tuned for maximal $k$-step performance and not to reproduce Transformer behavior. This is one additional point of evidence that our theoretical mesa-optimizers describe the computations performed by Transformers trained by next-token prediction error minimization.

\subsection*{Trained softmax self-attention layers behave like linear attention}

We return to standard Transformer models, which feature MLPs, LayerNorm and softmax self-attention layers. We train multi-layer versions of such networks on fully-observed linear tasks, under a standard tokenization scheme ($e_t = s_t$). Recalling that our theoretical mesa-optimizers (Propositions~\ref{prop:mesa-gradient-descent-construction} and \ref{prop:multi-layer-precond-gd}) rely on linear self-attention operations, we now ask whether base-optimization renders the softmax attention nonlinearity in an approximately linear regime, when driven by sequences such as those seen during training. 
\begin{figure}[htbp!]
    \begin{center}      \includegraphics[width=0.47\textwidth]{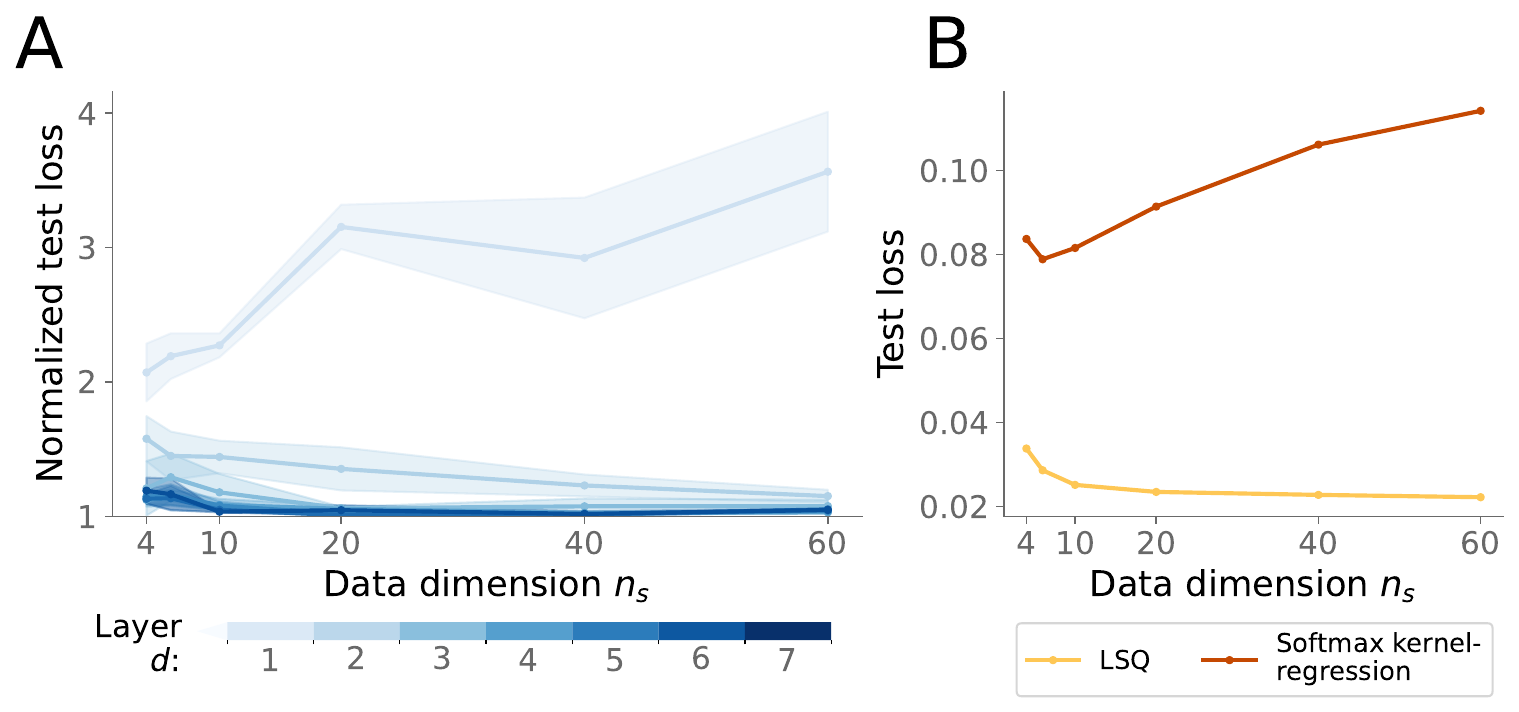}
    \caption{Linearization analysis of softmax Transformers. (\emph{A}) The test loss achieved by a linearized Transformer, where one attention layer at a given depth $d$ (intensity color-coded) is linearized, normalized relative to reference model loss. As the input dimension $n_s$ grows, the linear approximation improves for all layers except for the first. The highly-nonlinear behavior exhibited by this layer is consistent with its special role in implementing a token binding mechanism (Figs.~\ref{fig:illus-prop-1} and \ref{fig:linearization}). (\emph{B}) The test loss of an autoregressive linear model learned by regularized least-squares (LSQ, yellow line), the algorithm we hypothesize that a trained Transformer implements, does not suffer from the curse of dimensionality, whereas a generic interpolation algorithm (red line) that can be implemented in softmax attention layers does.}
    \label{fig:linearization}
    \end{center}
\end{figure}

In Fig.~\ref{fig:linearization}A, we analyze the test set loss achieved by a Transformer after replacing a softmax self-attention layer by its linear counterpart at a given depth, keeping the architecture otherwise intact. We obtain this control model through a process known as distillation \citep{hinton_distilling_2015}: we first record the outputs produced by the to-be replaced softmax attention layer, when the Transformer is applied to a set of training sequences, and then train a linear attention layer to reproduce these outputs by squared error minimization. As we observe in Fig.~\ref{fig:linearization}, for sufficiently large input dimension $n_s$, from the second layer onwards the linear attention models behave as their reference counterparts to a very good approximation. We further observe that the first attention layer behaves in an entirely different, nonlinear manner. This is consistent with the fact that softmax self-attention can implement near-exact token copying \citep{elhage_mathematical_2021}, as required by our token binding mechanism (cf.~Figs.~\ref{fig:illus-prop-1} and \ref{fig:induction-head-copying}).

\subsection*{On induction heads}
The low linearization error achieved at high enough data dimension seen in Fig.~\ref{fig:linearization} is at odds with previous theories explaining in-context learning as best-match (or nearest neighbor) pattern retrieval, which rely on the softmax nonlinearity \citep{elhage_mathematical_2021,olsson_-context_2022}. To better understand this phenomenon, let us compare the scaling behavior of two competing mechanistic explanations for in-context learning in Transformers, as we let the input dimension $n_s$ grow: the theory studied here, where a linear model is learned by regularized least-squares, and nonparametric regression under a softmax kernel. The latter is in fact the algorithm implemented by the full (two-layer) induction head mechanism \citep{elhage_mathematical_2021,olsson_-context_2022}. While we have seen previously that the first token-binding layer of an induction head circuit is precisely what Propositions~\ref{prop:mesa-gradient-descent-construction} and \ref{prop:multi-layer-precond-gd} require, the subsequent layers differ in the two theories, as we briefly review next.

In basic terms, an induction head predicts the next token by first retrieving the most similar past inputs, and then outputting a similarity-weighted combination of the tokens that appeared afterwards. This yields the next-token prediction $\hat{s}_{t+1}^\text{nn} = \sum_{t^\prime=1}^{t-1} s_{t^\prime+1} \softmax(\beta \, s_{t^\prime}^\top s_t)$. Unlike least-squares mesa-optimizers, this method operates on the highly nonlinear regime of the softmax attention function, with the scalar $\beta \in \mathbb{R}_+$ set large enough so as to approximate single-pattern retrieval ($\beta \to \infty$). Thus, with regards to the linearity of the attention function, an induction head and the mesa-optimizers studied here sit on two opposite extremes.

The theory of nonparametric regression has sought to characterize such interpolants, revealing that generalization error scales in general exponentially with input dimension \citep{gyorfi_distribution-free_2002}. By contrast, it can be shown analytically in the simpler non-autoregressive case that the generalization error is independent of input dimension for optimally-regularized linear regression \citep{krogh_generalization_1992,advani_high-dimensional_2020}, assuming that the task difficulty (measured as the context size per dimension $T/n_s$) is conserved, which is the regime we study here. These theoretical considerations are reflected in the experiments with fully-observed linear dynamics reported in Fig.~\ref{fig:linearization}B, where we report the scaling of cumulative next-token prediction mean-squared error loss for softmax kernel regression with  optimally-tuned $\beta$ (per dimension) against an autoregressive linear model learned by optimally-regularized least-squares (LSQ). We see that next-token prediction performance is always best and only weakly depends on $n_s$ for the latter, whereas it degrades for the former.

The findings presented in Fig.~\ref{fig:linearization}B highlight the merits of performing proper latent variable inference under the correct generative model, over applying a generic interpolation algorithm. This is the curse of dimensionality \citep{bellman_adaptive_1961}, here unveiled at the level of in-context learning. One strategy to deal with this problem is to embed the data in an appropriate learned space before applying a nearest-neighbor-type method \citep{snell_prototypical_2017}. For the synthetic autoregressive tasks considered in this paper, the curse of dimensionality can be defeated if base-optimization discovers the multi-layer mesa-optimizer of Proposition~\ref{prop:multi-layer-precond-gd}. Below, we provide further evidence that this actually occurs in trained Transformers.

\begin{figure}[htbp!]
    \begin{center}      \includegraphics[width=0.47\textwidth]{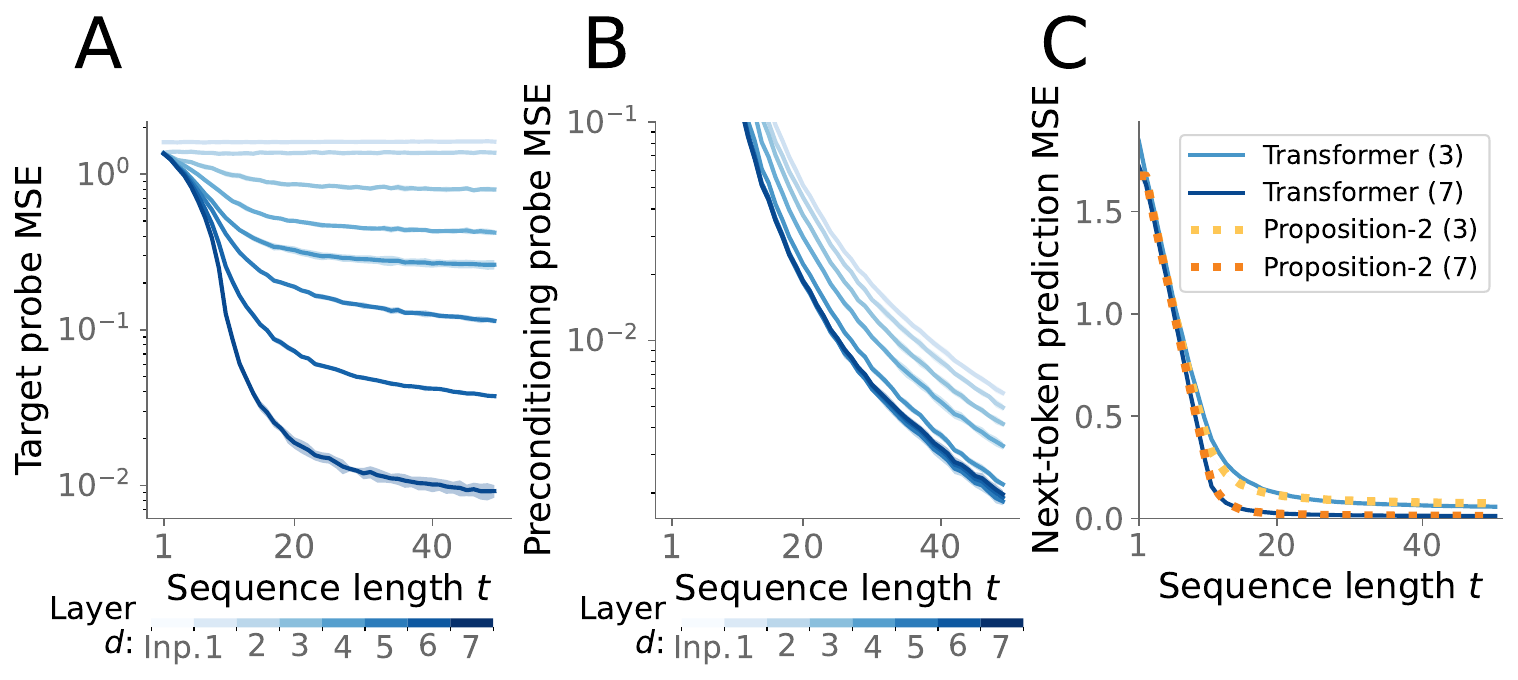}
    \caption{Evidence for mesa-optimization in standard (softmax) Transformers. \emph{(A)} Linear probes decode next-token target $s_{t+1}$ from internal Transformer activations, with decoding performance improving with depth (intensity color-coded) and context length, consistent with gradual optimization of an internal next-token prediction model. \emph({B}) Likewise for preconditioned input $(S_{t-1}S_{t-1}^\top + 1/\lambda I)^{-1}s_t$ probing, consistent with the mesa-optimizer of Proposition~\ref{prop:multi-layer-precond-gd}. \emph{(C)} Next-token prediction error of a 3-layer and a 7-layer Transformer (light and dark blue lines) decrease with context length in almost exactly the same way as 3 or respectively 7 steps of Proposition~\ref{prop:multi-layer-precond-gd} (light and dark dashed yellow lines), with hyperparameters of the latter set for best performance, not to match Transformer behavior.}
    \label{fig:full-fledged}
    \end{center}
\end{figure}

\subsection*{Mesa-optimization theory describes complete Transformers}
We continue studying complete Transformers, repeating the analyses carried out for deep linear attention models paired with special input tokens, as required by Propositions~\ref{prop:mesa-gradient-descent-construction} and \ref{prop:multi-layer-precond-gd}. Moreover, we examine all three task types — linear systems with either full or partial observability, as well as nonlinear systems — now always using conventional input token formatting ($e_t = s_t$).

\begin{figure}[b!]
    \begin{center}      \includegraphics[width=0.47\textwidth]{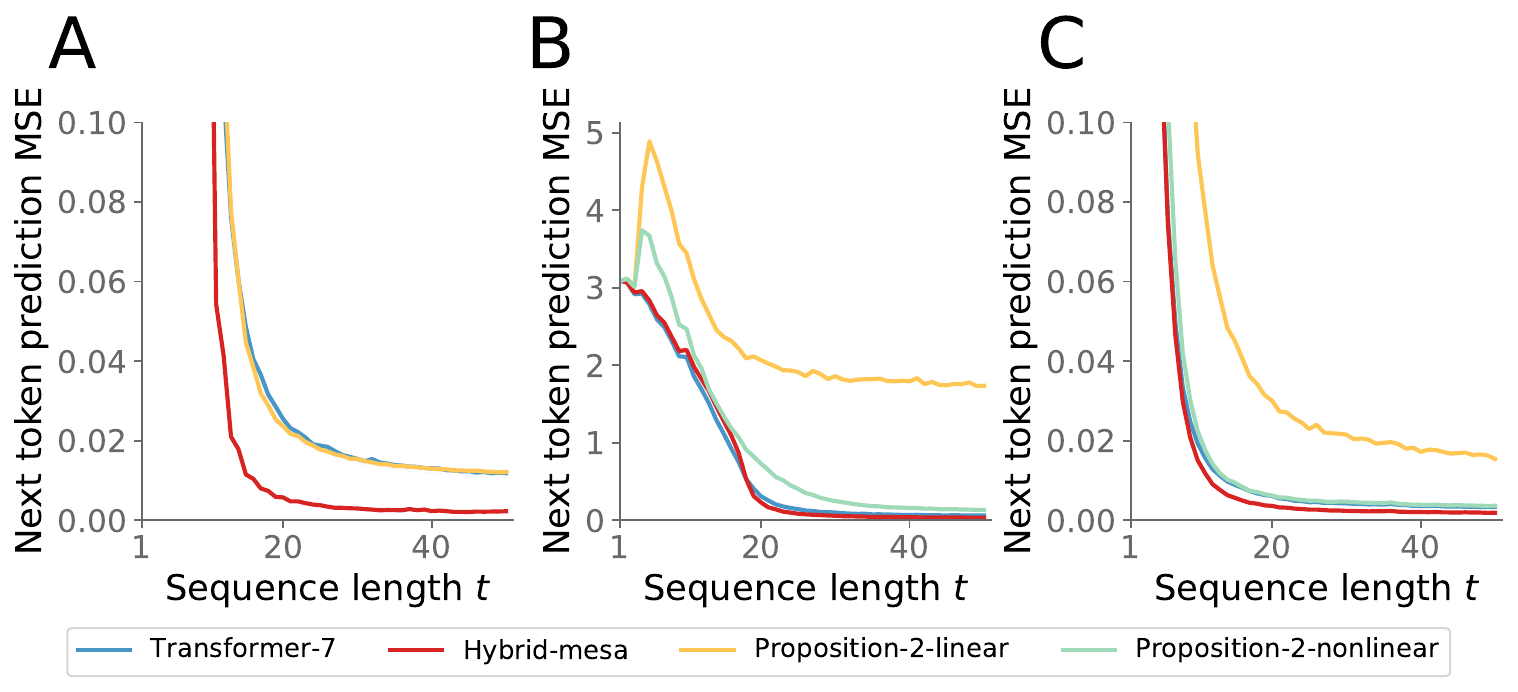}
    \caption{Comparison of the next-token prediction error of 7-layer softmax Transformers (blue lines) and 2-layer softmax-mesa Transformers (red lines) on three families of tasks: fully-observed linear systems (\emph{A}), partially-observed linear systems (\emph{B}), and nonlinear systems (\emph{C}). To validate the mesa-optimization theory developed here, we also report the performance achieved after applying 7 steps of the mesa-optimizer of Proposition~\ref{prop:multi-layer-precond-gd} to learn the parameters of a linear model (Proposition-2-linear; yellow lines). For partially-observed and nonlinear tasks, we further report the loss achieved when the Proposition~\ref{prop:multi-layer-precond-gd} is used to train a linear model applied to the groundtruth feature transformation, given by an optimal number of concatenated past tokens to resolve partial observability, or the $\text{MLP}^*(s_t)$ used by the nonlinear sequence generator, respectively (Proposition-2-nonlinear; light blue lines). These two control models accurately describe the behavior of actual trained standard Transformers. Moreover and also in accordance with the theory developed here, the hybrid-mesa architecture serves as a strong baseline for all three tasks. }
    \label{fig:performance}
    \end{center}
\end{figure}

In short, our main findings on simplified linear attention-only models translate to standard Transformers. We have already seen in Fig.~\ref{fig:induction-head-copying} that these models learn appropriate MLP basis functions when faced with nonlinear tasks, and that they construct internal training sets by binding tokens together. Repeating our probing analysis, we now confirm that  subsequent layers execute an algorithm that simultaneously improves next-token predictions and mesa-optimization conditioning (Fig.~\ref{fig:full-fledged}), as it was the case for linear attention-only models.

In terms of next-token prediction performance, we see that $k$ steps of Proposition~\ref{prop:multi-layer-precond-gd} can essentially describe the performance of $k$-attention-layer Transformers trained on all three task types considered here (Figs.~\ref{fig:full-fledged}C and \ref{fig:performance}), once again in line with our previous findings on simplified linear attention-only models. Moreover, we find that a hybrid two-attention-layer architecture, stacking one mesa-layer after a standard softmax attention layer, is the strongest of all models considered here despite its low parameter count and shallow depth. This hybrid architecture design is directly inspired by our mesa-optimization theory, leveraging the fact that softmax attention layers can easily implement a token binding operation, and that mesa-layers implement efficient in-context least-squares solvers. The fact that a fixed-depth, 2-layer softmax-mesa Transformer provides a performance upper bound approached as the depth of standard softmax Transformers increases provides additional evidence that such models are well described by the mesa-optimization theory developed here.

\subsection*{Autoregressive Transformers are few-shot learners}

Brown et al.~\citep{brown_language_2020} established in-context learning in large autoregressive language models, showing that LLMs can solve new tasks when provided with a small number of (`few-shot') labeled examples in-context. Here, we investigate whether a similar phenomenon occurs in the autoregressive models studied thus far. To that end, we take the Transformers analyzed above and present them post-training with in-context linear regression tasks (cf. \emph{Materials and Methods}).

\begin{figure}[t!]
    \begin{center}      \includegraphics[width=0.47\textwidth]{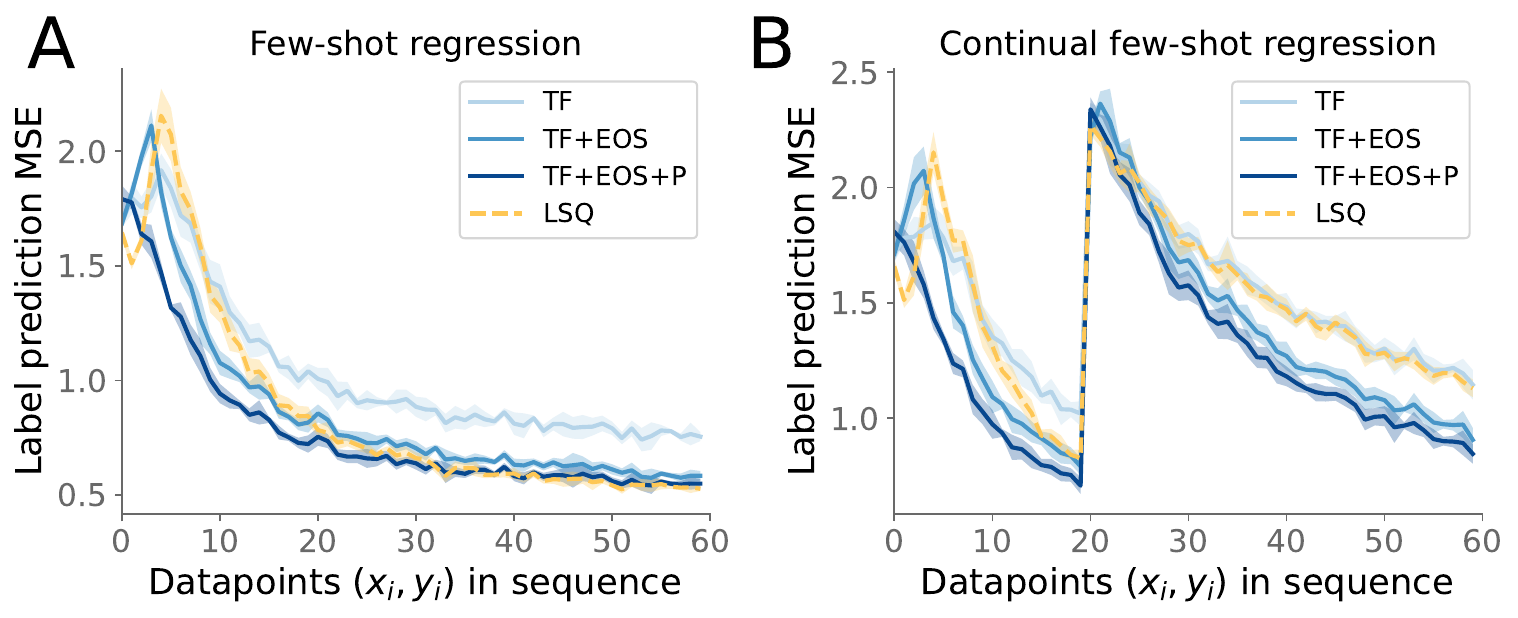}
    \caption{Autoregressive Transformers display in-context few-shot learning capabilities. After training a standard 7-layer Transformer on autoregressive sequence prediction problems, we measure its ability to solve linear regression tasks in-context, without further parameter fine-tuning. The task training set is presented to the model in sequence, with each token corresponding either to an input or to its corresponding label. A final test input is provided and the loss is measured after completing the sequence using the autoregressive Transformer. (\emph{A}) The mesa-optimizers installed by autoregressive pretraining can be leveraged off-the-shelf to solve in-context supervised regression tasks, but yield sub-optimal regression performance (lightest blue lines). In-context learning performance can be improved following the standard strategies of prompt (TF+EOS, light blue lines) and prefix fine-tuning (TF+EOS+P, dark blue lines). For comparison, we provide the loss achieved by an autoregressive linear model learned by least-squares (LSQ, yellow lines) (\emph{B}) Same analysis, now presenting two tasks in a row. The autoregressive models develop some in-context continual learning capabilities. \label{fig:icl}}
    \end{center}
    
\end{figure}

Despite the fact that the models were trained to predict the evolution of linear dynamical systems, and not to perform supervised in-context learning, we observe that regression loss decreases with sequence length (Fig.~\ref{fig:icl}A). The models can thus use additional in-context training data to improve predictions. Our results therefore show that training Transformers on simple autoregressive tasks can give rise to in-context few-shot learning, complementing previous evidence for this phenomenon in large-scale models \citep{brown_language_2020}. As a control, we report the performance reached by autoregressive least-squares on the same dataset, which yields a similar error curve. 

We note that the autoregressive in-context learning algorithm uncovered above is sub-optimal with respect to linear regression. Close inspection reveals that the origin of its sub-optimality lies in the learned token binding mechanism (analyzed in Fig.~\ref{fig:induction-head-copying}) that binds every consecutive pair of tokens, in an overlapping fashion. In a training set of size $n$, this introduces $n-1$ spurious associations, where a regression target $y_i$ is incorrectly associated to the next independent input $x_{i+1}$, whereas only inputs $x_i$ should be associated with their respective targets $y_i$. Interestingly, this gives rise not only to convergence to a sub-optimal solution, but also to the early ascent phenomenon present in LLMs \cite{lin_dual_2024}. This refers to in-context learning performance first undergoing a brief but statistically significant period of loss increase, before actual improvements start taking place. Note that early ascent is not specific to autoregressive Transformers; we can observe it on the autoregressive linear least-squares control model as well (LSQ; Fig.~\ref{fig:icl}A). We therefore identify one cause for this poorly-understood phenomenon, tracing it back to the internal mechanics of mesa-optimization for next-token prediction.

To mitigate this effect, we investigate a common approach, known as prompt-tuning, which can lead to significant performance improvements when applied to large language models \citep{li_prefix-tuning_2021,lester_power_2021}. Concretely, we fine-tune a single token, which we refer to as the $\texttt{EOS}$ token, on the linear regression objective. When presenting data sequentially as $[x_1, y_1, \texttt{EOS}, x_2, y_2,  \dots , \texttt{EOS}, x_N, y_N]$, where $x_i$ and $y_i$ resp.~denote regression inputs and labels, we observe a considerable performance improvement after prompt-tuning, see Fig.~\ref{fig:icl}A. Furthermore, to instruct the model to perform few-shot tasks, we learn a single prefix-prompt $\texttt{P}$ which we append at the beginning of a sequence with \texttt{EOS} tokens. This appears to further improve the few-shot performance for early data-pairs. Additional experimental details can be found in \emph{Materials and Methods}.

Lastly, we demonstrate the capability of autoregressive Transformers to learn multiple tasks in a row. We study the minimal setup where the model has to learn two tasks, generated from two distinct groundtruth linear models, resulting in a sequence of data of the form $[x^1_1, y^1_1, \dots , x^1_N, y^1_N, x^2_1, y^2_1, \dots , x^2_N, y^2_N]$. In Fig.~\ref{fig:icl}B, we see that the trained Transformer can learn a second task in-context, even though it was never explicitly trained to solve such sequential learning problems.  This behavior is expected, given the autoregressive linear model optimizer uncovered in the preceding sections. This finding suggests further characterizing the continual in-context learning abilities of Transformers, as Irie et al.~\citep{irie_automating_2023} have begun to investigate.

\section*{Discussion}
\label{sect:discussion}
We've presented evidence that Transformer models develop gradient-based learning algorithms when trained on sequence prediction tasks under a standard autoregressive objective. Moreover, we have seen that the resulting prediction algorithms can be repurposed without retraining to solve supervised in-context learning tasks, capturing LLM phenomena such as early ascent or the effectiveness of prompt fine-tuning techniques in improving in-context learning. The fact that we were able to reproduce these findings in our synthetic data setup is surprising, given that the state-space sequence generators studied here are far from language models---most notably, they operate in continuous space, and lack deep hierarchical structure. Our results serve as a case-in-point that autoregressive Transformers can exhibit in-context learning capabilities outside language modeling, and point towards the universality of certain properties of these acquired learning algorithms.

There has been significant debate on whether LLMs, and learned next-token predictors more generally, are limited to memorizing correlations present in the training set \citep[having been called stochastic parrots;][]{bender_dangers_2021}. This view has been challenged by a number of studies, analyzing for example autoregressive models trained to predict legal moves in board games \citep{toshniwal_chess_2022,li_dissecting_2023,nanda_emergent_2023}. In a purely observational manner and without any {\it a priori} game knowledge, self-supervised next-token prediction models learn latent representations of the board state and track the moves of each opponent. Our findings provide complementary evidence that next-token prediction objectives can lead to the discovery of algorithms that correctly infer the hidden state of the world: the in-context learning algorithm we identified can be precisely cast as maximum {\it a posteriori} inference under the correct Bayesian prior and likelihood function. Moreover, the multi-layer mesa-optimizers installed by next-token prediction objectives are highly efficient (i.e., achieve significant loss reduction in only a few layers) thanks to precise tuning of their hyperparameters to the sequence generative model. 

The idea that a Transformer generates its predictions by solving internal optimization problems has ties to many different lines of thought in machine learning. One closely related line of work explores the concept of a declarative node: a differentiable layer whose output is defined implicitly as the solution of an optimization problem \citep{amos_optnet_2017,gould_deep_2021,zucchet_beyond_2022}. We note that subsuming an entire chain of layers by a single declarative node is not only potentially 
more efficient, but also more interpretable. 
The mesa-layer is an example of such a node, adding to recent studies exploring the advantages of including declarative nodes within attention-based models \citep{ramsauer_hopfield_2021,martins_sparse_2020,garnelo_exploring_2023,hoover_energy_2023}.

Our analysis of trained models revealed that stochastic gradient descent in effect discovered a declarative node, preferring to pick an optimization algorithm among alternative solutions in the configuration space of autoregressive Transformers. This can be partly explained by the fact that recursive least-squares can be leveraged to solve the tasks considered here, and by the fact that Transformers can efficiently approximate this algorithm through Proposition~\ref{prop:multi-layer-precond-gd}. Our results complement the theoretical work of Hubinger et al.~\citep{hubinger_risks_2019}, by providing a concrete toy model where mesa-optimization occurs. However, more work is still needed to characterize this phenomenon outside the controlled experimental setting considered in this paper.

The mesa-layer developed here can also be seen as a locally optimal fast weight programmer \citep{schmidhuber_learning_1992}. In his seminal work \citep{schmidhuber_learning_1992}, Schimidhuber proposed to dynamically reprogram the weights of a feedforward neural network using a Hebbian rule. As pointed out by Schlag et al.~\citep{schlag_linear_2021}, this is precisely what a linear self-attention layer does: it generates predictions using an effective weight matrix that is learned by taking outer products of values and keys, a Hebbian associative rule \citep{hebb_organization_1949}. In this work, we instead frame fast weight learning as an optimization problem that is efficiently solved at every moment in time by the mesa-layer. This form of optimal fast learning is strictly superior to Hebb's rule, both in terms of generalization and memory capacity \citep{hertz_introduction_1991}. The mesa-layer is therefore also closely related to the Delta-Net \citep{schlag_linear_2021}, which uses the delta rule \citep{widrow_adaptive_1960} for fast weight learning. Unlike the mesa-layer, which is optimal at every step, the delta rule requires multiple steps to converge, though it is cheaper to implement. The strong performance of the mesa-layer observed here on synthetic tasks suggests investigating its application to natural data at larger scales, for which we provide preliminary language modeling results in Appendix \ref{apx:language}.

Our work has an unexpected connection to research on local learning rules, a question of great interest in theoretical neuroscience \citep{lillicrap_backpropagation_2020}. Decomposing a global learning problem into a series of local quadratic optimization problems, like the objective functions of the mesa-optimizers studied here, is at the heart of the target propagation \citep{lee_difference_2015}, predictive coding \citep{whittington_approximation_2017} and control-based \citep{meulemans_least-control_2022} theories of learning in the brain. Moreover, previous studies have proposed greedy layerwise learning algorithms that do not require global error information \citep{hinton_fast_2006,nokland_training_2019,belilovsky_greedy_2019,lowe_putting_2019,hinton_forward-forward_2022}. Much in the same vein but now on the fast timescale of inference, the mesa-optimizers uncovered here implement greedy, local learning algorithms which only use bottom-up information.

We conclude by discussing our findings in the light of predictive processing theories of intelligence, where learning predictive models is presumed to underwrite intelligent behavior \citep{hawkins_intelligence_2004,clark_whatever_2013}. A number of influential predictive processing models have adopted a Bayesian approach, starting from the assumption that the world obeys a certain generative model, and then hand-designing approximate inference algorithms for the assumed model \citep{hinton_wake-sleep_1995,rao_predictive_1999,lee_hierarchical_2003,friston_free_2006,keller_predictive_2018}. Here, directly inspired by LLMs, we took a powerful neural sequence model and trained it to maximize the likelihood of upcoming inputs given the past, without making explicit probabilistic assumptions about the latent structure of the world. The network was nonetheless able to discover the correct underlying model of the data, and appropriately exploit its knowledge to generate predictions.
This finding provides further evidence that direct maximization of future prediction performance by simple gradient-based methods --- as opposed to hierarchical probabilistic methods, and the typically intractable inference problems that they bring --- might be sufficient to build the predictive processing backbone of an intelligent system.

\section*{Methods}

\subsection*{Transformer architectures} The Transformer models studied here follow the widely-used GPT-2 specification \cite{radford_language_2018}. This architecture comprises multiple identical blocks, with one block consisting of the softmax self-attention layer defined in equation \ref{eq:attention-dynamics} followed by a one-hidden layer MLP. The inputs of both layers are normalized:
\begin{align}
        e_{t} \leftarrow e_{t} + \Delta e_{t}^{\text{sa}}(\text{LN}(e_t)) \nonumber\\
    e_{t} \leftarrow e_{t} + \Delta e_{t}^{\text{mlp}}(\text{LN}(e_t)),\nonumber
\end{align}
where $\text{LN}(\cdot)$ denotes the LayerNorm operation \cite{ba_layer_2016}, and $\Delta e_{t}^{\text{mlp}}(e_t) = W_2\text{GELU}(W_1e)$ with $\text{GELU}(e) := e \,\mathrm{G}(e)$, and $\mathrm{G}(\cdot)$ the Gaussian cumulative distribution function, applied elementwise \citep{hendrycks_gaussian_2016}. We set $W_1$ such that $W_1e$ has four times more neurons compared to $e$, which itself is four times larger than $s$. Additional architectural details are provided in the \emph{SI Appendix}.

The predictions are read-out directly from the first dimensions of last-layer token outputs, and we add a positional encoding to every input following the original method of Vaswani et al.~\citep{vaswani_attention_2017}. If not explicitly stated otherwise, for models that incorporate the mesa-layer, we leave the architecture configuration unchanged but replace $\Delta e_{t}^{\text{sa}}$ with $\Delta e_{t}^{\text{mesa}}$ in the appropriate places. A hybrid-mesa Transformer features two self-attention layers, with the first being standard softmax self-attention, and the second a mesa-layer.

\subsection*{Base optimizers} All models are trained by online autoregressive loss (Eq.~\ref{eq:autoregressive-loss}) minimization using the AdamW \cite{loshchilov_decoupled_2019} optimizer with learning rate warm-up followed by a cosine decay.

\subsection*{Statistics}
All numerical results are averaged across five random seeds, with shaded areas representing standard deviation.

\subsection*{Synthetic sequence generators}
The tasks considered in this paper involve predicting the next observation $s_{t+1} \in \mathbb{R}^{n_s}$ from a sequences of past observations $(s_{t^\prime})_{t^\prime=1}^t$ generated by discrete-time dynamical systems, whose state is denoted by $h_t \in \mathbb{R}^{n_h}$. Starting from a random initial state $h_1 \sim \mathcal{N}(0,1)$, we generate observations by letting a groundtruth system evolve according to
\begin{align}
h_{t+1} &= W^* f^*(h_t) + \epsilon_{h,t} \nonumber\\
s_t &= C^* h_t + \epsilon_{s,t}, \nonumber
\end{align}
where $\epsilon_{h,t} \sim \mathcal{N}(0,\sigma^2_h)$ is a noise input and $\epsilon_{s,t} \sim \mathcal{N}(0,\sigma^2_s)$ is an observation noise term. We set the transition matrix $W^* \in \mathbb{R}^{n_h \times n_h}$ to a random orthogonal matrix, and we consider both fully-observed ($C^* = I$) and partially-observed tasks, where $n_s < n_h$, and $C^*_{ij} \sim \mathcal{N}(0,0.5)$. Our tasks can be further categorized as linear (by setting $f^*$ to the identity function) or nonlinear. For the nonlinear case, we always take $C^*=I$ and introduce a nonlinear transformation $\text{MLP}^*(\cdot)$ in state-space, $h_{t+1} = W^* \, \text{MLP}^*(h_t) + \epsilon_t$. The MLP computation is described by $\text{MLP}^*(h_t) = B\cdot\text{GELU}(A\cdot h_t)$, where $A \in \mathbb{R}^{n_m \times n_h} \sim \mathcal{N}(0,1.1)$ and $B \in \mathbb{R}^{n_h \times n_m} \sim \mathcal{N}(0,1.1)$.

Importantly, we draw new transition and readout matrices $W^*$ and $C^*$ for every sequence. These parameters are analogous to task-specific variables in multi-task learning \cite{caruana_multitask_1997}, adapted to the problem of unsupervised sequence modeling. We introduce sequence-specific variables to reflect the high degree of variability that is observed in large datasets of real-world data, such as in LLM training corpora \citep{radford_language_2018}. Under such a generative model, rote memorization solutions are excluded from the global minimizers of Eq.~\ref{eq:autoregressive-loss}: a trained Transformer cannot achieve minimal loss by memorizing a single set of $W^*$ and $C^*$ in its parameters $\theta$. Instead, it must deal with inherent uncertainty in every sequence, and infer in-context a set of latent variables whose values vary from sequence to sequence. The main goal of this paper is to characterize this in-context inference process. 

\subsection*{Proof of Proposition 1}
Starting with the token construction 
$e_t = \begin{bmatrix} \Phi_0 s_t,s_t,s_{t-1}\end{bmatrix}$, we now show that the parameter construction of Proposition \ref{prop:mesa-gradient-descent-construction} induces the following gradient-based change to all tokens in parallel 
$e_t \leftarrow \begin{bmatrix} (\Phi_0 - \eta \nabla L_t(\Phi_0))s_t,s_t,s_{t-1}\end{bmatrix}$.
When plugging in the proposed weights into a linear self-attention layer head we obtain
\begin{align*}
    \begin{bmatrix}
\Delta\hat{\Phi}_ts_t \\
 0 \\
  0  \\
\end{bmatrix} & = PW_V \sum_{t'=1}^{t} \begin{bmatrix}
0 \\
 s_{t'} \\
   s_{t'-1}
\end{bmatrix}\begin{bmatrix}
0 \\
 s_{t'} \\
   s_{t'-1}
\end{bmatrix}^{\mathbf{\intercal}} \begin{bmatrix}
 0 & 0 & 0  \\
  0 & 0 & 0  \\
 0 & I_s & 0  \\
\end{bmatrix} \begin{bmatrix}
0 \\
 s_{t}  \\
   s_{t-1}  \\
\end{bmatrix} \\
& = PW_V \sum_{t'=1}^{t} \begin{bmatrix}
0 \\
 s_{t'} s_{t'-1}^\top s_t \\
   s_{t'-1}s_{t'-1}^\top s_t
\end{bmatrix} \\
& =\begin{bmatrix} 
 0 & \eta I_s
 & -\eta\Phi_0  \\
  0 & 0 & 0  \\
 0 & 0 & 0  \\
\end{bmatrix} \sum_{t'=1}^{t} \begin{bmatrix}
0 \\
 s_{t'} s_{t'-1}^\top s_t \\
   s_{t'-1}s_{t'-1}^\top s_t
\end{bmatrix} \\ & =  -\eta  \sum_{t'=1}^{t}  \begin{bmatrix}
 (\Phi_0 s_{t'-1} - s_{t'})s_{t'-1}^\top s_t \\
 0 \\
 0 \\
\end{bmatrix} = \begin{bmatrix}
 -\eta \nabla L_t(\Phi_0)\\
 0 \\
 0 \\
\end{bmatrix}. 
\end{align*}
Adding the above result to the layer input, an operation that is supported in Transformers by a residual connection or a second attention head, yields the desired output.

\subsection*{Full statement and proof of Proposition 2}
We present here the linear self-attention parameter construction which supports the claim of Proposition~\ref{prop:multi-layer-precond-gd}. First, we restate the goal of the autoregressive Transformer, namely, to solve a regularized least-squares problem:
\begin{equation}
\min_\Phi \sum_{t^\prime=1}^{t-1} \frac{1}{2} \|s_{t^\prime+1} - \Phi s_{t^\prime} \|^2 + \frac{1}{2\lambda}||\Phi||_\mathrm{F}^2,\nonumber
\end{equation}
for all time steps simultaneously. This amounts to computing a (recursive) least squares solution, where time-shifted (by one) sequence elements play the role of inputs and desired outputs in a dataset, with inputs $S_{t-1}$, targets $S_{t}$, and test input $s_t$.

With the limited expressivity of one layer, we have already established that Transformers can, and do, implement a single gradient step on the corresponding regression problems $\sum_{t'=1}^{t-1}\|s_{t'+1} - \Phi s_{t'}\|^2 \quad \forall t$ in parallel both in theory and in practice. The key observation here is that given a preconditioning matrix $ H^*_{t}=(S_{t-1}S_{t-1}^\top  + \frac{1}{\lambda} I)^{-1}$ which changes the loss to $\sum_{t'=1}^{t-1}\|s_{t'+1} - \Phi H^*_{t}s_{t'}\|^2$, a single gradient descent step immediately yields the desired regularized least-squares solution.

Based on this simple observation, we provide a theoretical construction that shows how Transformers can approximate $(S_{t-1}S_{t-1}^\top  + \frac{1}{\lambda} I)^{-1}s_{t}$ layer by layer in their forward pass, leading to improved single-step gradient descent performance. Note that this is equivalent to iteratively solving the systems of linear equations $\{(S_{t^\prime-1}S_{t^\prime-1}^\top  + \frac{1}{\lambda} I) x =  s_{t'}\}_{t'=1}^t$. 
Let us now approximate the above expression with a truncated Neumann series:
\begin{align}
    H^*_{t}s_{t} & \approx 
     \tilde{s}^{K=1}_t = \sum_{k=0}^K (I-(S_{t-1}S_{t-1}^\top  + \frac{1}{\lambda} I))^k s_{t} \nonumber\\
    & = \sum_{k=0}^{K} ((1 - \frac{1}{\lambda})I - S_{t-1}S_{t-1}^\top)^ks_{t} \nonumber\\
    & = \tilde{s}^{K}_t + ((1 - \frac{1}{\lambda})I - S_{t-1}S_{t-1}^\top)\tilde{s}^{K}_t = \tilde{s}^{K}_t + \tilde{H}_t^*\tilde{s}^{K}_t \nonumber
\end{align}
with $\tilde{H}_t^* := ((1 - \frac{1}{\lambda})I - S_{t-1}S_{t-1}^\top)$.
This corresponds to the Richardson iteration \cite{rirchardson_approximate_1911} method for solving linear systems iteratively, which can be augmented with a stepwise parameter (or learning rate) $\alpha_K$ and an additional term adding the difference between former approximations, resembling a momentum term. This variant is termed second-order Richardson or Chebyshev \cite{golub_chebyshev_1961} iteration, and it can speed up convergence:
\begin{equation}
\label{eq:chebyshev}
    \tilde{s}^{K+1}_t  = \tilde{s}^{K}_t - \alpha_K\tilde{H}_t^*\tilde{s}^{K}_t - \beta_K(\tilde{s}^{K}_t - \tilde{s}^{K-1}_t).
\end{equation}

We now show that a single step of these iteration methods can be mapped to a single layer of linear self-attention, allowing deep Transformers to solve the aforementioned set of linear equations efficiently in parallel. Starting with a token construction similar to the one of Proposition 1, i.e., with aggregate tokens $\begin{bmatrix}
\tilde{s}_{t'}^{K} ,
\tilde{s}_{t'}^{K-1},
   s_{t'-1}
\end{bmatrix}$ with $\tilde{s}_t^{0} = s_t$, we can compute $\tilde{s}_t^{K+1}$  with a single causally masked linear self-attention, in parallel for $\forall t$. Indeed, with $W_k^\top W_q= \begin{bmatrix}
0 & 0 &  -\alpha_K I_s  \\
0 & 0 & 0  \\
 0 & 0 &  0  \\
\end{bmatrix}$ and $PW_v = \begin{bmatrix}
 0 & 0 & 0  \\
  0 & 0 & 0  \\
 - \alpha_K I_s & 0 &  0  \\
\end{bmatrix}$ the linear self-attention equation, similar to the derivation above, results in $PW_v \sum_{t'=1}^{t}  \begin{bmatrix}
\tilde{s}_{t'}^{K} \\
\tilde{s}_{t'}^{K-1}\\
   s_{t'-1}
\end{bmatrix}\begin{bmatrix}
\tilde{s}_{t'}^{K} \\
\tilde{s}_{t'}^{K-1}\\
   s_{t'-1}
\end{bmatrix}^{\mathbf{\intercal}} W_k^\top W_q \begin{bmatrix}
\tilde{s}_t^{K} \\
\tilde{s}_t^{K-1}\\
   s_{t-1}
\end{bmatrix} = \begin{bmatrix}
- \alpha_K S_{t-1} S_{t-1}^\top\tilde{s}_t^{K}\\
0 \\
   0
\end{bmatrix}$. Therefore, the matrix-matrix-vector products needed to compute equation \ref{eq:chebyshev} can be computed inside a single linear self-attention layer in parallel, for all time steps. The remaining terms in equation~\ref{eq:chebyshev} are simple scaled additions of $\tilde{s}_t^{K}, \tilde{s}_t^{K+1}$ for which multiple alternative constructions exist. Note that for the construction above to hold, we need to have $s_{t-1}$ available at every layer and push forward $\tilde{s}_t^{K}$ such that it can be used to compute $(\tilde{s}_t^{K+1} - \tilde{s}_t^{K})$ in the next iteration which again is easy to accomplish within the residual stream. 

We therefore conclude that deep Transformer models can approximate the solutions of the set of systems of linear equations $\{(S_{t'-1}S_{t'-1}^\top  + \frac{1}{\lambda} I) x =  s_{t}\}_{t'=1}^t$ efficiently in parallel. This results in a preconditioning of the least-squares problems $\{\sum_{t''=1}^{t'-1}\|s_{t''+1} - \Phi H^*_{t'}s_{t''}\|^2\}_{t'=1}^t$, which can then be solved with a single gradient step, again in parallel and by a single additional linear self-attention layer, built after Proposition \ref{prop:mesa-gradient-descent-construction}.

\subsection*{Mesa-optimizers solve partially-observed linear tasks}
We now show that Propositions~\ref{prop:mesa-gradient-descent-construction} and \ref{prop:multi-layer-precond-gd} can be leveraged to solve next-token prediction problems involving linear latent variable dynamics, as in our experiments with partially-observed linear dynamical systems. We analyze here the deterministic setting, i.e., when no noise is added to the state transitions and observations; for an extension to the stochastic case, see the \emph{SI Appendix}. We investigate a simple construction where we concatenate the last $k$ observations into a single `state' vector $z$, and use this state vector in a least-squares problem to estimate the linear transition between $z_{t+1}$ and $z_t$. As $z_{t+1}$ contains $s_{t+1}$, this state prediction can be used straight-forwardly to predict the next observation. Let us define
\begin{align}
    z_t^k = \begin{bmatrix}
        s_{t-k+1} \\
        \vdots \\
        s_t
    \end{bmatrix}.\nonumber
\end{align}
We first investigate whether the transition between $z^k_{t}$ and $z^k_{t+1}$ is a linear operator. For this, let us define the observation matrix as
\begin{align}
    \mathcal{O}_k = \begin{bmatrix}
        C^* \\
        C^*W^* \\
        \vdots \\
        C^*{W^*}^{k-1}
    \end{bmatrix}.\nonumber
\end{align}
Now we have that $z_t^k = \mathcal{O}_k h_{t-k+1}$ and $z_{t+1}^k = \mathcal{O}_k W^* h_{t-k+1}$. We want to find a matrix $\Phi_k$ such that $z_{t+1}^k = \Phi_k z_{t}^k$. As this should hold for all possible system initializations and hence $h_{t-k+1}$, we have that $\mathcal{O}_k W^* = \Phi_k\mathcal{O}_k$. If $kn_s \geq n_h$, we have an underdetermined or fully-determined (in case of equality) set of linear equations, assuming no rank-deficient matrices. The minimum-norm solution for $\Phi_k$ is given by
\begin{align}
    \Phi_k = \mathcal{O}_k W^* \mathcal{O}_k^{\dagger},\nonumber
\end{align}
with $\mathcal{O}_k^{\dagger}$ the Moore-Penrose pseudoinverse of $\mathcal{O}_k$. If the dimension of the concatenated observations $z^k$ is smaller than the dimension of the groundtruth state $h$ ($kn_s < n_h$), the linear system is overdetermined and in general there does not exist a solution for $\Phi_k$. Hence, in order to do optimal predictions, we need to concatenate enough observations into $z_t^k$ such that $kn_s \geq n_h$.

As there exists a linear map between $z_{t+1}^k$ and $z_t^k$, and $z_{t+1}^k$ can be used directly to predict $s_{t+1}$, a Transformer can solve the least-squares problem in-context on $z_{t+1}^k$. One possible implementation is the following construction:
\begin{inlinelist}
    \item copy the last $k$ observations into a concatenated state vector $z_t^k$;
    \item format tokens as required by Propositions~\ref{prop:mesa-gradient-descent-construction}~and~\ref{prop:multi-layer-precond-gd}, now with $z_t^k$ instead of $h_t$, which can be done by the same self-attention layer as the first step;
    \item solve the mesa-optimization problem by directly leveraging the aforementioned propositions.
\end{inlinelist}

\subsection*{CompressedAlg-$d$} After training a single- or multi-layer linear attention model, we obtain structured matrix products $W_K^\top W_Q, PW_v$ per head and layer. When inspecting the trained weight matrix products, we observe strong block-diagonal structure across all layers. We extract the mean values of these block-diagonals and construct sparse weight matrices, consisting only of identity sub-matrices scaled by the resp.~obtained mean value, and compute the evaluation of this constructed compressed algorithm on test sequences. Then, during a second training run (for the same initial conditions), we compute the test loss achieved by an a control model with interpolated parameters, obtained by averaging (with equal averaging weight) the compressed per-head weight-matrix-products and the actual trained layer parameters. 

\subsection*{Probing analyses}
In Figs.~\ref{fig:induction-head-copying}, \ref{fig:reverse-engineering-construction-linear} and \ref{fig:full-fledged} we show the performance of linear decoders trained to predict certain features (e.g., a given past input token $e_{t^\prime}$, in Fig.~\ref{fig:induction-head-copying}A) from internal model activations at various depths, time steps, and stages of training. For every such probing experiment (i.e., for each layer, context length, or base training step, depending on the analysis at hand) we train a separate linear decoder on a batch of activations to predict the respective probing targets by mean-squared error minimization (linear regression). For the preconditioning probings, we compute the 6-step Chebyshev approximation of $(S_{t'-1} S_{t'-1}^\top  + \frac{1}{\lambda} I) s_{t'}$ at each time step $t'$, and linearly regress the activations after each layer at the respective time step against this preconditioning target.

\subsection*{In-context few-shot learning: generative model}
To generate a few-shot task, we sample a groundtruth $W^*$ random orthogonal matrix as done during training, but now use this groundtruth model to generate a labeled training set $\{x_i,y_i\}_{i=1}^N$, with inputs $x_i \sim \mathcal{N}(0,I_x)$ and targets $y_i = W^*x_i$. We then present this dataset to our autoregressive Transformers as a sequence of tokens, $e^\text{few-shot}=[x_1, y_1, \dots , x_N, y_N ]$ of length $T=2N$, cf.~Figure~\ref{fig:icl}. As the sequence unfolds, and more training data is presented, we measure in-context learning performance through the mean squared error between the Transformer output $f_{\theta}(e_{2i-1};e^\text{few-shot}_{1:2i-1})$ and the corresponding target $y_i = e_{2i}$. We emphasize that both the sequence generative model and loss function differ from the ones used during training; compare the task performance metric  $L^\text{few-shot} = \frac{1}{2}\sum_{i=1}^N \|e_{2i} - f_{\theta}(e_{2i-1};e^\text{few-shot}_{1:2i-1})\|^2$ used to evaluate in-context learning performance in this section with the actual loss used to train the Transformer, Eq.~\ref{eq:autoregressive-loss}.

As a control, we further report the performance reached by the least-squares solution (LSQ) obtained on the dataset $D_N^{\text{mesa}} =\{(x_i, y_i)\}_{i=1}^N \cup  \{(y_i, x_{i+1})\}_{i=1}^{N-1}$, and observe a similar decrease in loss after a phase of early ascent. This dataset,  where half of the associations consist of wrong input-output pairs $D^\text{spurious}_N = \{(y_i, x_{i+1})\}_{i=1}^{N-1}$ as illustrated in Figure~\ref{fig:icl}A, corresponds to the training set an autoregressive Transformer imbued with the mesa-optimizers uncovered in the previous section learns from.

\paragraph{Acknowledgements} João Sacramento and Johannes von Oswald thank Angelika Steger and Jyrki Alakuijala for their support and guidance. The authors also thank Marc Kaufmann, Yassir Akram, Andrey Zhmoginov, Yanick Schimpf, Oliver Sieberling and Luca Versari for fruitful discussions and insights, and to Luke Sernau, Maciej Wolczyk, Simon Schug and Robert T.~Lange for valuable comments on the manuscript. João Sacramento and Nicolas Zucchet were supported by an Ambizione grant (PZ00P3\_186027) from the Swiss National Science Foundation and ETH Research Grant (ETH-23 21-1).

\bibliography{transformers}
\appendix
\onecolumn
\section{Visualization of weights and attention maps of trained multi-layer Transformers}
\label{apx:visualizations}
\begin{figure}[htbp!]
    \centering
    \includegraphics[width=0.8\textwidth]{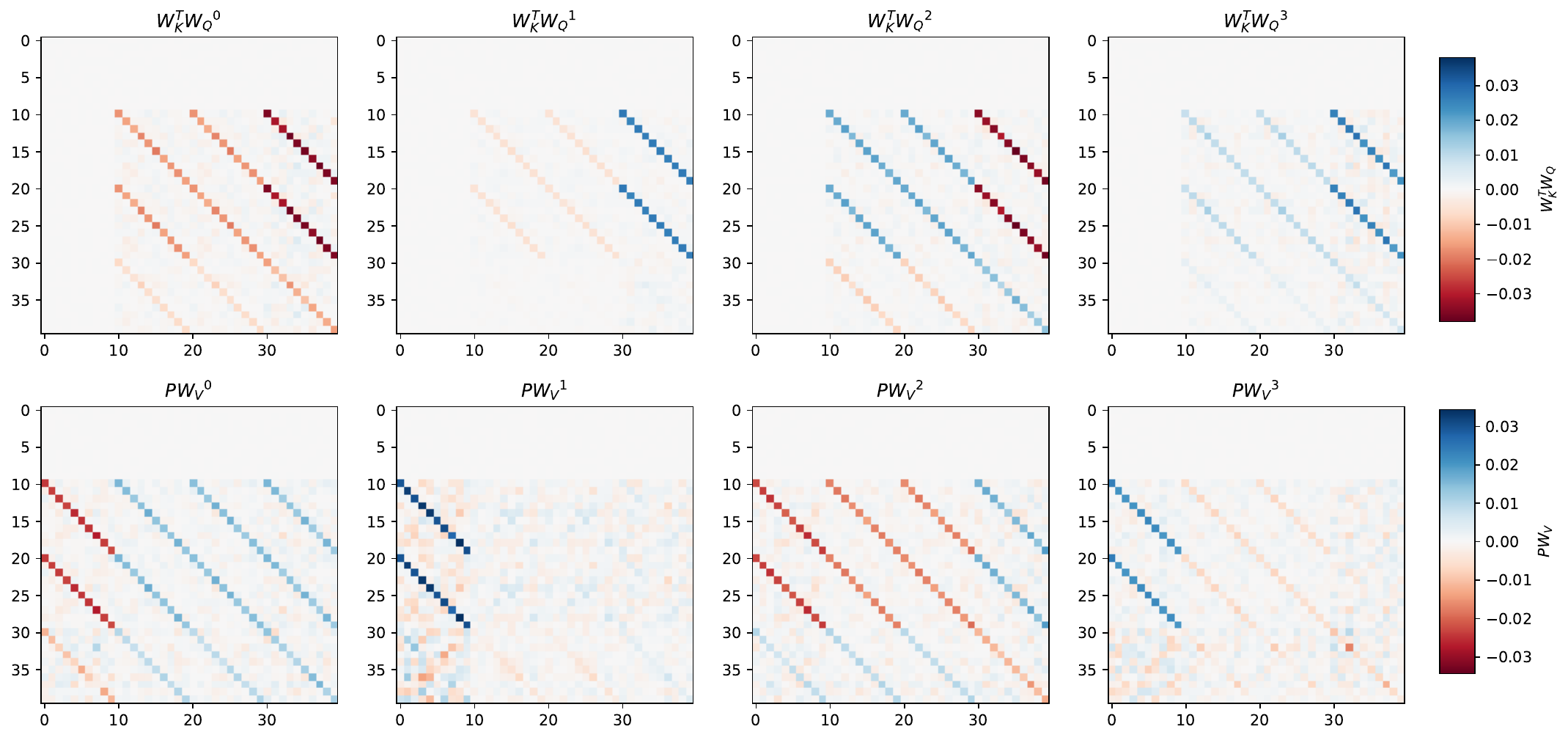}
    \includegraphics[width=0.8\textwidth]{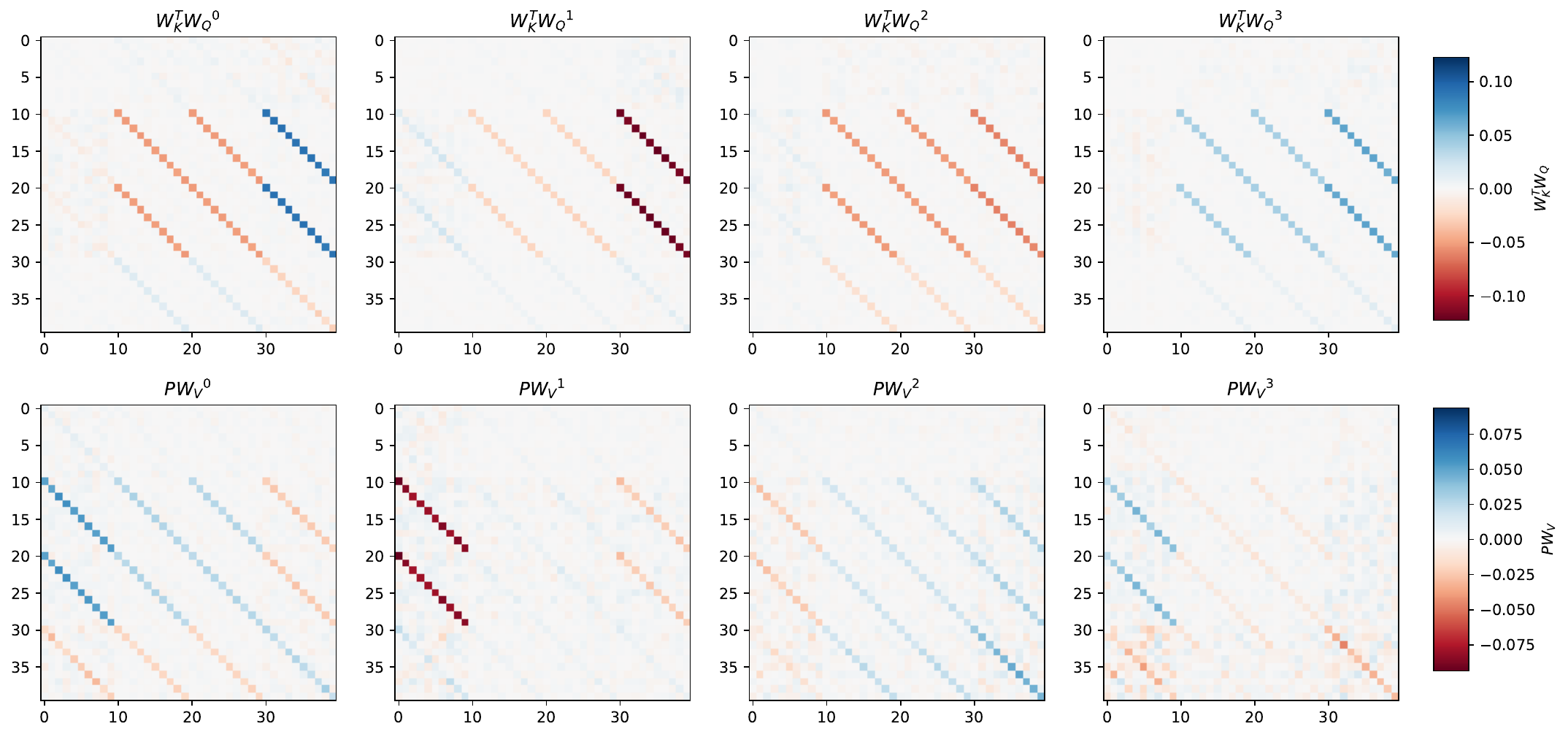}
    \includegraphics[width=0.8\textwidth]{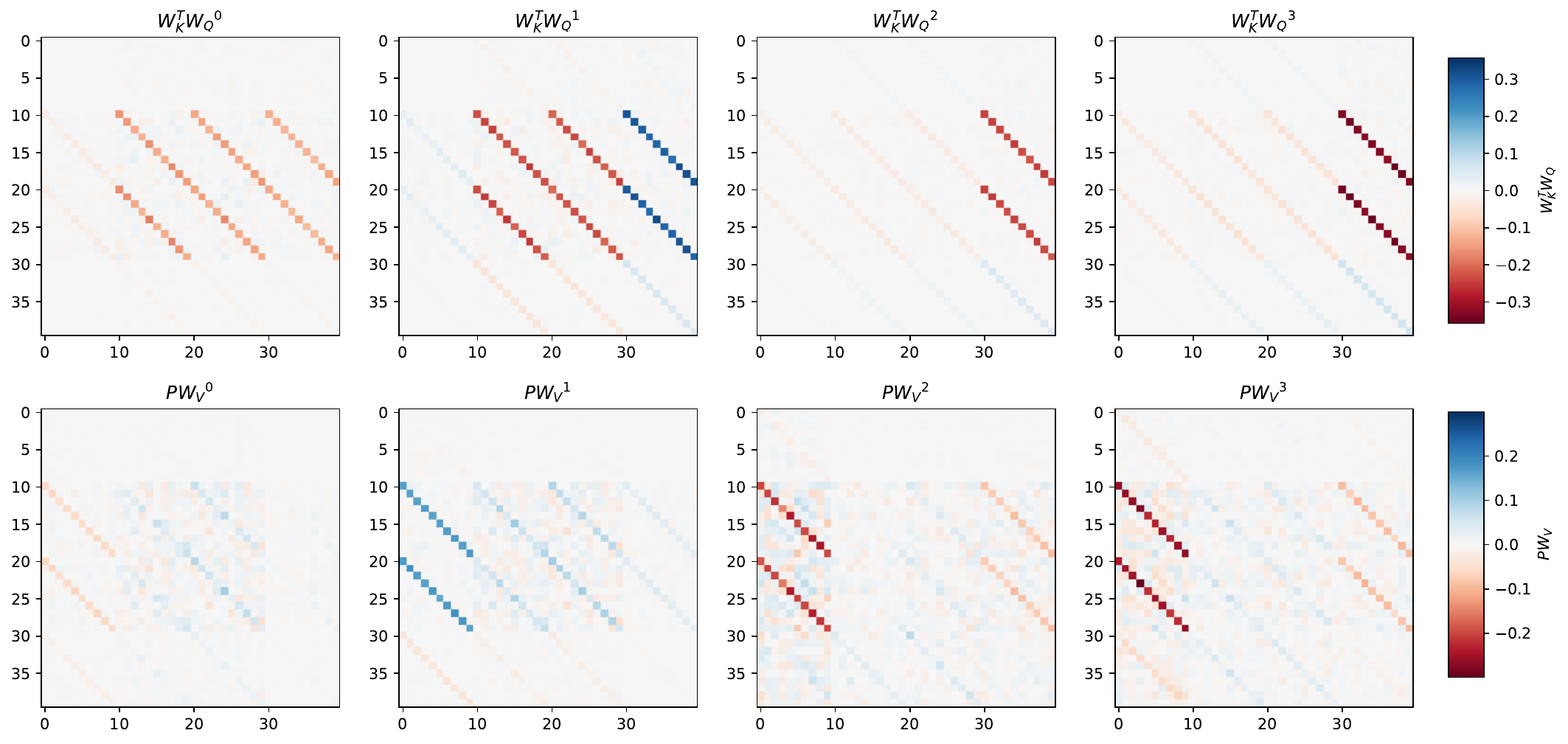}
    \caption{\textbf{Weights of the deep 6-layer linear Transformer trained on constructed tokens $e_t=(0,s_t,s_t,s_{t-1})$.} We observe clear structure in the trained Transformer weight products $W_K^\top W_Q$ as well as $PW_V$ in all 4 heads. Note that this structure seems to be sufficient to approximate $(S_{t-1}S_{t-1}^\top + 1/\lambda I)^{-1}s_t$, cf. probing experiments and weight construction in the main text. We show here all 4 heads (f.l.t.r.) of the first (top 2 rows), the second (next 2 rows), and the fourth (last 2 rows) linear layer.}
    \label{fig:weight-deep-construction-linear}
\end{figure}

\begin{figure}[htbp!]
    \centering
    \includegraphics[width=0.8\textwidth]{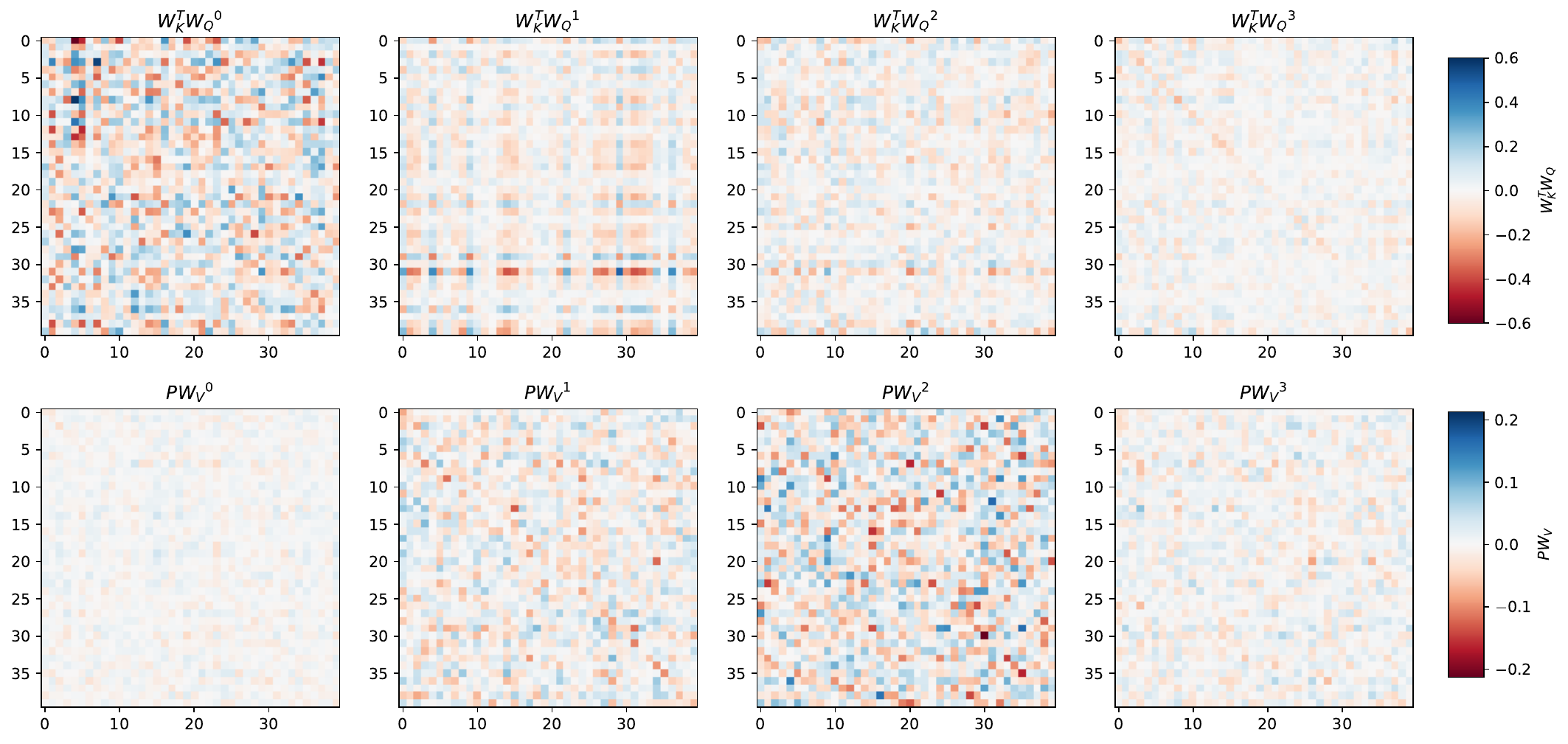}
    \includegraphics[width=0.8\textwidth]{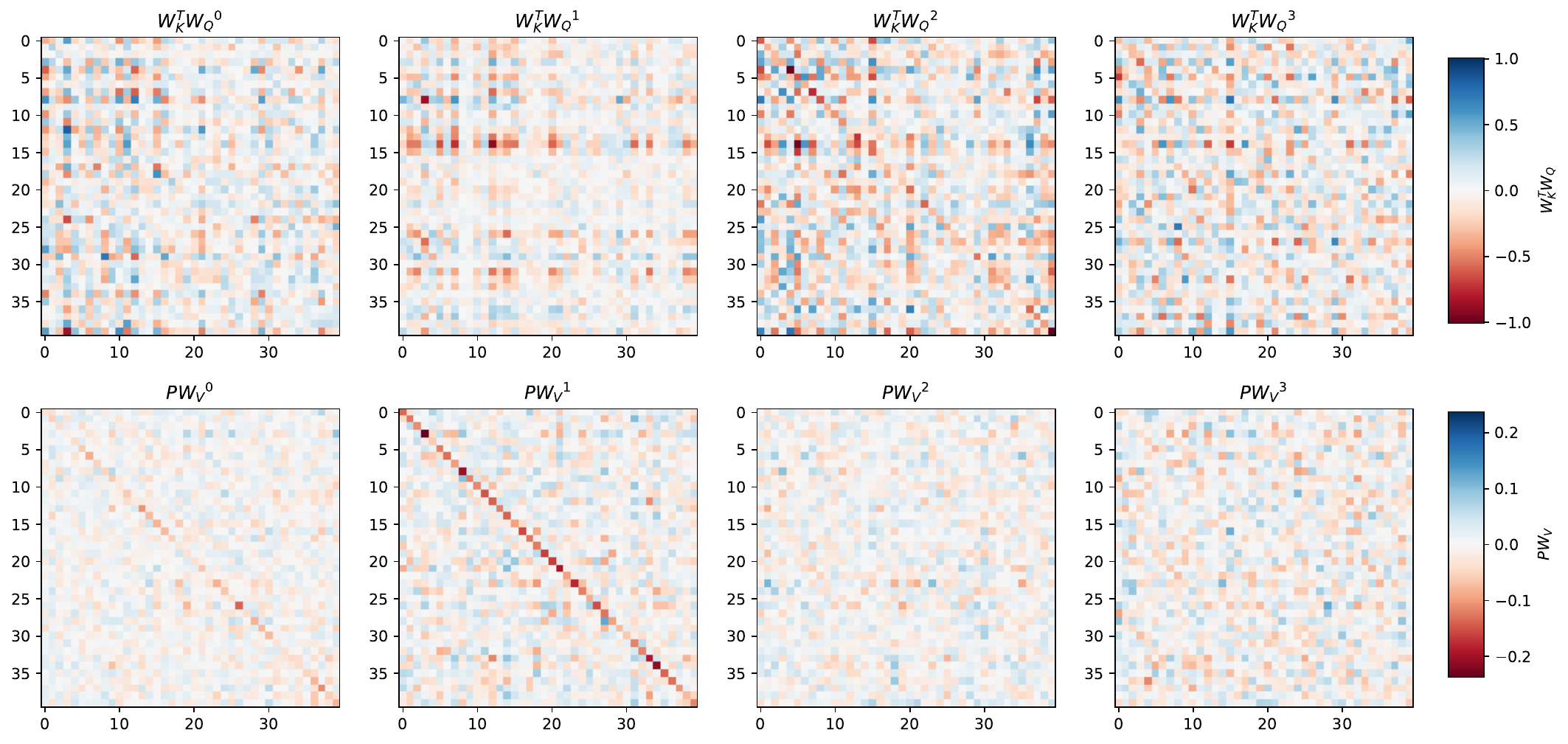}
    \includegraphics[width=0.8\textwidth]{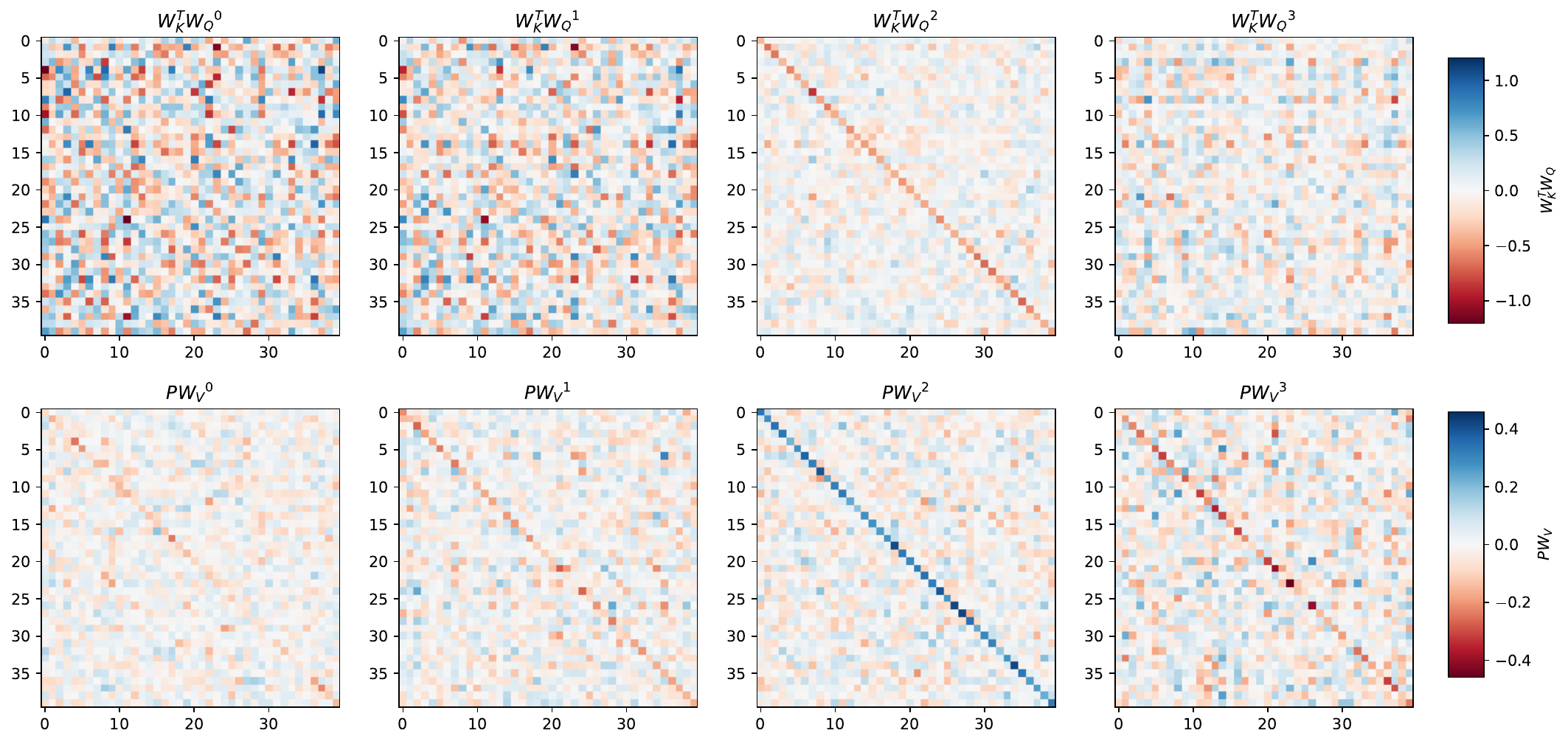}
    \caption{\textbf{Weight products of the deep 7-layer softmax Transformers trained on unconstructed tokens $e_t=s_t$.} We observe diagonal structure in the trained Transformer weight products $W_K^\top W_Q$ as well as $PW_V$. Note that this structure seems to be sufficient to approximate layer-wise the final prediction $s_{t+1}$ as well as $(S_{t-1}S_{t-1}^\top + 1/\lambda I)^{-1}s_t$, cf. probing experiments and weight construction in the main text. We show here all 4 heads (f.l.t.r.) of the first (top 2 rows) the second (middle 2 rows) and the fourth (last 2 rows) layers after the first (potential) copying-softmax-layer.}
    \label{fig:weight-deep-hybrid-softmax}
\end{figure}

\newpage
\section{Additional details on the mesa-layer} \label{sec:mesa-forget}
In this section, we provide a detailed derivation of the forward and backward (reverse-mode differentiation) pass of the mesa-layer. For completeness, we consider a generalized version of the mesa-layer, which includes an additional forget factor $\Gamma_{h,t}=(\gamma_{h,t'})_{t'=1}^{t}$, where $\gamma_{h,t'} \in (0,1]$, inspired by the recursive least-squares with forget factor algorithm \cite{johnstone_exponential_1982}. Given again a set of tokens $E_t$, the generalized mesa-layer changes the tokens as follows:
\begin{align}
\Delta e_{t}^\text{mesa} &= \sum_{h=1}^H P_h \hat{\Phi}_{h,t}^\text{mesa} q_{h, t},\\
\text{with} \qquad
     \hat{\Phi}_{h,t}^\text{mesa} = \argmin_\Phi &\left\{ \frac{1}{2}\sum_{t^\prime=1}^t \left(\prod_{t''=t'+1}^{t}\gamma_{h,t’'}\right) ||\Phi k_{h,t^\prime} - v_{h,t^\prime}||^2 + \frac{\prod_{t''=1}^t\gamma_{h,t''} }{2\lambda_h}||\Phi||_\mathrm{F}^2 \right\}.
\end{align}
For notational simplicity we drop the subscript in $h$ and ignore the sum over the heads in the following derivation. It can be shown that the analytical solution of the optimization problem is 
\begin{equation}
 \hat{\Phi}^\mathrm{mesa}_t = \left ( \sum_{t^\prime=1}^t \left(\prod_{t''=t'+1}^t\gamma_{t’'}\right) v_{t’} k_{t’}^\top \right )\left ( \sum_{t^\prime=1}^t \left(\prod_{t''=t'+1}^t\gamma_{t’'}\right) k_{t’} k_{t’}^\top + \frac{\prod_{t''=1}^t\gamma_{t''} }{\lambda}I \right )^{-1} 
\end{equation}
We will now see how $\Delta e_{t}^\text{mesa}$ can be efficiently computed in a forward pass.

\subsection{Computing the inverse term within $\hat{\Phi}^\mathrm{mesa}_t$}

Computing the full-fledged inverse at every timestep is computationally too expensive. We resort to using the Sherman-Morrison formula to efficiently compute the inverse term for all timestep sequentially in time. We redefine
\begin{equation}
    R_{t} = \left ( \sum_{t^\prime=1}^t \left(\prod_{t''=t'+1}^t\gamma_{t’'}\right) k_{t’} k_{t’}^\top + \frac{\prod_{t''=1}^t\gamma_{t''} }{\lambda}I \right )^{-1} \!.
\end{equation}
It satisfies the recursive formula
\begin{equation}
R_{t+1} = \left (\gamma_t R_t^{-1} + k_{t+1}k_{t+1}^\top \right)^{-1}
\end{equation}
with $ R_0 = \lambda I$, and the Sherman-Morrison formula thus gives
 \begin{align}
    R_{t+1} &= \gamma_{t+1}^{-1}\left( R_t^{-1} + \gamma_{t+1}^{-1} k_{t+1}k_{t+1}^\top \right )^{-1}\\
    & = \gamma_{t+1}^{-1} \left ( R_t - \frac{ \gamma_{t+1}^{-1} R_t k_{t+1}k_{t+1}^\top R_t}{1+ \gamma_{t+1}^{-1} k_{t+1}^\top R_t k_{t+1}} \right )\\ 
    &= \gamma_{t+1}^{-1} \left ( R_t - \frac{ R_t k_{t+1}k_{t+1}^\top R_t}{\gamma_{t+1}+ k_{t+1}^\top R_t k_{t+1}}\right ).
\end{align}

\subsection{Computing $\Delta e_{t}^\mathrm{mesa}$}

Given $R_{h,t}$ for all heads, we can rewrite the token update as
\begin{align}
\Delta e_{t}^\text{mesa} &= \sum_{h=1}^H P_h \left ( \sum_{t^\prime=1}^t \left(\prod_{t''=t'+1}^t\gamma_{h,t’'}\right) v_{h,t’} k_{h,t’}^\top \right ) R_{h,t} q_{h,t}\\
&=\sum_{h=1}^H P_h V_h \left (\left(\mathbbm{1}_{t' \leq t}\prod_{t''=t'+1}^t\gamma_{h,t’'}\right)_{t'=1}^\top \odot K_{h}^\top \tilde{q}_{h,t} \right )\\
&=\sum_{h=1}^H P_h V_h \left ( M_{:, t} \odot K_{h}^\top \tilde{q}_{h,t} \right ) \label{eq:masked_delta_e}
\end{align}

where $\tilde{q}_{h,t} = R_{h,t} q_{h,t}$ and $M_{t',t} := \mathbbm{1}_{t' \leq t}\prod_{t''=t'+1}^t\gamma_{h,t’'}$. Note that we apply some form causal masking here: we take the key $K_h \in \mathbb{R}^{D_a\times T}$ and value matrices $V_h \in \mathrm{R}^{D_a\times T}$ with all the sequence timesteps and select the entries occurring before time $t$. The main difference with the usual causal mask $(\mathbbm{1}_{t' \leq t})_{t', t}$ is the inclusion of the forget factors. It can be efficiently computed leveraging partial products. We conclude by remarking that the same mask can be applied to softmax attention layers, applying it to the key-queries products before the softmax.

\section{Mesa-layer differentiation}

\subsection{Mesa-layer backward pass computation via Sherman-Morrison}
\label{apx:mesa-backward}

We now detail how to compute the backward pass of the mesa-layer. Summarizing the results above, its forward pass is computed recursively following:
\begin{align}
  R_{h,t+1} &= \gamma_{h,t+1}^{-1} \left(R_{h,t} - \frac{R_{h,t}k_{h,t+1}k_{h,t+1}^\top R_{h,t}}{\gamma_{h,t+1} + k_{h,t+1}^\top R_{h,t}k_{h,t+1}}\right)\\
    \Delta e_{t, \text{mesa}} &= \sum_{h=1}^H P_h V_h \left ( M_{:, t} \odot K_{h}^\top \tilde{q}_{h,t} \right )
\end{align}
with $R_{h,0} = \lambda_h I$. These computations can be decomposed into 3 steps:
\begin{enumerate}
    \item First, the matrices $R_{t,h}$ are computed sequentially.
    \item Then, for all $t$ and $h$, the transformed queries $\tilde{q}_{h,t} = R_{h,t} q_{h, t}$ are computed.
    \item Finally, using the transformed queries $\tilde{Q}_{h}=(\tilde{q}_{h,t})_t$ as the queries, a standard cross-attention operation is computed from $(V_h,K_h,\tilde{Q}_h)$ using the causal mask $M$ that includes forgetting rates.
\end{enumerate}

While the backward pass of steps 2 and 3 can be computed easily with automatic differentiation tools without much overhead compared to standard attention layers, the same thing cannot be said about 1. We will here discuss how the backward pass of the computation of $\tilde{Q}_{h}$ can be computed in a memory-efficient way. Without loss of generality, we drop the subscript $h$ for notational simplicity.

\paragraph{The issue with out-of-the-box automatic differentiation.}  For all time steps $t$, $\tilde{q}_{t}=R_{t} q_{t}$ depends on $q_{t}$, but also $K_t$, $\Gamma_t$ and $\lambda$ through the variable $R_t$. 

In the backward pass, we are given as input the gradient of the loss function w.r.t.~$\tilde{Q}$, namely $\frac{\mathrm{d} \mathcal{L}}{\mathrm{d} \tilde{q}_t}$ for all $t$. The goal is then to compute the gradient of the loss w.r.t.~the input of $\tilde{Q}$, namely $\frac{\mathrm{d} \mathcal{L}}{\mathrm{d} k_t}$,$\frac{\mathrm{d} \mathcal{L}}{\mathrm{d} \gamma_t}$, $\frac{\mathrm{d} \mathcal{L}}{\mathrm{d} q_t}$ and $\frac{\mathrm{d} \mathcal{L}}{\mathrm{d} \lambda}$, which can be achieved via the chain rule.

While using automatic differentiation out of the box would take care of this computation, it would require in particular the storing of all intermediate variables $R_t$, which can be prohibitively expensive. 

\paragraph{Memory efficient custom backward pass.} Instead, we will show that storing the matrices $K, \Gamma, Q$ as well as $R_T$ where $T$ is the last time step of the training sequence, is sufficient to exactly compute the backward pass. Indeed, given the aforementioned inputs, all $R_t$ can be recomputed in linear complexity w.r.t.~$T$, which means we can reconstruct recursively the inputs of $\tilde{q}_t$  at all time steps. 

By noticing that $R_{t-1}=\gamma_t(R_{t}^{-1} - k_{t}k_{t}^\top)^{-1}$, we can apply the Sherman-Morrison formula backwards to obtain $R_{t-1}$ as
\begin{align}
  R_{t-1} &=\gamma_t \left ( R_{t} - \frac{R_{t}(-k_{t})k_{t}^\top R_{t}}{1 + (-k_{t})^\top R_{t}k_{t}} \right ) \\
  &=\gamma_t \left (R_{t} - \frac{R_{t}k_{t}k_{t}^\top R_{t}}{ k_{t}^\top R_{t}k_{t}-1} \right )
\end{align}

We will now show how accumulating the right error signal and leveraging the vector-jacobian product trick together with automatic differentiation tools is sufficient for computing the full backward pass recursively.

Firstly, given the error signal and reconstructed $R_{t}$ allows the computation of $\frac{\mathrm{d} \mathcal{L}}{\mathrm{d} q_t}$ via
\begin{equation}
\frac{\mathrm{d} \mathcal{L}}{\mathrm{d} q_t} =\frac{\mathrm{d} \mathcal{L}}{\mathrm{d} \tilde{q}_{t}}\frac{\mathrm{d} \tilde{q}_{t}}{\mathrm{d} q_t}=\frac{\mathrm{d} \mathcal{L}}{\mathrm{d} \tilde{q}_{t}}S_{t}
\end{equation}

Secondly, we rewrite $\tilde{q}_{t}$ as a function of $k_t, \gamma_t, R_{t-1}$ and $q_t$, i.e. 
\begin{equation}
\tilde{q}_{t} = \mathcal{R}^{\text{forward}} (R_{t-1}, k_{t}, \gamma_t) q_t
\end{equation}
Since $\mathcal{L}$ depends on $k_t$ only via both $\tilde{q}_{t} $ and $R_t$, we can then rewrite 
\begin{align}
\frac{\mathrm{d} \mathcal{L}}{\mathrm{d} k_t} &= \frac{\mathrm{d} \mathcal{L}}{\mathrm{d} \tilde{q}_{t}}\frac{\mathrm{d} \tilde{q}_{t}}{\mathrm{d} k_t} + \frac{\mathrm{d} \mathcal{L}}{\mathrm{d} {R}_{t}}\frac{\mathrm{d} {R}_{t}}{\mathrm{d} k_t}\\
&= \frac{\mathrm{d} \mathcal{L}}{\mathrm{d} \tilde{q}_{t}}\frac{\partial \tilde{q}_{t}}{\partial k_t} + \frac{\mathrm{d} \mathcal{L}}{\mathrm{d} {R}_{t}}\frac{\partial {R}_{t}}{\partial k_t}
\end{align}
where, provided $R_{t-1}, k_t, \gamma_t$ and $q_t$, $\frac{\partial \tilde{q}_{t}}{\partial k_t}$ can be computed easily using e.g.~automatic differentiation tools. Similarly, we have, 
\begin{align}
\frac{\mathrm{d} \mathcal{L}}{\mathrm{d} \gamma_t} &= \frac{\mathrm{d} \mathcal{L}}{\mathrm{d} \tilde{q}_{t}}\frac{\partial \tilde{q}_{t}}{\partial \gamma_t} + \frac{\mathrm{d} \mathcal{L}}{\mathrm{d} {R}_{t}}\frac{\partial {R}_{t}}{\partial \gamma_t}
\end{align}
Notice that $\frac{\mathrm{d} \mathcal{L}}{\mathrm{d} {R}_{t}}$ can be computed recursively following the chain rule
\begin{align}
    \frac{\mathrm{d} \mathcal{L}}{\mathrm{d} {R}_{t-1}} &= \frac{\mathrm{d} \mathcal{L}}{\mathrm{d} {R}_{t}}\frac{\partial R_{t}}{\partial R_{t-1}} + \frac{\mathrm{d} \mathcal{L}}{\mathrm{d} \tilde{q}_{t}} \frac{\partial \tilde{q}_{t}}{\partial R_{t-1}}
\end{align}
where again, provided $R_{t-1}, k_t, \gamma_t$ and $q_t$, both terms can be computed efficiently with standard automatic differentiation tools coupled with the well known vector-Jacobian product trick given the quantities $\frac{\mathrm{d} \mathcal{L}}{\mathrm{d} {R}_{t}}$ and $\frac{\mathrm{d} \mathcal{L}}{\mathrm{d} \tilde{q}_{t}}$.

Thirdly, we can show that 
\begin{equation}
\frac{\mathrm{d} \mathcal{L}}{\mathrm{d} \lambda} = \mathrm{Tr}\left [ \frac{\mathrm{d} \mathcal{L}}{\mathrm{d} R_0} \right ]
\end{equation}

Combining everything, we can now implement the backward computation recursively via the following equations:
\begin{align}
   R_{t-1} & = \gamma_t \left (R_{t} - \frac{R_{t}k_{t}k_{t}^\top R_{t}}{ k_{t}^\top R_{t}k_{t}-1} \right ) \\
    \frac{\mathrm{d} \mathcal{L}}{\mathrm{d} R_{t-1}} &= \frac{\mathrm{d} \mathcal{L}}{\mathrm{d} R_{t}} \frac{\partial R_{t}}{\partial R_{t-1}} + \frac{\partial \mathcal{L}}{\partial \tilde{q}_{t}} \frac{\partial \tilde{q}_{t}}{\partial R_{t-1}} \\
    \frac{\mathrm{d} \mathcal{L}}{\mathrm{d} k_t} &=\frac{\mathrm{d} \mathcal{L}}{\mathrm{d} \tilde{q}_{t}}\frac{\partial \tilde{q}_{t}}{\partial k_t} + \frac{\mathrm{d} \mathcal{L}}{\mathrm{d} R_{t}} \frac{\partial {R}_{t}}{\partial k_t}\\
    \frac{\mathrm{d} \mathcal{L}}{\mathrm{d} \gamma_t} &=\frac{\mathrm{d} \mathcal{L}}{\mathrm{d} \tilde{q}_{t}}\frac{\partial \tilde{q}_{t}}{\partial \gamma_t} + \frac{\mathrm{d} \mathcal{L}}{\mathrm{d} R_{t}} \frac{\partial {R}_{t}}{\partial \gamma_t}\\
    \frac{\mathrm{d} \mathcal{L}}{\mathrm{d} q_t} &=\frac{\mathrm{d} \mathcal{L}}{\mathrm{d} \tilde{q}_{t}}R_{t}\\
    \frac{\mathrm{d}\mathcal{L}}{\mathrm{d}\lambda} &= \mathrm{Tr}\left [ \frac{\mathrm{d} \mathcal{L}}{\mathrm{d} R_0}\right ]
\end{align}
$R_T$ is assumed to be given and $\frac{\mathrm{d} \mathcal{L}}{\mathrm{d}R_T}=0$. The above equations only require the storage of $\frac{\mathrm{d} \mathcal{L}}{\mathrm{d} R_{t}}, \frac{\mathrm{d} \mathcal{L}}{\mathrm{d} R_{t-1}}, R_t, R_{t-1}$ at all time, and computes the backward pass in a similar time and memory complexity as for the forward pass. The derivation is identical without forgetting factors, by setting all $\gamma$ to $1$.

\textbf{Comment on runtime.}  We highlight that, although this implementation of the mesa-layer reduces the memory footprint of the forward and backward pass substantially, the layer still runs forward (and backward) in time. This prevents the computation of all mesa-layer outputs in parallelization during training, a crucial advantage of softmax as well as linear attention. On the other hand, during test time, the mesa-layer benefits from the same advantages of linear self-attention or RNNs and predicts the next token without the necessity to store and attend to the past. In the next sections, we present two potential avenue to improve the training time by a solution based in linear solvers or by a solution approximating the necessary inversions by a Neumann series running in parallel.

\subsection{Alternative derivation through the implicit function theorem}

We here present an alternative way of deriving the gradients presented above that leverages the implicit function theorem. The key here is to remark that $\hat{\Phi}_{t}^\mathrm{mesa}$ satisfies that the gradient of the least-square regression loss $L$ is 0. For simplicity, we restrict ourselves to the case in which the output dimension of $\hat{\Phi}_{t}^\mathrm{mesa}$ is one, that is $\hat{\Phi}_{t}^\mathrm{mesa}=\hat{\phi}_{t}^{\top}$ for $\hat{\phi}_{t}$ some column vector, and remark that we have to repeat the same operation over all rows of $\hat{\Phi}_{t}^\mathrm{mesa}$ to obtain the full gradient, as all output coordinates are independent in the least-square regression problem. Therefore, we $w$ defined through the implicit function
\begin{equation}
    \frac{\mathrm{d} L}{\mathrm{d} \phi}(\hat{\phi}_{t}) = \sum_{t'=1}^t M_{t',t} (\hat{\phi}_{t}^\top k_{t'} - v_{t'}) k_{t'}^\top + \frac{M_{1, t}}{\lambda} \hat{\phi}_{t}^\top = 0.
\end{equation}
We can then use the implicit function theorem and compute the derivative of $w$ with respect to any quantity $\cdot$ through
\begin{align}
    \frac{\mathrm{d} \hat{\phi}_t}{\mathrm{d}~\cdot} &= - \left ( \frac{\mathrm{d}^2 L_t}{\mathrm{d} \phi^2}(\phi_t) \right )^{\!\!-1} \frac{\mathrm{d}^2 L_t(\hat{\phi}_t)}{\mathrm{d} \cdot \mathrm{d} \phi}\\
    &= - R_t \frac{\mathrm{d}^2 L_t(\hat{\phi}_t)}{\mathrm{d} \cdot \mathrm{d} \phi}.
\end{align}
For example, this yields
\begin{equation}
    \frac{\mathrm{d} \hat{\phi}_t}{\mathrm{d}v_{t'}} = M_{t',t}R_t k_{t'}.
\end{equation}
Finally, we can recover the desired gradient by combining the previous equation with the chain rule.

\subsection{Parallel backward pass through Neumann series approximation}
\label{apx:mesa-approx}

Although the previous custom backward gradient computation allows for dramatic memory savings during training, the underlying recursive least squares computation still suffers from linear scaling in time, similar to recurrent neural networks, as we cannot parallelize computation across time dimension. 

Here, we discuss an alternative forward pass that can be used when one can afford storing all intermediate matrices $R_{h,t}$ in time. This forward pass leverages a $K$-step truncated Neumann series to approximate the inverses in parallel, and is compatible with automatic differentiation tools out of the box. Interestingly, we can do this by simply repeating (with the same weights) a slightly altered linear self-attention layer $K$ times.   

Our goal is now to efficiently compute the terms $\tilde{q}_{t} := R_t q_{t}= (K_t K_t^\top + \frac{1}{\lambda} I)^{-1}q_{t}$ for all time steps in parallel. Indeed, once give these vectors, one can leverage Equation~\ref{eq:masked_delta_e} and efficient dot-product attention (DPA) layers implementations\footnote{See \url{https://flax.readthedocs.io/en/latest/_modules/flax/linen/attention.html} for an implementation of DPA in JAX \citep{bradbury_jax_2018}.}. Note that we here ignore the forgetting factors, but their partial products can easily be integrated in one of the $K_t$ in $K_t K_t^\top$ to recover the version with forget rates described above.

Given an invertible matrix $X$ with operator norm less than $1$, the truncated Neumann series approximates its inverse by 
\begin{equation}
    X^{-1} \approx \tilde{X}^{-1}_{(K)} := \sum_{k=0}^K (I-X)^k.
\end{equation}
When multiplying a vector from the right, we see that 
\begin{align}
    \tilde{x}^{(K)} := \tilde{X}^{-1}_{(K)} x & = \sum_{k=0}^K (I-X)^kx \\
    & = \sum_{k=1}^{K} (I-X)^kx + x\\
    & = (I-X)\sum_{k=0}^{K-1} (I-X)^kx + x\\
    & = (I-X)\tilde{x}^{(K-1)} + x
\end{align}
An advantage of the truncated Neumann series compared to other approximate inverse techniques such as Newton-Iteration is that we can compute more series elements without passing intermediate matrices across algorithmic steps -- which in turn makes it memory efficient and straightforward to use in the light of automatic differentiation. We only need to keep the original matrix we wish to invert in memory at all times and store the intermediate vectors $\tilde{x}^{(k)}$ for the backward pass.

We now look at the quantities we wish to compute, that is $\tilde{q}_{t} = (K_t K_t^\top + \frac{1}{\lambda} I)^{-1}q_{t}$, and approximate it by $\tilde{q}_t^{(K)}$, obtained by multiplying $q_{t}$ to the $K$-step truncated Neumann series approximating the inverse term $(K_t K_t^\top + \frac{1}{\lambda} I)^{-1}$. Note that a normalization by the operator norm of the matrix inside the inverse is necessary for the approximation to hold.

Then, $\tilde{q}_t^{(K)}$ can be computed recursively as 
\begin{align}
    \tilde{q}_t^{(k+1)} & = \left(I- \left (K_t K_t^\top + \frac{1}{\lambda} I \right )\right)\tilde{q}_t^{(k)} + q_t\\
    &=q_t + \left (1-\frac{1}{\lambda} \right )\tilde{q}^{(k)}_t -K_t K_t^\top\tilde{q}_t^{(k)}
\end{align}
and thus by denoting $\tilde{Q}_{t}^{(k)} := (\tilde{q}_{t'}^{(k)})_{t'=1}^t$, we have 
\begin{align} \label{eq:neumann-iter} 
    \tilde{Q}_{k+1}^{(k+1)} &=Q_t + \left(1-\frac{1}{\lambda}\right)\tilde{Q}_{t}^{(k)} -K_t K_t^\top\tilde{Q}_{t}^{(k)}
\end{align}
which is the sum of simple terms with a DPA computed between $K_t,K_t,\tilde{Q}_{t}^{(k)}$.

After obtaining $\tilde{Q}_{t}^{(K)}$ to approximate $\tilde{Q}_{t}$, we compute the approximate least-squares solution as described above. Note that other implementations could save us from effectively recomputing $(K_{t}K_{t}^\top)$ at every iteration of Equation \ref{eq:neumann-iter} by simply pre-computing these terms before running the Neumann approximation. We nevertheless observe the former version to be faster when timing for forward and backward computation and speculate the reason being the highly optimized implementation of DPA as the backbone of the self-attention layer.
Note that a simple byproduct of the derivations here is the insight that chaining linear self-attention layers can actually easily implement truncated Neumann series computation -- especially if the goal is an inverse multiplied by a known vector. See materials and methods section of the main text for an in-depth analysis.

%\newpage

\newpage

\section{Probabilistic latent-state inference in Transformers}
In this section, we generalize our results on latent-state inference in partially-observed deterministic linear systems towards noisy linear systems. Our aim is to show that the optimal maximum-likelihood estimator (MLE) of the next observation $s_{t+1}$ is a linear map of the concatenated previous observations, possibly encoded into a lower-dimensional subspace by a linear encoder. First, we show that in the Gaussian noise setting, the MLE of $s_{t+1}$ is a linear map of the MLE of the latent state $h_{t+1}$. Second, we show that the MLE of the latent state $h_{t+1}$ is a linear map of a concatenation of the previous $k$ observations. Finally, we generalize our setting to allow for a linear encoding of all the previous observations into a fixed low-dimensional subspace, instead of explicitly concatenating $k$ observations. Taken together, these results show that performing least-squares linear regression on tokens that encode or concatenate previous observations is an optimal strategy for predicting the next observation according to the maximum-likelihood estimator.

\subsection{The MLE of $s_{t+1}$ is a linear map of the MLE of $h_{t+1}$}
As we consider linear dynamics with additive Gaussian noise, the distributions $p(s_{t+1} \mid z_t^k)$ and $p(s_{t+1}, h_{t+1} \mid z_t^k)$ are multivariate Gaussians. Let us now consider the MLE estimators of the marginal $p(s_{t+1} \mid z_t^k)$ and joint distribution $p(s_{t+1}, h_{t+1} \mid z_t^k)$. 
\begin{align*}
    \hat{s}^{\text{marginal}}_{t+1} &= \argmax_{s_{t+1}} p(s_{t+1} \mid z_t^k) \\
    \hat{s}^{\text{joint}}_{t+1}, \hat{h}^{\text{joint}}_{t+1} &= \argmax_{s_{t+1}, h_{t+1}} p(s_{t+1} \mid z_t^k)p(h_{t+1} \mid s_{t+1}, z_t^k)
\end{align*}
$p(h_{t+1} \mid s_{t+1}, z_t^k)$ is Gaussian, as conditional distributions of jointly distributed Gaussian variables are also Gaussian. Furthermore, the covariance of a Gaussian conditional distribution only depends on the covariance of the joint distribution, not on the specific value of the conditioned variable $s_{t+1}$. Hence, the maximum (not the $\argmax$) of $p(h_{t+1} \mid s_{t+1}, z_t^k)$ is independent from $s_{t+1}$, and we hence have that the MLE $\hat{s}^{\text{marginal}}_{t+1}$ is equal to $\hat{s}^{\text{joint}}_{t+1}$. Rewriting the joint distribution as $p(h_{t+1} \mid z_t^k)p(s_{t+1} \mid h_{t+1}, z_t^k)$, and repeating the same arguments, we have that 
\begin{align*}
    \hat{s}^{\text{marginal}}_{t+1} = C^* \hat{h}^{\text{joint}}_{t+1} = C^* \hat{h}^{\text{marginal}}_{t+1}
\end{align*}
with $\hat{h}^{\text{marginal}}_{t+1}$ the MLE of $p(h_{t+1} \mid z_t^k)$. Hence, the MLE of $s_{t+1}$ is a linear map of the MLE of the latent state $h_{t+1}$. 

\subsection{The MLE of $h_{t+1}$ is a linear map of $z_t^k$}
Now we turn our focus on showing that $\hat{h}_{t+1}^{\text{MLE}} = \argmax_{h_{t+1}} p(h_{t+1} \mid z_t^k)$ as a linear map of $z_t^k$. First, by similar arguments as before, we have that $\hat{h}_{t+1}^{\text{MLE}} = A\hat{h}_{t}^{\text{MLE}}$, with $\hat{h}_{t}^{\text{MLE}} = \argmax p(h_t \mid z_t^k)$. In the following, we show that $\hat{h}_{t}^{\text{MLE}}$ is a linear map of $z_t^k$, thereby completing our goal of this section. 

Running the noisy dynamics backwards gives us $h_{t-1}= W^{*-1}(h_t - \epsilon_{h,t-1})$. Repeating this $k$ times gives us
\begin{align}
    z_t^k &= v_t^k +  \begin{bmatrix}
        C^* {W^*}^{-(k-1)} \\
        \vdots \\
        C^* W^{*-1} \\
        C^*
    \end{bmatrix} h_t - \begin{bmatrix}
        C^* W^{*-(k-1)} \\
        \vdots \\
        C^*W^{*-1} \\
        0
    \end{bmatrix} \epsilon_{h,t-1} - \begin{bmatrix}
        C^*W^{*-(k-2)} \\
        \vdots \\
        C^*W^{*-1} \\
        0 \\
        0
    \end{bmatrix} \epsilon_{h,t-2} - \hdots \\
    &= v_t^k + \mathcal{F}_k h_t - \mathcal{F}_k^1 \epsilon_{h,t-1} - \mathcal{F}_k^2 \epsilon_{h,t-2} - \hdots \\
    &= v_t^k + \mathcal{F}_k h_t - \sum_{l=1}^{k-1} \mathcal{F}_k^l \epsilon_{h,t-l}
\end{align}
with $v_t^k$  the concatenated observation noise variables $\epsilon_{s,t}$ of the last $k$ timesteps, and $\mathcal{F}_k^l$ shifted versions of the filter matrix $\mathcal{F}_k$ by inserting $l$ zero blockmatrices from below:
\begin{align*}
\mathcal{F}_k = \begin{bmatrix}
        C^* W^{*-(k-1)} \\
        \vdots \\
        C^* W^{*-1} \\
        C^*
    \end{bmatrix}
\end{align*}

Now we want to extract the maximum-likelihood estimate of $h_t$. We have that 
\begin{align}
   p(h_t, v_t^k, \epsilon_{h,t-(k-1):t-1} \mid s_{t-(k-1):t}) =  p(h_t \mid s_{t-(k-1):t}) p(v_t^k, \epsilon_{h,t-(k-1):t-1} \mid h_t, s_{t-(k-1):t})
\end{align}
Importantly, all variables are Gaussian, as we have linear dynamics and Gaussian noise. Due to the property of Gaussian conditional distribution conditioned before, 

the maximum of $p(v_t^k, \epsilon_{h,t-(k-1):t-1} \mid h_t, s_{t-(k-1):t})$ only depends on the covariance matrix of the distribution and hence does not depend on the value of $h_t$. Consequently, we have that the value of $h_t$ that maximizes $p(h_t, v_t^k, \epsilon_{h,t-(k-1):t-1} \mid s_{t-(k-1):t})$ is the same one that maximizes $p(h_t \mid s_{t-(k-1):t})$. This is convenient, as it is much more tractable to maximize the joint distribution w.r.t.~$h_t$ and the noise variables, compared to maximizing the marginal distribution w.r.t.~$h_t$, for which we need to compute integrals. 

As the noise variables are Gaussian (with covariances which we assume to be equal to $\sigma I$ for simplicity), maximizing the joint log-probability is equivalent to the following optimization problem: 
\begin{align}\label{eq:mle_objective_x}
    \argmin_{h_t, \epsilon_{h,t-1:t-k+1}, v_t^k} \frac{1}{2\sigma^2} \|v_t^k\|^2 + \frac{1}{2\sigma^2} \sum_{l=1}^{k-1} \|\epsilon_{h,t-1}\|^2 \quad ~~ \text{s.t.} ~~ z_{t}^k = v_t^k + \mathcal{F}_k h_t - \sum_{l=1}^{k-1} \mathcal{F}_k^l \epsilon_{h,t-l}.
\end{align}
We solve it with the Lagrange multiplier method: 
\begin{align}
    {\cal L} = \frac{1}{2\sigma^2} \|v_t^k\|^2 + \frac{1}{2\sigma^2} \sum_{l=1}^{k-1} \|\epsilon_{h,t-1}\|^2 + \lambda^\top \left(-z_{t}^k + v_t^k + \mathcal{F}_k h_t - \sum_{l=1}^{k-1} \mathcal{F}_k^l \epsilon_{h,t-l}\right)
\end{align}
Taking the gradients of this Lagrangian and equating them to zero gives us the following linear system with $kn_h + 2kn_s$ equations and $kn_h + 2kn_s$ variables:
\begin{align}
    \nabla_{h_t} {\cal L} &= \mathcal{F}_k^\top \lambda = 0\\
    \nabla_{\epsilon_{h,t-l}} {\cal L} &= \epsilon_{h,t-l} - \mathcal{F}_k^{l\top} \lambda = 0\\
    \nabla_{v_t^k} {\cal L} &= v_t^k + \lambda = 0 \\
    \nabla_{\lambda} {\cal L} &= \-z_{t}^k + v_t^k + \mathcal{F}_k h_t - \sum_{l=1}^{k-1} \mathcal{F}_k^l \epsilon_{h,t-l} = 0
\end{align}
We can structure this set of equations in a big matrix equation
\begin{align}\label{eq:linear_system_mle_x}
    S \begin{bmatrix}
        h_t \\ \epsilon_{h,t-1} \\ \vdots \\ \epsilon_{h,t-(k-1)} \\ v_t^k \\ \lambda
    \end{bmatrix} = \begin{bmatrix}
        0 \\ 0 \\ \vdots \\ 0 \\z_t^k
    \end{bmatrix}
\end{align}
Where $S$ contains the terms of the equations that multiply with the variables, and the right-hand-side of the above equation contains all other terms (only $z_t^k$ in our case). We can solve this system by inverting $S$ (assuming it is invertible). Now we can extract our maximum likelihood estimate of $h_t$ as 
\begin{align}
    \hat{h}_t = \begin{bmatrix} I & 0 & \hdots & 0 \end{bmatrix} S^{-1} \begin{bmatrix}
        0 \\ 0 \\ \vdots \\ 0 \\z_t^k
    \end{bmatrix} = \left[S^{-1}\right]_{0,k+2} z_t^k
\end{align}
with $\left[S^{-1}\right]_{0,k+2}$ the upper right block of $S^{-1}$. So after this slightly more complicated derivation, we again end up with a simple linear map from $z_t^k$ to decode the maximum likelihood hidden state. Let us rename it for ease of notation: $U = \left[S^{-1}\right]_{0,k+2}$: 
\begin{align}
    \hat{h}_t = Uz_t^k
\end{align}
Using this state estimation, we can predict the next observation as $\hat{s}_{t+1} = C^* W^* \hat{h}_t$. This leads us to the following optimal candidate for the linear map $z_{t+1}^k = \Phi z_t$:
\begin{align}
    \Phi = \begin{bmatrix}
        0 & I & 0 & \hdots & 0 \\
        \vdots & \ddots & \ddots & \ddots & \vdots \\
        0 & 0 & \hdots & 0 & I \\
        & & C^* W^* U & & 
    \end{bmatrix}
\end{align}
As there exists an optimal map between $z_t^k$ and $z_{t+1}^k$ that is linear, this map can be found by performing least-squares on an autoregressive dataset with $z_t^k$.

\subsection{Capacity constraints on the representation}
Previously, we derived results for a fixed $k$. Now, we consider the case with a capacity bottleneck on the representations of the transformer. Let us assume that the transformer can allocate a $d$-dimensional subspace to store some representation of the past observations $s_{0:t-1}$. Instead of concatenating $k$ previous observations into this subspace with the constraint that $k\leq d/n_s$ with $n_s$ the observation dimension, we can consider a more general case where we have an encoding $u_t = E s_{0:t} = E z_t^T$. Here, $E \in \mathbb{R}^{d \times Tn_s}$, with $T$ the sequence length. For $t<T$, we prepend zeros to $s_{0:t}$ to make the dimensions fit. When $E$ consists of identity matrices on the diagonals corresponding to the $k$ last observations, we recover the previous case. However, it might be more optimal to copy partial information from more than $k$ observations, resulting in a different encoding matrix $E$.

We are interested in three main points. First, we need to formalize a bottleneck objective that the encoding matrix $E$ should optimize. Second, we need to show that the MLE for $s_t$ is still a linear map of the encoded observations $u_t$. Finally, we need some algorithm or strategy to compute the optimal encoding matrix $E$, such that we can compare it to the learned weights of the transformer. 

\paragraph{Bottleneck objective.}
We want the encoding to capture as much useful information about past observations as possible, to predict the future observation. Hence, we want the MLE $\hat{y}_{t+1}$ conditioned on $u_t$ to be as close as possible to the MLE conditioned on the full past $y_{0:t}$. We can formalize this in the following bilevel optimization problem
\begin{align}
    \min_{E} ||\argmax_{s_{t+1}} p(s_{t+1} \mid s_{0:t}) - \argmax_{s_{t+1}} p(s_{t+1} \mid u_t)||^2
\end{align}
As both $p(s_{t+1} \mid s_{0:t})$ and $p(s_{t+1} \mid u_t)$ are Gaussian, we have that the MLE of $p(s_{t+1} \mid s_{0:t})$ and $p(s_{t+1}, h_{t+1} \mid s_{0:t})$ are the same (see previous section), and hence we can rewrite the bilevel optimization problem into an equivalent form: 
\begin{align}\label{eq:encoding_objective}
    \min_{E} ||C\left[\argmax_{h_{t+1}} p(h_{t+1} \mid s_{0:t}) - \argmax_{h_{t+1}} p(h_{t+1} \mid u_t) \right]||^2
\end{align}

\paragraph{MLE of $h_{t+1}$ is a linear map of $u_t$.}
For a fixed encoding $E$, it is easy to see that the MLE $\hat{h}_{t+1}$, and hence the MLE $\hat{s}_{t+1} = C\hat{h}_{t+1}$ as well, are a linear map of $u_t$. We have that $u_t = E z_t^T$. Hence, we can repeat the calculations of the previous section, now with a new linear constraint $u_t = E\left[ v_t^T + \mathcal{F}_T h_t - \sum_{l=1}^{T-1} \mathcal{F}_T^l \epsilon_{h,t-l}\right]$ for the MLE objective \eqref{eq:mle_objective_x}. The main result that the MLE $\hat{h}_{t+1}$ is a linear map of $u_t$ holds in this case as well, as all equations for the first-order optimality conditions remain linear. 

\paragraph{How to compute the optimal encoding?}
Now that we derived $\argmax_{h_{t+1}} p(h_{t+1} \mid u_t)$ as a function of $E$, we can use this to optimize the encoding objective \eqref{eq:encoding_objective} w.r.t.~$E$, by computing its gradients. Concretely, we need to iterate the following two steps:
\begin{enumerate}
    \item Compute the MLE $\hat{s}_t = C^* \hat{h}_t$ conditioned on $u_t$, by solving the linear system resulting from the MLE objective \eqref{eq:mle_objective_x} with the new constraint $u_t = E\left[ v_t^T + \mathcal{F}_T h_t - \sum_{l=1}^{T-1} \mathcal{F}_T^l \epsilon_{h,t-l}\right]$. Use a differentiable linear solver (e.g.~torch.linalg.solve), such that we can backpropagate through it in step 2. 
    \item Compute the encoding loss \eqref{eq:encoding_objective} and compute the gradients w.r.t.~$E$ on a training dataset consisting of multiple teacher systems. 
\end{enumerate}

\section{Additional experiments with different sequence generator distributions}

For our main text experiments, the groundtruth transition matrix $W^*$ was set to a random orthogonal matrix. Here we briefly analyze Transformers trained on systems with different transition matrix statistics. For all settings in this section, we assume full observability, that is $s_t = h_t$ for all time steps $t$.

\subsection{Contracting linear dynamics}

We show here the preliminary result when diverging from purely orthogonal teachers $W$ to construct the sequence presented to the Transformer and restrict the eigenvalues of $W \sim \mathcal{N}(0, I)$ in a band of $[0.3, 0.9]$. We notice that with these $W$ approximately $2\%$ of the sequences lead to very large values. To ease trainability, we therefore clip all the values of those sequences to values between $[-2,2]$.

\begin{figure}[h!]
    \centering
    \begin{minipage}{0.68\textwidth}
        \centering
        \includegraphics[width=\linewidth]{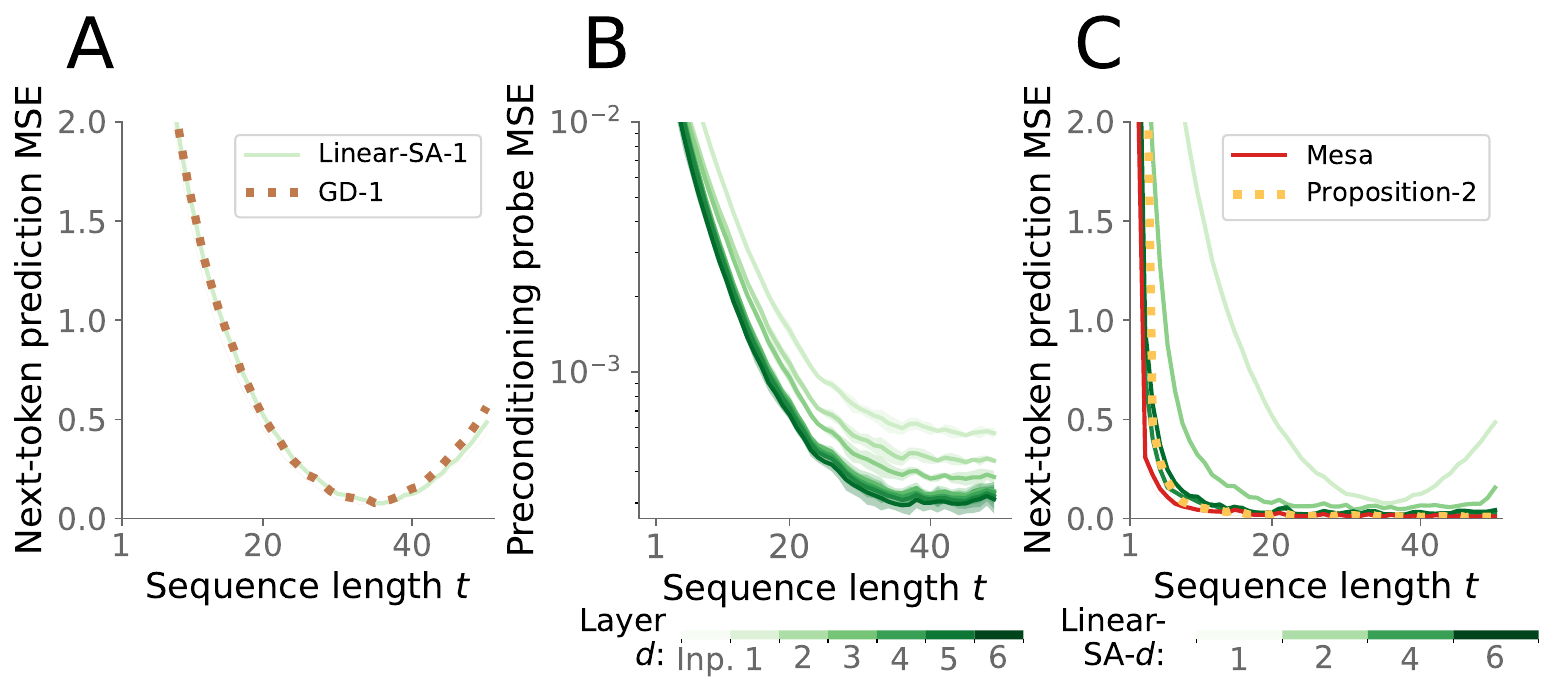}
    \end{minipage}
    \begin{minipage}{0.28\textwidth}
        \centering
        \raisebox{1pt}{\includegraphics[width=\linewidth]{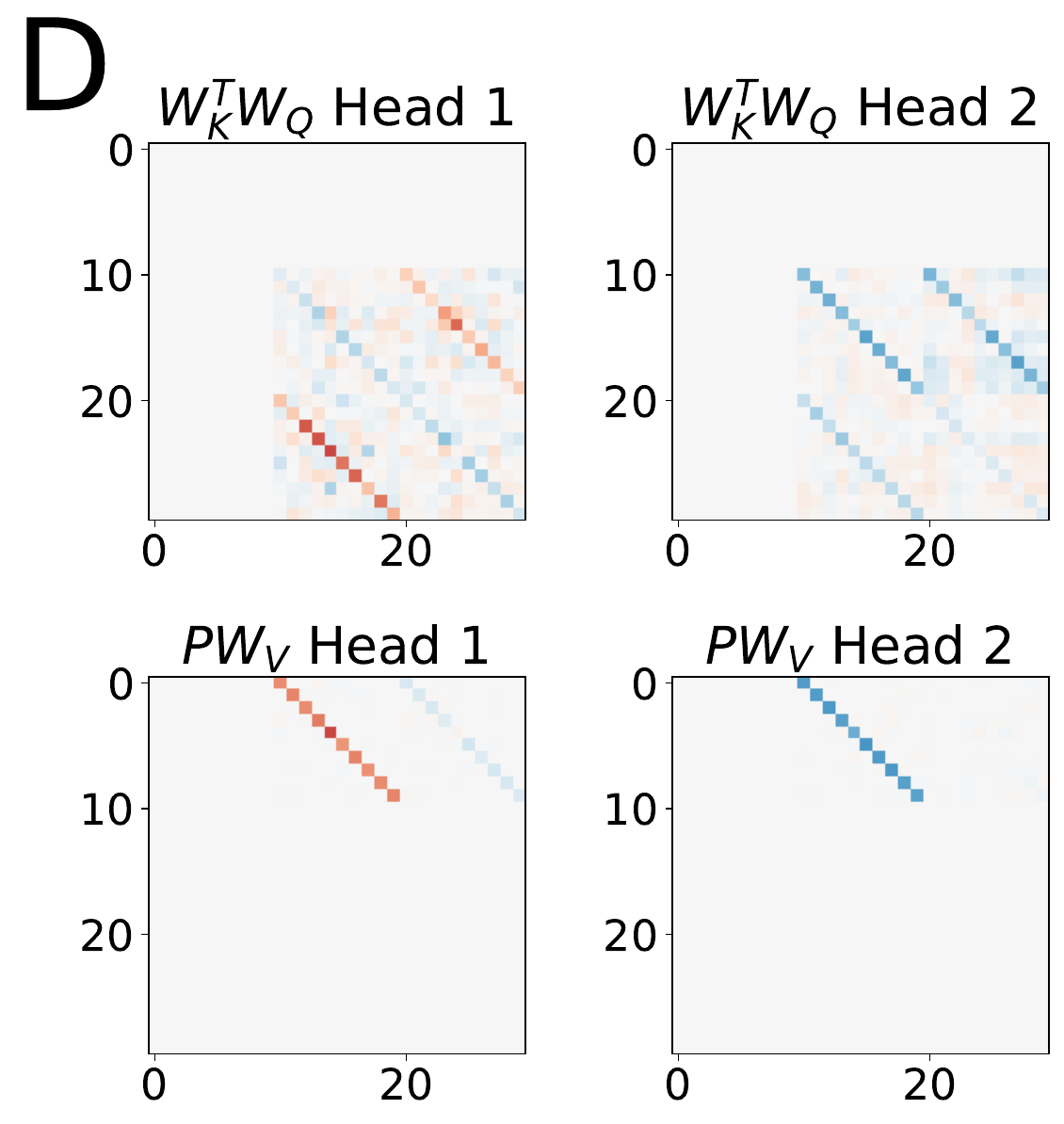}}
    \end{minipage}
    \caption{Evidence for mesa-optimization in Transformers trained on contracting linear dynamics. (\emph{A}) At convergence, models trained on contracting sequences exhibit the same in-context learning performance (measured as the loss as a function of sequence length) as 1 step of gradient descent (dashed line), as in our findings for models trained on data generated by an orthogonal teacher. (\emph{B}) In six-layer linear self-attention models trained on constructed tokens, we find that linear probing of preconditioned inputs $(S_{t-1}S_{t-1}^\top + 1/\lambda I)^{-1}s_t$ improves with depth and context length, consistent with the mesa-optimizer of Proposition $2$ and our findings for the orthogonal-teacher setting. (\emph{C}) For deeper models, performance in this setting increases. We find that the mesa-layer outperforms any other model and that a six-layer linear self-attention model can be explained by Proposition $2$. (\emph{D}) We again find highly structured weights that, in the shown two-head-one-layer case, can implement an update step of gradient descent.}
    \label{fig:contracting}
\end{figure}

When training a single layer of linear self-attention, see Figure \ref{fig:contracting}, we again observe that the trained layer matches the performance of a single step of gradient descent. We furthermore find clean weight structure, comparable to the weights trained on sequences which are generated by an orthogonal teacher, see Figure \ref{fig:2-head-1-LSA-identification-linear-dynamics}.

For multi-layer linear transformers we find both gradually increasing probing of preconditioned inputs as necessary for our hypothesis, Proposition $2$, as well as gradual performance improvement for deeper Transformers.

\subsection{Fixed-teacher linear dynamics} Here we analyse the setting where every sequence shares the same single fixed orthogonal transition matrix $W^* \in \mathbb{R}^{n_h \times n_h}$, and only the initial state $h_1 \sim \mathcal{N}(0,1)$ is sequence-specific. Thus, in this setting there is no need to infer $W^*$ in-context.

\begin{figure}[h!]
    \centering
    \begin{minipage}{0.68\textwidth}
        \centering
        \includegraphics[width=\linewidth]{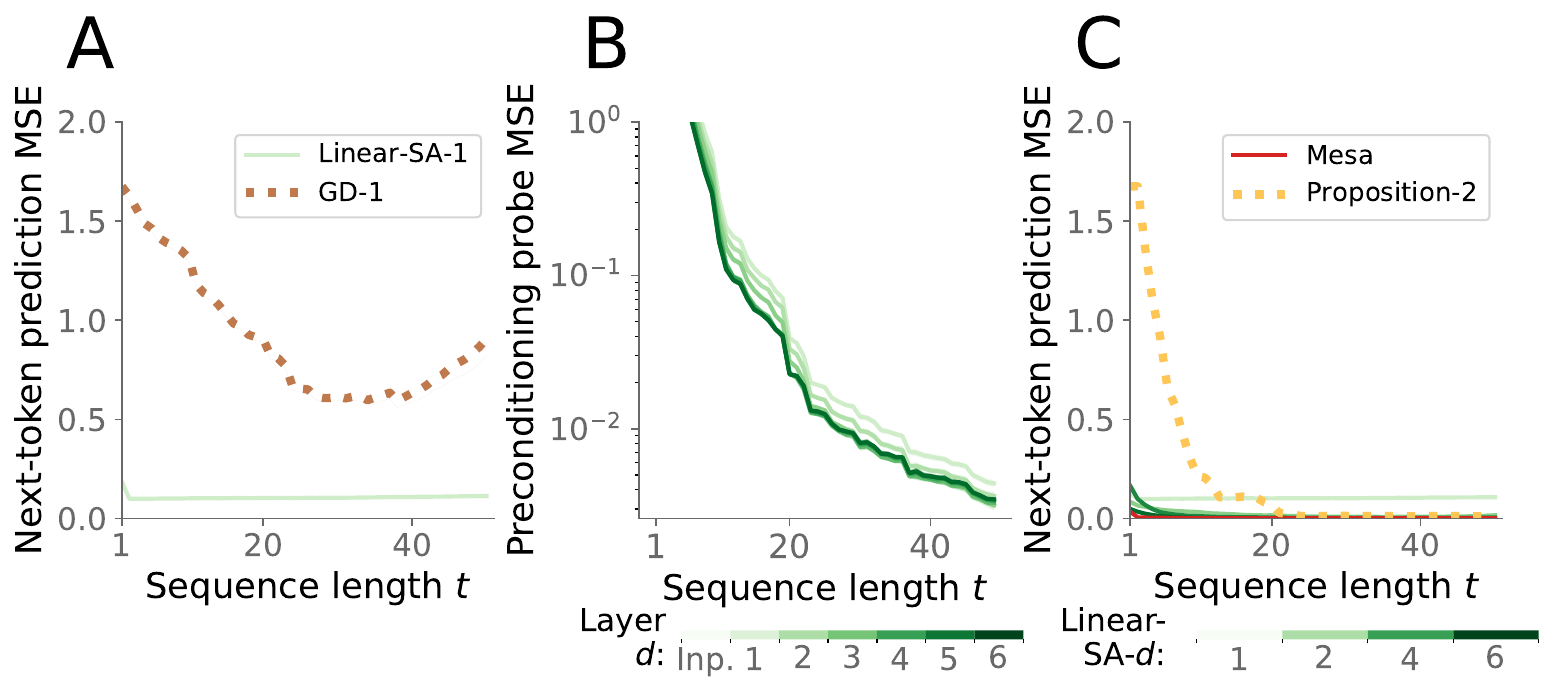}
    \end{minipage}
    \begin{minipage}{0.28\textwidth}
        \centering
        \raisebox{1pt}{\includegraphics[width=\linewidth]{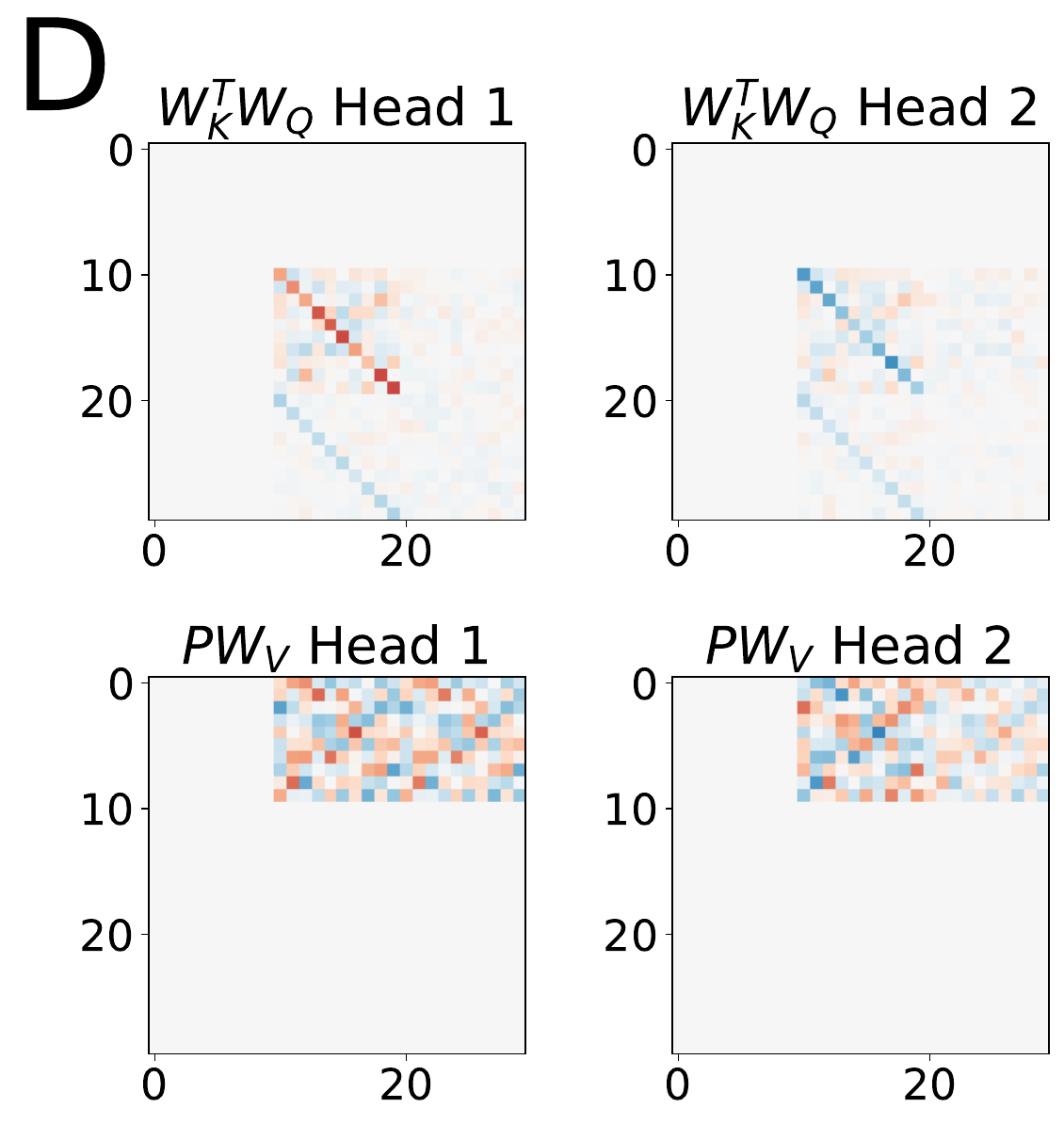}}
    \end{minipage}
    \caption{No evidence for mesa-optimization in Transformers trained on fixed-teacher linear dynamics, as predicted by our theory. (\emph{A}) At convergence, one layer linear self-attention transformers trained on fixed-teacher linear sequences significantly outperform the performance achieved by a single update step of gradient descent (dashed line). (\emph{B}) In six-layer linear self-attention models trained on constructed tokens, we only find very weak linear probing of preconditioned inputs $(S_{t-1}S_{t-1}^\top + 1/\lambda I)^{-1}s_t$ and only barely see gradual improvement over depth. (\emph{C}) Various deep linear self-attention and mesa- transformers drastically outperform optimization algorithms when evaluated on test sequences for the same fixed teacher. (\emph{D}) We find less structured and less interpretable weights in a trained one-layer transformer.}
    \label{fig:fixed_w}
\end{figure}

We report the results for the experiments in Figure \ref{fig:fixed_w}. We observe that for this case even a one-layer linear self-attention Transformer drastically outperforms an update step of gradient descent. Furthermore, we find no evidence for the mesa-optimizers of Propositions 1 and 2, neither in the weights, which appear less structured and less interpretable, nor in linear probings of preconditioned tokens, where we barely observe a gradual improvement over layers as well as an overall worse probing performance. Lastly, all trained transformers, including a single mesa-layer seem to outperform optimization-algorithms in this settings, indicating that the models learn the fixed teacher and thereby predict with very low error already very early in the sequence, as we also find in next-token prediction analyses.

\section{Experimental details}

\subsection{Training Transformers on fully observable linear dynamical systems}
\label{apx:exp-details-linear}

We provide here details about the training details of the Transformer models when training on the fully observable linear dynamics setting.
As already stated in the main text, we train all Transformer models by minimizing the following classical autoregressive prediction error objective regression loss:
\begin{equation}
\mathcal{L}(\theta) = \mathbb{E}_{e\sim p(e)}\!\left[\frac{1}{2}\sum_{t=1}^{T-1}\|e_{t+1} - f_{t}(e_{1:t}, \theta) \|^2\right].
\end{equation} \label{apx:transformer-training-objective}

In all of our experiments, we employ causal masking during self-attention, implemented in the same way as in the majority of auto-regressive language modeling experiments. Specifically, during the self-attention operation we zero out the elements corresponding to the upper triangular matrix of the attention map, except for the diagonal itself. We do this both for the linear attention layer and for the mesa-layer. In practice, for softmax self-attention the incoming logits to the softmax are set to $-1e^{30}$. We ran into stability issues especially when training models with linear layers. To mitigate those, we simply clipped the activations of the forward pass to values between $[-4, 4]$ for linear self-attention Transformer-layers, which stabilized training significantly. 
Hyperparameters and other experimental details can be found in table \ref{tab:hps-linear}.

\begin{table}[bt]
\begin{center}
\caption{Hyperparameters for all settings and model variants when training on simple fully observable linear dynamics.}

\label{tab:hps-linear}
\footnotesize
\begin{tabular}{@{}ll@{}}
\toprule
Hyperparameter            & Value/Description \\
\toprule
 Context size & We used length 50, except for the ICL experiments, where we used length 224 \\& and the softmax-linearization experiments where we vary the context size \\& according to the ratio context size = $4\cdot n_h$. \\
 Optimizer &Adam \citep{kingma_adam_2015} with $\epsilon= 1e^{-8}, \beta_1 =0.9, \beta_2=0.999$ \\
 Weight decay & 0.1 for constructed tokens, 0.05 otherwise \\
  Batchsize &  256, except for ICL and Linearization due to memory constraints, here 128 and 64, resp.\\
 Gradient clipping & 1.0 across models \\
 Activation clipping & Clip $[-4,4]$ for all linear models trained on constructed tokens, no clipping otherwise.\\
  Positional encodings &  We concatenate positional encodings of dimension 40 to queries and keys before \\& computing the self-attention in the first layer for models trained on \\& unconstructed tokens, otherwise no positional encodings. \\
   Dropout & We do not use Dropout for any model. \\
  Architecture 1-L., Constr. &  We use a 1-layer, 2-head, key-size 20, dim-40-tokens, no input- or output-embedding\\& architecture for single-layer models trained on constructed tokens. \\
  Architecture k-L. $(k>1)$, Constr. & We use a $k$-layer, 4-head, key-size 20, dim-40-token, no input- or output- \\&embedding architecture for the multi-layer models (softmax and linear)\\& trained on constructed tokens for the probing analysis and used \\& key-size 40 for the interpolation.\\
  Architecture Full-softmax, No Constr. & We use a 7-layer, 4-head, key-size 20, dim-10-tokens, dim-40-\\&embedding- architecture with input- and output-embedding layers for \\& full-fledged softmax-only-models.\\ 
  Architecture Hybrid-mesa, No Constr. & We use 2-layer, 4-head, key-size 20, dim-10-tokens, dim-40-\\&embedding-architecture with inputs- and output embedding layers. First a \\& softmax-self-attention layer, then a single Mesa-layer.\\
  Architecture Full-mesa, No Constr. & We use 2-layer, 4-head, key-size 20, dim-10-tokens, dim-40-\\&embedding-architecture with inputs- and output embedding layers. Both layers are \\& mesa-layers.\\
 Weight initialization &  $W \sim \mathcal{N}(0, \sigma^2)$ with $\sigma^2= 0.0002$ for models trained on constructed tokens\\& and $\sigma= 0.05$ for all other models. We always fixed the bias parameters to zero. \\
Learning rate (\& scheduler) &  For models trained on non-constructed tokens, we used linear warm-up \\& starting from $0$ to $7e^{-4}$ in $1000$ steps, Cosine annealing to $1e-5$ for the next \\& $10000$ (single-layer interpolation experiments), $30000$ (other experiments) steps. \\& We note here that we train the models only for at most $10000$ steps, except for the ICL-\\& setting where we do Cosine annealing for $60000$ steps and train for $40000$ steps. \\&For models trained on constructed tokens, we used a fixed learning rate of $1e^{-4}$.\\
Mesa regularization $\lambda$ & We initialize the learnable regularization parameter $\lambda$  for every mesa-head to 1. \\
\bottomrule
\end{tabular}
\end{center}
\end{table}

\subsubsection{Single-layer linear self-attention Transformer}
\label{apx:single-layer-details}
We analyze single-layer, two-head, key-size-20 linear self-attention Transformers, trained on constructed tokens, by comparing their performance with other models and providing an interpolation in parameter space between trained Transformers and the provided construction for Proposition $1$, which is described by only a few hyper-parameters. We read out the predictions from the first $D_s$ entries of the outputs (which initially contain a zero-vector). For the performance analysis, these models are compared to a Proposition $1$, thus a single gradient descent update step on the auto-regressive loss. The optimal learning rate for this gradient descent step is line-searched. 

\begin{figure}[h!]
    \centering    \includegraphics[width=0.85\textwidth]{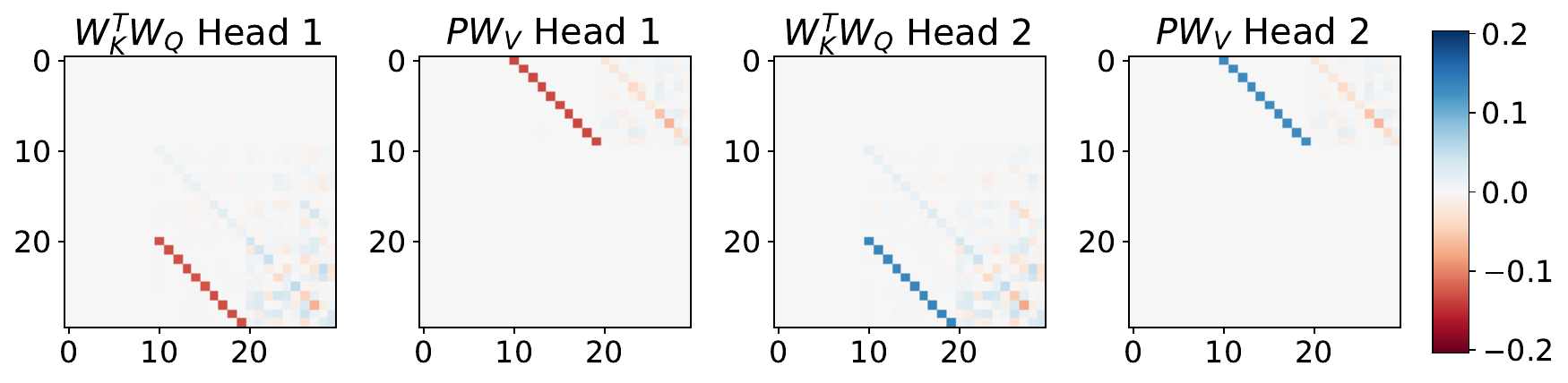}
    \caption{Mesa-optimization in a trained linear self-attention layer. We inspect the parameters of a two-headed, linear self-attention layer trained to predict the future state of a linear dynamical system. The dominant pattern obtained after learning corresponds to our mesa-gradient descent construction. The faint additional structure can be further reverse-engineered, and results from a modified mesa-objective function, $L_t(\Phi) = \sum_{t^\prime=1}^{t-1} \frac{1}{2} \|s_{t^\prime+1} - \Phi s_{t^\prime} \|^2$, discovered by base-optimization of Equation ~\ref{apx:transformer-training-objective}. Please compare to the similar structure of the weight matrix products of our construction. Please note that these matrices are actually of shape $40\times 40$. Here we only show the $30\times 30$ dimensional sub-matrix containing nonzero entries.}
    \label{fig:2-head-1-LSA-identification-linear-dynamics}
\end{figure}

\emph{Interpolation details:} We first train a Transformer, then extract scalar parameters of the mesa-optimization algorithm, from the $D_s \times D_s$-shaped sub-matrices by taking the mean of the sub-diagonals of the matrix products $W_k^\top W_q$, $PW_v$ (cf. \ref{fig:2-head-1-LSA-identification-linear-dynamics}). We proceed by using these to both build a construction of sparse weight matrices, each consisting only of identity-sub-matrices (scaled by the resp. parameters), and, for the single-layer case, also directly compute a loss for the hard-coded implementation of Proposition $1$ with the respective hyper-parameters. Then, during a second training-run of a Transformer for the same initial conditions, we simultaneously compute the test loss for an interpolation, where we average equally not between the single weight matrices, but between the correct weight-matrix-products per head to obtain a new, interpolated model. The reason for this procedure is the non-uniqueness of weight matrices to obtain the found matrix products. We repeat this procedure for 5 different seeds, train a newly initialized Transformer each time and plot the obtained mean and standard deviation values for the test loss during training.

\subsubsection{Multi-layer linear self-attention Transformer}
\label{apx:multi-layer-details}
For the multi-layer experiments, we use different settings: For the experiments with constructed tokens, we use a $k$-layer ($k>1$), no input- or output-embedding layer architecture, we found that forward-pass activation clipping in linear self-attention based Transformers after each layer greatly stabilized training and hence clip activations in a band of $[-4,4]$.

\emph{Interpolation details:} The interpolation of multi-layer transformers when training on the token construction, we follow the procedure described in the previous subsection, per layer, but extend it to 4-head key-size 40 self-attention layers: We read off the parameters as the mean of the diagonals of the respective $n_s \times n_s$ sub-matrices of the resulting matrix weight products $W_k^\top W_q$, $PW_v$ per head of a trained Transformer. Then we construct sparse weight matrices consisting of identity-sub-matrices (scaled by the resp. parameters). We name this algorithm Compressed-Alg-$6$. We proceed as for the single-layer experiment and re-train the Transformer from the initial conditions, but during training also report the test loss of a model that is obtained by equally averaging the weight products of our construction for Compressed-Alg-$6$ and the Transformer. We average the products and not the single weight matrices for the same reasons stated in the previous subsection \ref{apx:single-layer-details} and report the loss obtained in runs for $5$ different seeds.

\subsubsection{Full-fledged Transformers}
\label{apx:full-fledged-details}
For the experiments with full-fledged Transformers, we use either a 7-layer full-softmax architecture or 1+1 softmax-mesa and mesa-mesa hybrid-models. In all full-fledged models, we have input- and output-embedding layers, and the first layer always incorporates the logic for the positional encodings, while the other Transformer layers are either $6$ softmax self-attention layers, or $1$ mesa layer (1+1-layer architecture). The positional encodings are concatenated to the outputs of the key- and query projections before the computation of the attention.

\begin{figure}[h!]
    \centering
    \includegraphics[width=.2\textwidth]{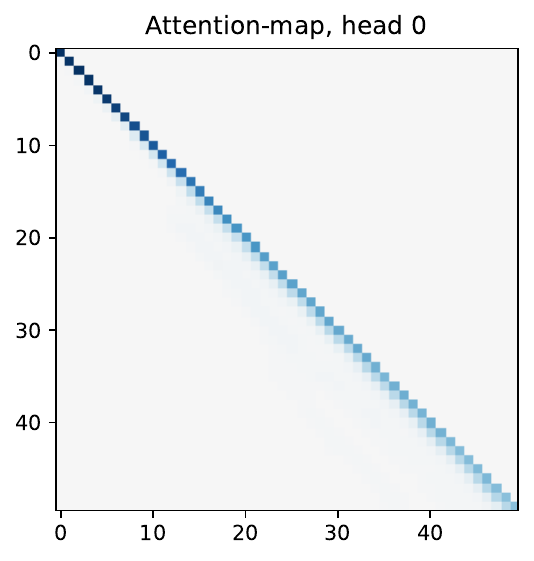}
    \includegraphics[width=.2\textwidth]{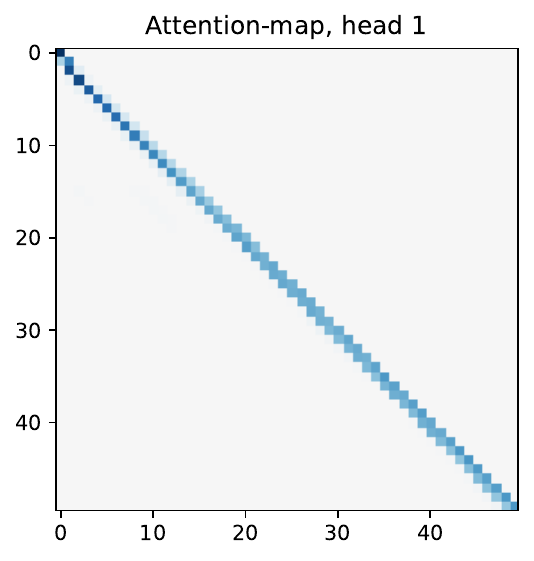}
    \includegraphics[width=.2\textwidth]{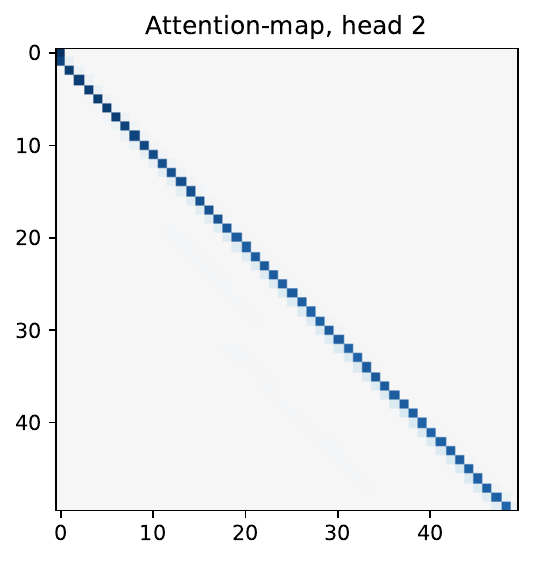}
    \includegraphics[width=.2\textwidth]{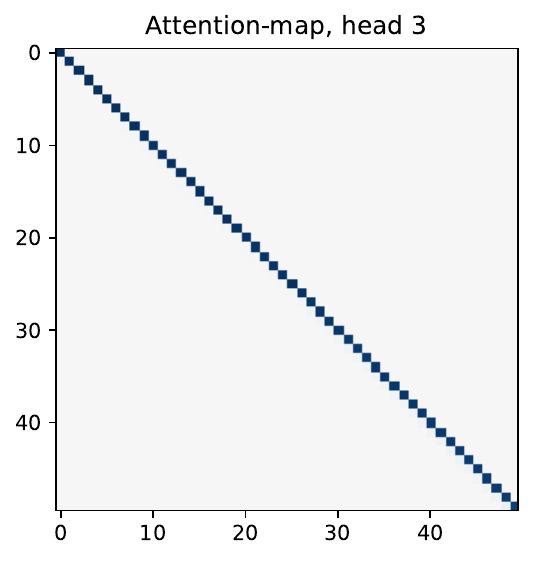}
    \caption{Softmax attention maps of the first softmax self-attention layer when training a softmax-only Transformer on unconstructed inputs. We visualize all four heads of the first softmax-attention layer and observe strong copying behavior, as predicted by the provided theory, in the heads i.e. full attention on the current and the previous token. We average the attention maps over a batch of 2048.}
    \label{apx:full-softmax-attention}
\end{figure}

\begin{figure}[h!]
    \centering
    \includegraphics[width=.2\textwidth]{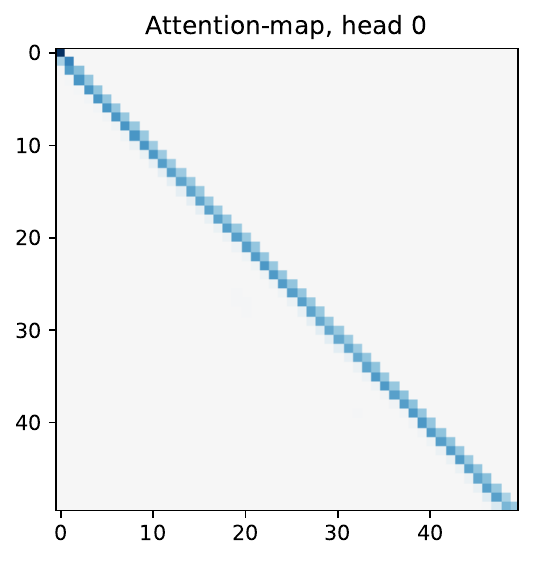}
    \includegraphics[width=.2\textwidth]{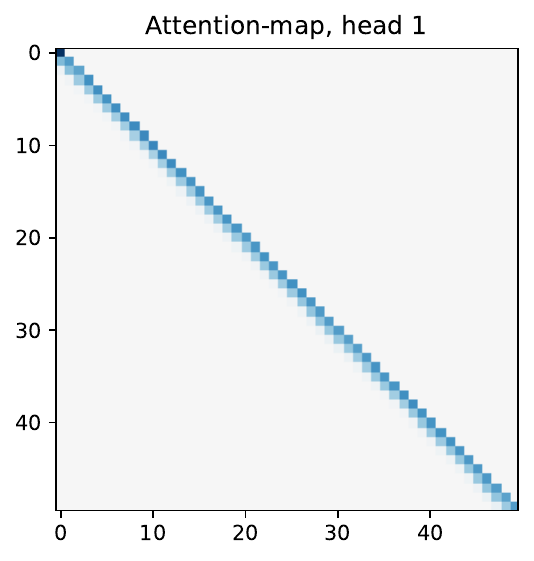}
    \includegraphics[width=.2\textwidth]{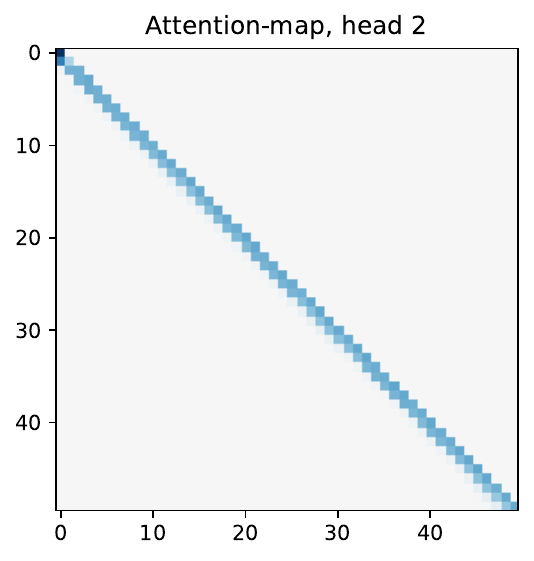}
    \includegraphics[width=.2\textwidth]{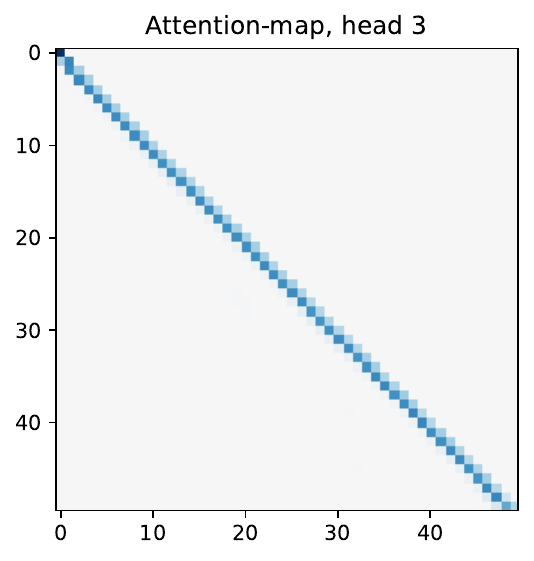}
    \caption{Softmax attention maps of the first softmax self-attention layer when training a hybrid-mesa Transformer on unconstructed inputs. We visualize all four heads of the first softmax-attention layer and observe strong copying behavior, as predicted by the provided theory, in the heads i.e. full attention on the current and the previous token. We average the attention maps over a batch of 2048.}
    \label{apx:hybrid-mesa-attention}
\end{figure}

\begin{figure}[h!]
    \centering
    \includegraphics[width=.75\textwidth]{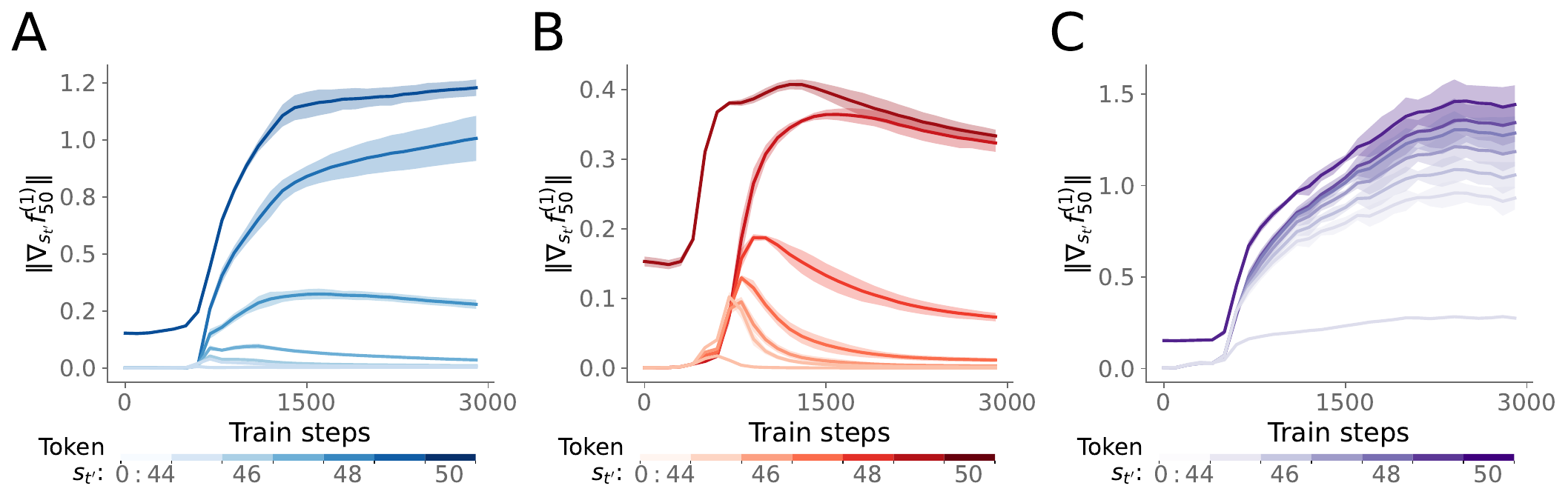}
    \caption{Gradient sensitivity analysis of activations after the first layer in various Transformer models over the course of training. The first softmax layer groups together neighboring tokens. This can be seen in the high sensitivity to the current and previous tokens of the outputs of the first layer of a softmax-only Transformer. For full-mesa models we find less clear binding of all previous tokens, which is also reflected in the token probing analyses, cf. \ref{fig:token-probing-fully}.}
    \label{fig:sensitivities-fully}
\end{figure}

\begin{figure}[h!]
    \centering
    \includegraphics[width=.75\textwidth]{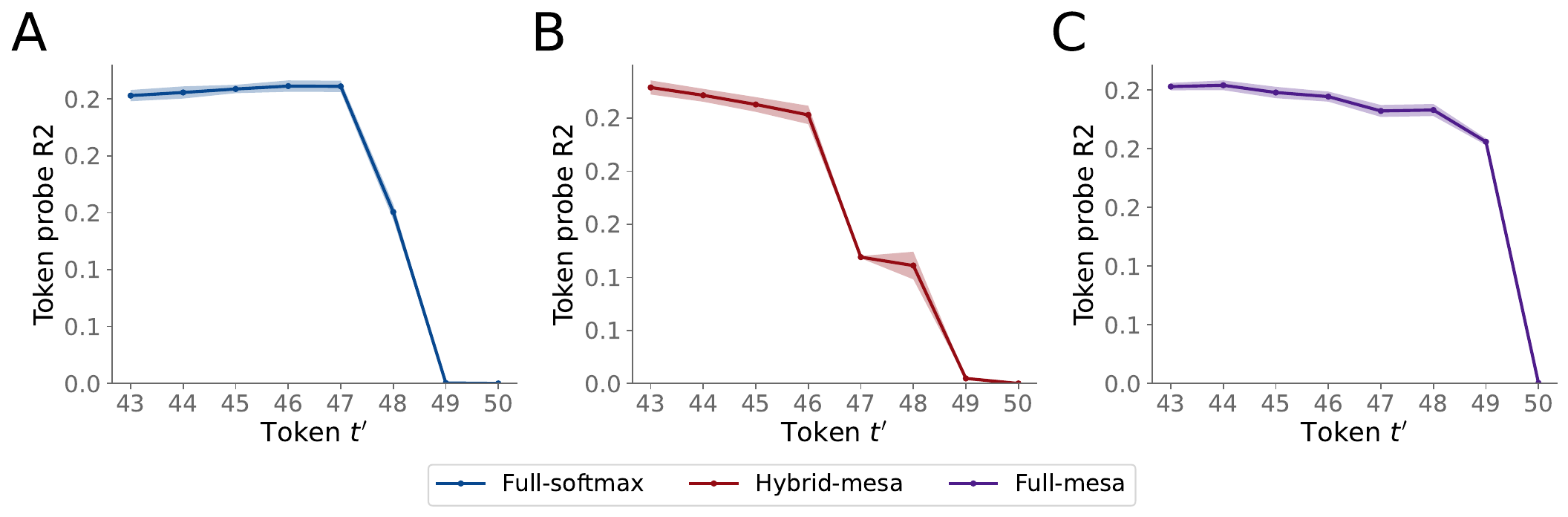}
    \caption{Token probing for various full-fledged Transformer models trained on fully observable linear sequences models. We find further evidence for a learned token binding process in the first layer, indicated by a very low decoding-loss for both the current and the previous token at a chosen time-step ($50$) over batches of test-sequences.}
    \label{fig:token-probing-fully}
\end{figure}

\emph{Analysing copying behaviour in full-fledged Transformers:} We examine Transformers trained on linear sequence models to understand if they learn a token-binding process in early layers to construct aggregate internal token representations, which are necessary for the proposed mesa-optimization algorithms in subsequent layers. We analyse the causally masked attention maps of trained models (cf. \ref{apx:full-softmax-attention}, \ref{apx:hybrid-mesa-attention}) and find clear data-independent attention on both the current and the previous token at each time-step. Furthermore, we propose a token-probing and a gradient sensitivity experiment (cf. \ref{fig:sensitivities-fully}, \ref{fig:token-probing-fully}) to understand if the transformed tokens after the first Transformer layer contain both the current as well as the previous token in the sequence, as necessary for our hypothesis. For the token probing, we report the performance of linear decoders trained to predict previous tokens from output. There, we linearly regress a batch of sequences at a single time-step against a range of previous time-steps and report the obtained MSE loss. We find that, as predicted by our hypothesis, Transformers learn a process that data-independently binds previous and current tokens at each time steps to construct the proposed representations internally. We support this evidence by further analyses where we compute the sensitivity norm $\| \nabla_{s_{t'}} f_{t}^{(1)}(s_{1:t},\theta)\|$ of the output of the first layer for all time steps $t' \leq t$. Furthermore we analyse full-mesa (first and second layer mesa) models and report the findings for the above experiments. Here, we find weaker and less clear - but still existing binding of previous tokens at each time-step.

\emph{Analysing optimization algorithms in full-fledged Transformers:} We proceed by analysing later layers in a variety of experiments. First, we compare the performance across fresh test sequences of the full-fledged model architectures and a hard-coded implementation of our proposed mesa-optimization that consists of six steps of preconditioning an internal optimization problem which is then solved in the last layer by an update step of gradient descent. Previously, we learn the parameters for the Chebyshev-iteration method for inverting matrices (as necessary for the proposed optimization procedure) by optimizing directly for solving fully observable linear sequence models generated by the same teacher as used in this setting. Furthermore, we find strong evidence for mesa-optimization in various activation-probing experiments. We linearly regress activations separately per time-step against targets and preconditioned inputs as predicted by our Proposition-$2$, $(S_{t-1}S_{t-1}^\top + 1/\lambda I)^{-1}s_t$ and find gradually increasing performance over layers in both experiments. 

\subsubsection{Testing autoregressively trained Transformers on few-shot in-context-learning}
\label{apx:icl-details}
\begin{figure}[h!]
    \centering    
    \includegraphics[width=0.7\textwidth]{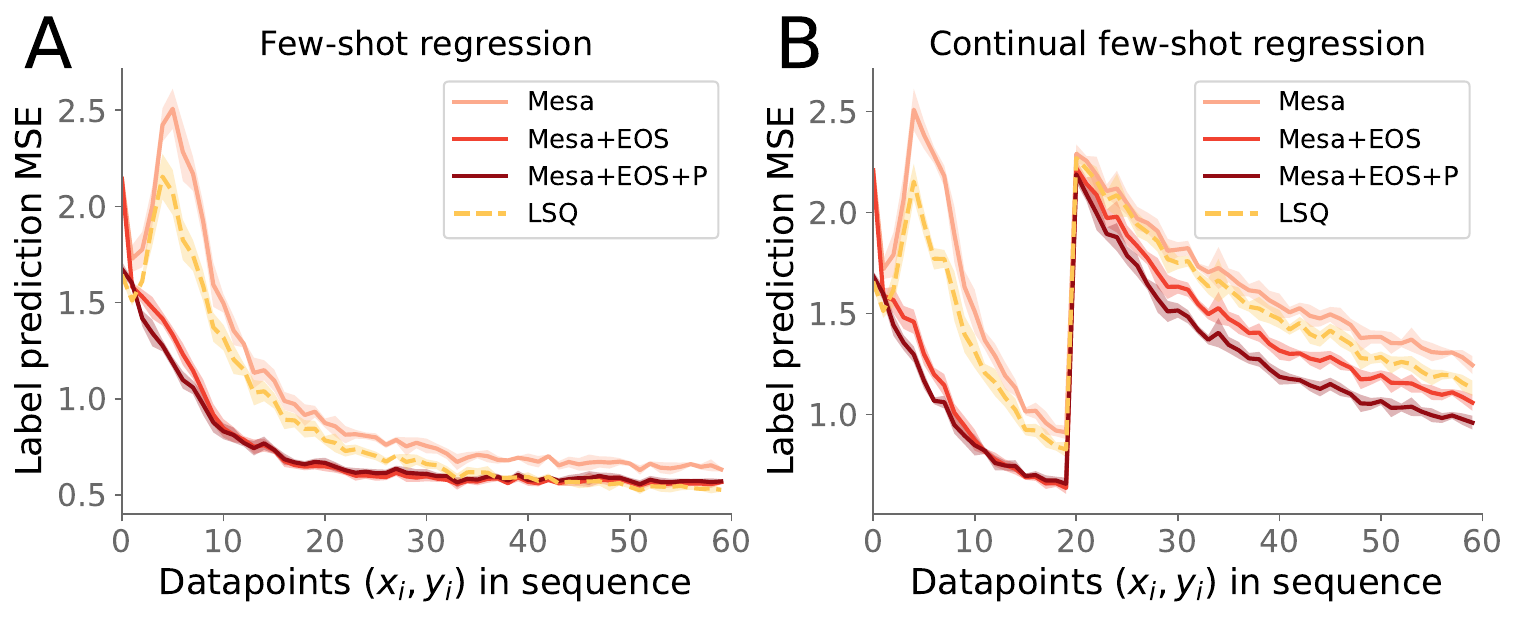}
    \caption{Autoregressive Transformers display in-context few-shot learning capabilities. After training a hybrid-mesa Transformer on autoregressive sequence prediction problems, we measure its ability to solve linear regression tasks in-context, without further parameter fine-tuning. The task training set is presented to the model in sequence, with each token corresponding either to an input or to its corresponding label. A final test input is provided and the loss is measured after completing the sequence using the autoregressive Transformer. (\emph{A}) The mesa-optimizers installed by autoregressive pretraining can be leveraged off-the-shelf to solve in-context supervised regression tasks, but yield sub-optimal regression performance (lightest red lines). In-context learning performance can be improved following the standard strategies of prompt (TF+EOS, light red lines) and prefix fine-tuning (TF+EOS+P, dark red lines). For comparison, we provide the loss achieved by an autoregressive linear model learned by least-squares (LSQ, yellow lines) (\emph{B}) Same analysis, now presenting two tasks in a row. The autoregressive models develop some in-context continual learning capabilities.}
    \label{fig:icl-mesa}
\end{figure}

We provide here details about the \textit{post}-training in-context learning experiment. For this experiment, we exclusively analyse full-fledged Transformers. After training, we "prompt" the model with few-shot regression datasets i.e. simply switch from sequences $[x_1, x_2, \dots, x_{t-1}, x_{t}]$ where $x_{t+1} = Wx_t$ and $x_0 \sim \mathcal{N}(0,I)$ to $[x_1, y_1, \dots, x_N, y_N]$ where $y_i = Wx_i$ and all $x_i \sim \mathcal{N}(0,I)$. Note that there is no  relation between $y_i, x_{i+1}$ as in the autoregressive case.
In both cases we sample $W$, if not stated otherwise from the same distribution i.e. as random orthogonal matrices. This results in a sequence length of $t=2N$ and $t=3N$ when incorporating $\texttt{EOS}$ tokens. Throughout the sequence we measure

\begin{equation}
\label{eq:icl-loss}
\mathcal{L}_i = \mathbb{E}\!\left[\frac{1}{2}\|y_{i} - f_{2i-1}(x_{i}; \{(y_j, x_j)\}_{j=1}^{i-1}) \|^2\right].
\end{equation}

for $i\geq 2$ depicted e.g. in Figure \ref{fig:icl-mesa}.

For the EOS-token fine-tuning experiments, we initialize a single vector $\texttt{EOS} \sim \mathcal{N}(0,I)$ and optimize this single vector on the same loss

\begin{equation}
\label{eq:eos-fine-tuning}
\mathcal{L}(\texttt{EOS} ) = \mathbb{E}\!\left[\frac{1}{2}\sum_{i=1}^{N}\|y_{i} - f_{3i-2}(x_{i}, \texttt{EOS}; \{(y_j, x_j)\}_{j=1}^{i-1}) \|^2\right]
\end{equation}

via batch gradient descent for 5000 steps with batchsize 256 on randomly sampled training data. Note that we interleave every datapair with an $\texttt{EOS}$ token i.e. $[x_1, y_1, \texttt{EOS}, x_2,  \dots, y_{N-1}, \texttt{EOS}, x_N, y_N]$ and we therefore increase the sequence length from $2N$ to $3N$.

For the prefix-prompt \texttt{P}, we fine-tune a single sequence of $20$ tokens which we append at the beginning of every in-context learning sequence. We initialize here again all vectors before training of the soft-prompt $\texttt{P}_i \sim \mathcal{N}(0,I)$ and optimize again the same loss with or without the additional (pre-trained, see above) $\texttt{EOS}$ token, 

\begin{equation}
\label{eq:prefix-fine-tuning}
\mathcal{L}(\texttt{P} ) = \mathbb{E}\!\left[\frac{1}{2}\sum_{i=21}^{N-20}\|y_{i-20} - f_{3i-2 + 20}(x_{i-20}, \texttt{P},\texttt{EOS}; \{(y_j, x_j)\}_{j=1}^{i-21}) \|^2\right],
\end{equation}

via batch gradient descent for 5000 steps with batchsize 256 on randomly sampled training data resulting in sequences $[P_1, \dots, P_{20}, x_1, y_1, \texttt{EOS}, x_2,  \dots, y_{N-1}, \texttt{EOS}, x_N, y_N]$.

We extend this analysis by a continual-in-context learning experiment where we demonstrate the in-context learning capabilities of autoregressively trained Transformers on two tasks shown in sequence in context.

\subsection{Linearizing softmax-Transformers}
\begin{figure}[t!]
    \centering    
    \includegraphics[width=1\textwidth]{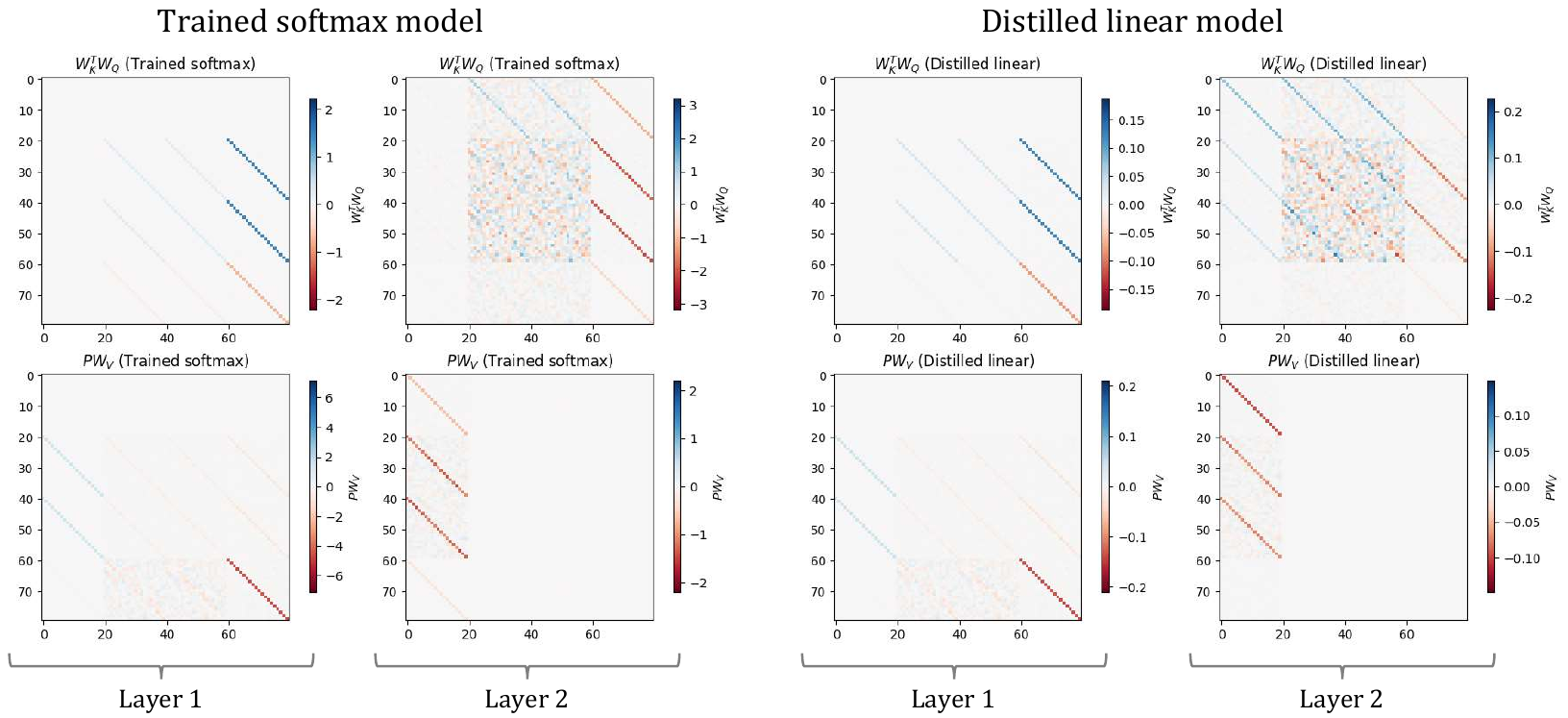}
    \caption{The weights of a distilled linear layer are surprisingly similar to those of the original full-softmax model. Here, we present the resulting weights for a linearization of a full-softmax, 2-layer-1-head model trained on constructed data with $T=80, n_s=20$.}
    \label{fig:linearization-weights}
\end{figure}

We provide here details and additional results about the linearization experiments. For the linearization analysis presented in the main text, we proceed as follows: First, we fix the ratio of context-size to (observed) data-dimension to $4:1$. Then, for each of the listed settings ($n_s \in \left[4,6,10,20,40,60\right]$ and $T$ according to the fixed ratio) we first train a classical full-fledged softmax-attention Transformer model on data generated by a linear-sequence generating teacher. We note here that for larger dimensions, the training becomes significantly more difficult in this setting. Then, for each layer in the model, we distill a separate linear self-attention layer by training it to `behave' like its softmax-counterpart. To this end, we record the outputs of the softmax-layer for a new input-sequence. Note that the inputs to the linear layer that we are training are not the original input-sequences, but rather the (transformed) sequences that are the activations \emph{before} the softmax-layer in the multi-layer softmax-Transformer. Hence, the distillation process is described by optimizing this objective:
\begin{equation}
\mathcal{L}(\theta_{linear}) = \mathbb{E}\left[\frac{1}{2}\sum_{t=1}^{T-1}||\text{SA}^{(l)}(s_{1:t},\theta_{\text{softmax}, l}) - \text{LSA}(f_t^{(l-1)}(s_{1:t},\theta_{\text{TF}}), \theta_{\text{linear}})||\right].
\end{equation}
Here, $\text{SA}^{(l)}$ denotes the softmax attention operation at the $l$-th layer of the full-softmax transformer, $\theta_{\text{softmax}, l}$ the (learned) parameters for this operation, LSA the linear self-attention layer and $f_t^{(l-1)}(s_{1:t},\theta_{\text{TF}})$ the activation after the $(l-1)$-th layer in the trained full-softmax Transformer, which will be the input to the linear layer we aim to distill. After this distiallation process is completed, we construct a model where we swap out the softmax operation at the respective layer and replace it by the distilled layer in the full-softmax model. Then we compare the performance of this new `linearized' Transformer with the original full-softmax model on a batch of test sequences and report the measured test loss.
Furthermore, we find that the distilled weights that were trained on the in- and outputs of a specific softmax-layer appear to be very similar to softmax-attention layers in structure, cf. \ref{fig:linearization-weights}.

\begin{figure}[t!]
    \centering    
    \includegraphics[width=0.8\textwidth]{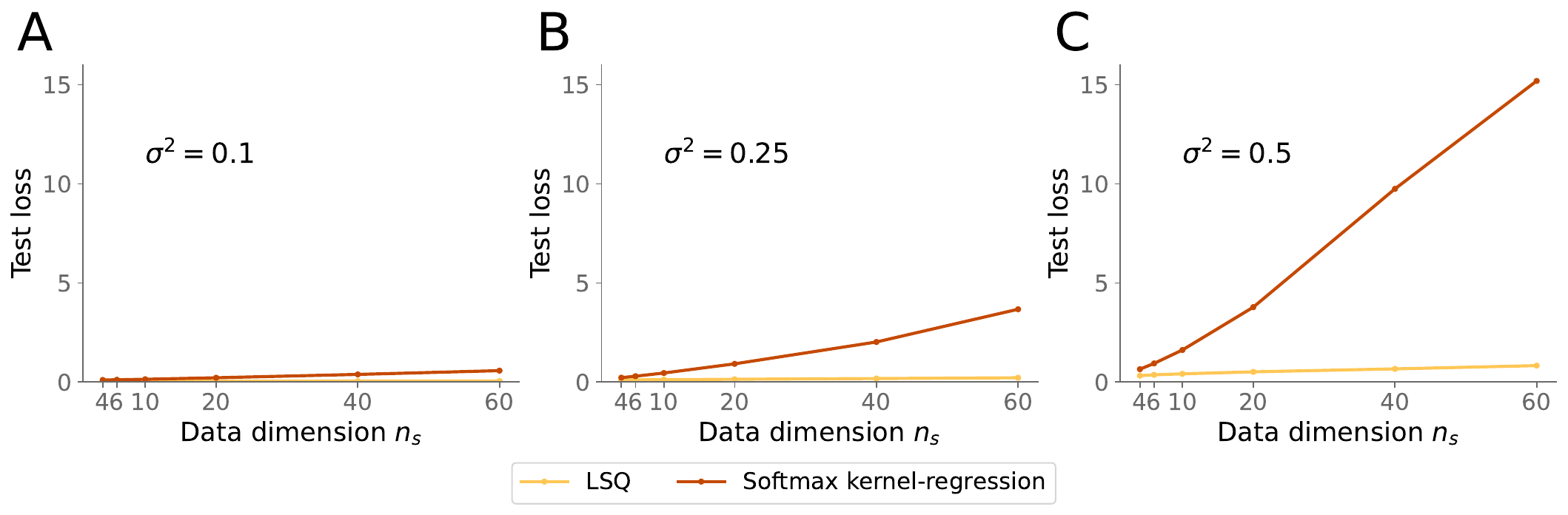}
    \caption{For different noise levels $\sigma_h^2$ in the sequence generation process, we analyse the performance of autoregressive models learned by regularized least squares and softmax kernel regression for increasing dimensions to underline the effect of the `curse of dimensionality' in our setting.}
    \label{fig:linearization-noise}
\end{figure}

Furthermore, we analyse and compare the performance of autoregressive models learned by regularized least squares and an generic interpolation algorithm, softmax-kernel-regression, in various settings as described above. We line-search the parameters necessary for regularization. Here, we extend these results and also analyse these settings for varying noise settings in the generating model. We report mean and standard deviation for three different seeds, each using generated data of batch-size $32$, in \ref{fig:linearization-noise}.

\subsection{Training Transformers on partially observable linear dynamical systems}
\label{apx:partobs-details}
For the experiments with partially observable linear dynamical systems, we directly analyze full-fledged Transformers trained on the observations. In detail, we use either a 7-layer full-softmax architecture or 1+1 softmax-mesa hybrid-models. In all models, we have input- and output-embedding layers, and the first layer always incorporates the logic for the positional encodings, while the other Transformer layers are either $6$ softmax self-attention layers, or $1$ mesa layer. The positional encodings are concatenated to the outputs of the key- and query projections before the computation of the attention. Generally, our models are trained on $n_s = 5$ - dimensional observations from a process with $n_h = 15$ - dimensional hidden states. Further training details can be found in Table \ref{tab:hps-partobs}.

\begin{table}[h]
\begin{center}
\caption{Hyperparameters for all settings and model variants when training on partially observable linear dynamics.}
\label{tab:hps-partobs}
\footnotesize
\begin{tabular}{@{}ll@{}}
\toprule
Hyperparameter            & Value/Description \\
\toprule
 Context size & We use context size $T=50$ \\
 Optimizer &Adam \citep{kingma_adam_2015} with $\epsilon= 1e^{-8}, \beta_1 =0.9, \beta_2=0.999$ \\
 Weight decay & 0.05 across models \\
  Batchsize &  Batchsize 256\\
 Gradient clipping & 1.0 across models \\
 Activation clipping & No activation clipping\\
  Positional encodings &  We concatenate positional encodings of dimension = embedding-dimension \\& to queries and keys before computing the self-attention in the first layer for all models \\
   Dropout & We do not use Dropout for any model. \\
  Architecture Full-softmax, No Constr. & We use a 7-layer, 4-head, key-size $\min{(20, \text{embedding-dim.})}$, dim-5-input-tokens,\\& architecture with varying embedding dimensions in $[5,10,15,20,30,50,80]$ with \\& input- and output-embedding layers for full-fledged softmax-only-models.\\ 
  Architecture Hybrid-mesa, No Constr. & We use 2-layer, 4-head, key-size $\min{(20, \text{embedding-dim.})}$, dim-5-input-tokens, \\&architecture with embedding dimensions in $[5,10,15,20,30,50,80]$ with inputs- \\& and output embedding layers. First a softmax-self-attention layer, then a single Mesa-layer.\\
 Weight initialization &  $W \sim \mathcal{N}(0, \sigma^2)$ with $\sigma^2= 0.05$ for all models. We always fixed the bias parameters to zero. \\
Learning rate (\& scheduler) &  We used linear warm-up starting from $0$ to $4e^{-4}$ in $1000$ steps, \\& Cosine annealing to $1e-5$ for the next $30000$ steps. \\
Mesa regularization $\lambda$ & We initialize the learnable regularization parameter $\lambda$  for every mesa-head to 1. \\
\bottomrule
\end{tabular}
\end{center}
\end{table}

\begin{figure}[h!]
    \centering
    \includegraphics[width=.65\textwidth]{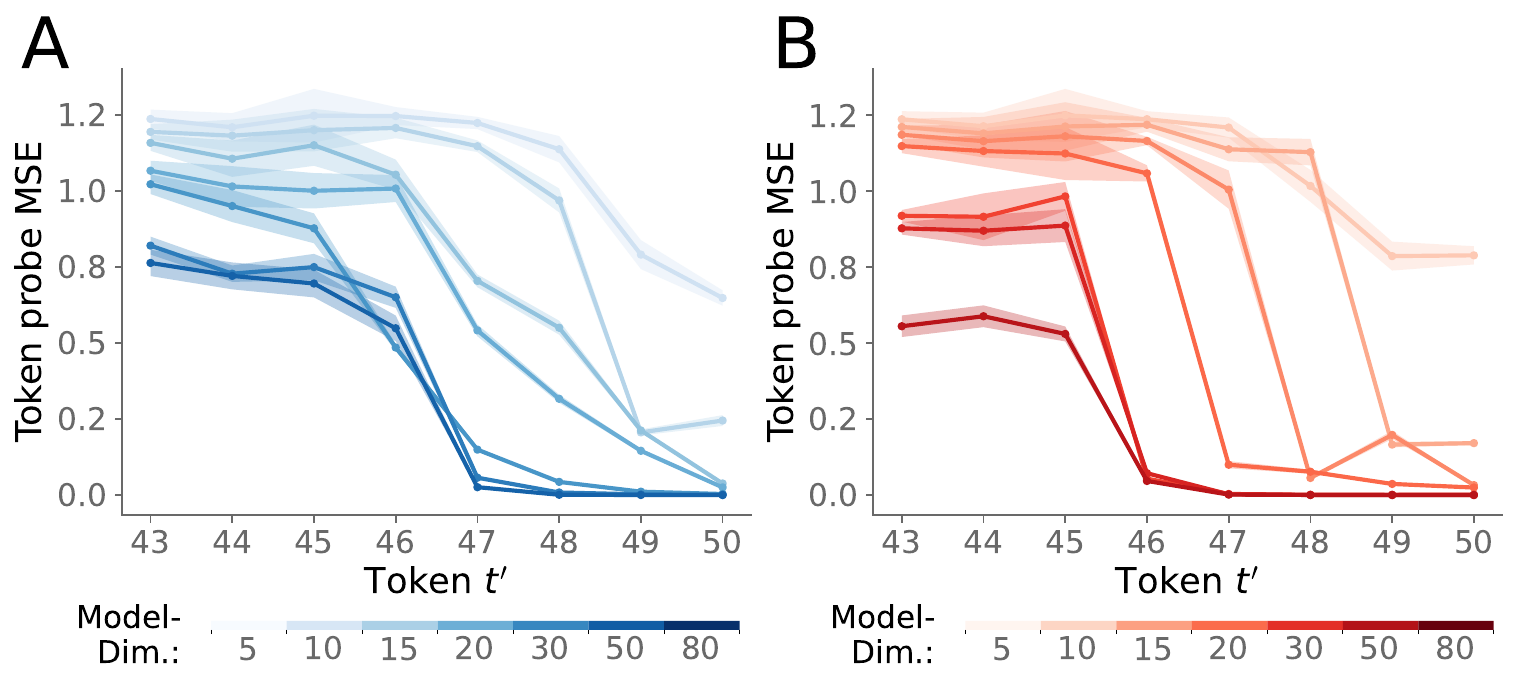}
    \caption{Token probing for Transformers trained on partially observable data. If we vary the embedding-dimension of the Transformers, we find that larger Transformers use the provided space to copy over relevant tokens.}
    \label{fig:token-probing-partobs}
\end{figure}

\begin{figure}[h!]
    \centering
    \includegraphics[width=.65\textwidth]{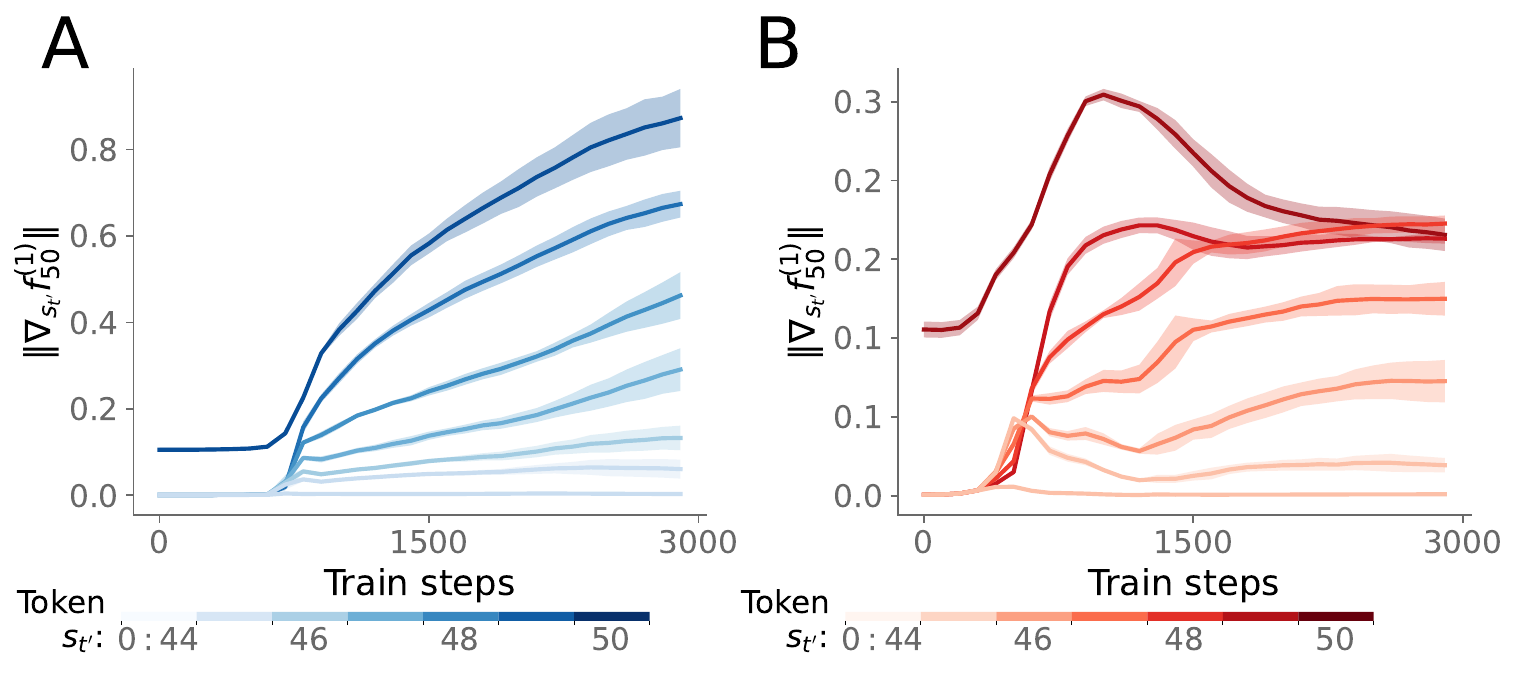}
    \caption{Gradient sensitivity analysis of activations after the first layer in full-softmax (A) and hybrid-mesa (B) Transformer models over the course of training. The first softmax layer groups together the current and multiple previous tokens as predicted by our hypothesis. This can be seen in the high sensitivity to the current and previous tokens of the outputs of the first layer of the Transformer models.}
    \label{fig:sensitivities-partobs}
\end{figure}

\emph{Analysing copying behaviour in full-fledged Transformers:} We examine Transformers trained on partially observable linear sequence models to understand if they learn a token-binding process in early layers to construct aggregate internal token representations, which are necessary for the proposed mesa-optimization algorithms in subsequent layers. As in the fully-observable setting, we use both a token-probing and a gradient sensitivity experiment to understand if the transformed tokens after the first Transformer layer contain the previous tokens in the sequence, as necessary for our hypothesis for partially observable models. For the token probing (cf. \ref{fig:token-probing-partobs}), we report the performance of linear decoders trained to predict previous tokens from output. There, we linearly regress a batch of sequences at a single time-step against a range of previous time-steps and report the obtained MSE loss for models of \emph{varying embedding dimension}. We find that as the embedding dimension grows, the probing of previous tokens becomes more clear and stable. Hence we infer that, as expected by our hypothesis, Transformers learn a process that data-independently binds previous and current tokens at each time steps to construct the proposed representations internally. We support this evidence by further analyses where we compute the sensitivity norm $\| \nabla_{s_{t'}} f_{t}^{(1)}(s_{1:t},\theta)\|$ of the output of the first layer for all time steps $t' \leq t$ (cf. \ref{fig:sensitivities-partobs}).

\begin{figure}[htbp!]
    \centering
    \includegraphics[width=.6\textwidth]{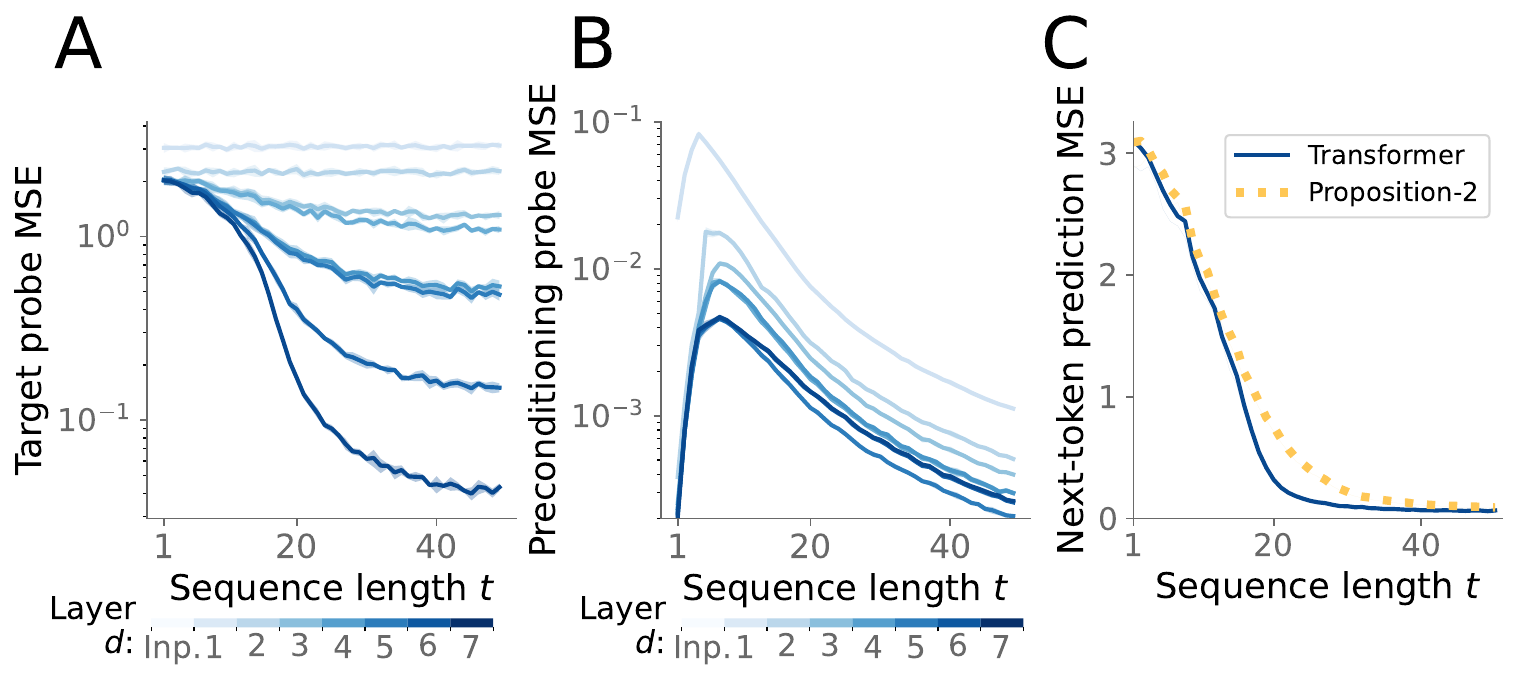}
    \caption{Evidence for mesa-optimization in standard (softmax) Transformers trained on partially observable linear dynamical systems. \emph{(A)} Linear probes decode next-token target $s_{t+1}$ from internal Transformer activations, with decoding performance improving with depth (intensity color-coded) and context length, consistent with gradual optimization of an internal next-token prediction model. \emph({B}) Likewise for preconditioned input $(Z_{t-1}Z_{t-1}^\top + 1/\lambda I)^{-1}z_t$ probing, where $z_t$ are constructed tokens, comprising of the past 5 observations, consistent with our findings in token probings for Transformers trained on partially observable dynamics and the mesa-optimizer of Proposition $2$. \emph{(C)} Next-token prediction error of a 7-layer Transformer (blue line) decreases with context length in a very similar way as 7 steps of Proposition $2$ on constructed tokens as predicted by our hypothesis for partially observable linear dynamical systems (dashed yellow line), with hyperparameters of the latter set for best performance, not to match Transformer behavior.}
    \label{fig:layer-probing-partobs}
\end{figure}

\emph{Analysing optimization algorithms in full-fledged Transformers trained on partially observable linear dynamical systems:} We proceed by analysing later layers using the same method as in the fully observable setting (cf. \ref{fig:layer-probing-partobs}). We compare the performance across fresh test sequences of the full-fledged model architectures and a hard-coded implementation of our proposed mesa-optimization that consists of six steps of preconditioning an internal optimization problem which is then solved in the last layer by an update step of gradient descent. Previously, we learn the parameters for the Chebyshev-iteration method for inverting matrices (as necessary for the proposed optimization procedure) by optimizing directly for solving fully observable linear sequence models generated by the same teacher as used in this setting. Furthermore, we find strong evidence for mesa-optimization in various activation-probing experiments. We linearly regress activations separately per time-step against targets and preconditioned inputs as predicted by our Proposition $2$ for partially observable linear sequence models, $(Z_{t-1}Z_{t-1}^\top + 1/\lambda I)^{-1}z_t$ (here $z_t$ refers to an aggregation of previous $k = 5$ tokens in one constructed token) and find gradually increasing performance over layers in both experiments. 

\subsection{Training Transformers on fully observable nonlinear dynamical systems}
\label{apx:nonlin-details}

For the experiments with fully observable nonlinear dynamical systems, we also directly analyze full-fledged Transformers trained on non-constructed observation-tokens. In detail, we use either a 7-layer full-softmax architecture or 1+1 softmax-mesa hybrid-models. In all models, we have input- and output-embedding layers, and the first layer always incorporates the logic for the positional encodings, while the other Transformer layers are either $6$ softmax self-attention layers, or $1$ mesa layer (1+1-layer architecture). The positional encodings are concatenated to the outputs of the key- and query projections before the computation of the attention. Furthermore, we use MLPs with hidden dimension $300$ (factor $5\times$ if compared with embedding-dimension for the models, which we set to $50$-dimensional) and a specialized version of normalization, sum normalization, as introduced by \cite{schlag_linear_2021}, where we divide the query and key projections by their respective sums of components. Further training details can be found in \ref{tab:hps-nonlin}.

\begin{table}[h]
\begin{center}
\caption{Hyperparameters for all settings and model variants when training on fully observable nonlinear dynamics.}
\label{tab:hps-nonlin}
\footnotesize
\begin{tabular}{@{}ll@{}}
\toprule
Hyperparameter            & Value/Description \\
\toprule
 Context size & We use context size $T=50$ \\
 Optimizer &Adam \citep{kingma_adam_2015} with $\epsilon= 1e^{-8}, \beta_1 =0.9, \beta_2=0.999$ \\
 Weight decay & 0.05 across models \\
  Batchsize &  Batchsize 256\\
 Gradient clipping & 1.0 across models \\
 Activation clipping & No activation clipping\\
  Positional encodings &  We concatenate positional encodings of dimension 60 to queries and keys before \\& computing the self-attention in the first layer for all models \\
   Dropout & We do not use Dropout for any model. \\
  Architecture Full-softmax, No Constr. & We use a 7-layer, 4-head, key-size $20$, dim-10-input-tokens architecture \\& with varying embedding dimensions $60$ with input- and output-embedding-\\&  layers for full-fledged softmax-only-models. The models comprise of MLPs \\& with hidden dimension $300$ and layer-normalization of query- and key-projections at each layer.\\ 
  Architecture Hybrid-mesa, No Constr. & We use 2-layer, 4-head, key-size $20$, dim-10-input-tokens architecture \\& with varying embedding dimensions $60$ with input- and output-embedding-\\&  layers. First a softmax-self-attention layer, then a single Mesa-layer. The models comprise of MLPs \\& with hidden dimension $300$ and layer-normalization of query- and key-projections at each layer.\\
 Weight initialization &  $W \sim \mathcal{N}(0, \sigma^2)$ with $\sigma^2 = 0.05$ for all models. We always fixed the bias parameters to zero. \\
Learning rate (\& scheduler) &  We used linear warm-up starting from $0$ to $4e^{-4}$ for hybrid-mesa and $1e^{-3}$ for full-softmax\\& models in $1000$ steps, Cosine annealing to $1e-5$ for the next $50000$ steps. \\&We only train for $40000$ steps.\\
Mesa regularization $\lambda$ & We initialize the learnable regularization parameter $\lambda$  for every mesa-head to 1. \\
\bottomrule
\end{tabular}
\end{center}
\end{table}

\begin{figure}[h!]
    \centering
    \includegraphics[width=.65\textwidth]{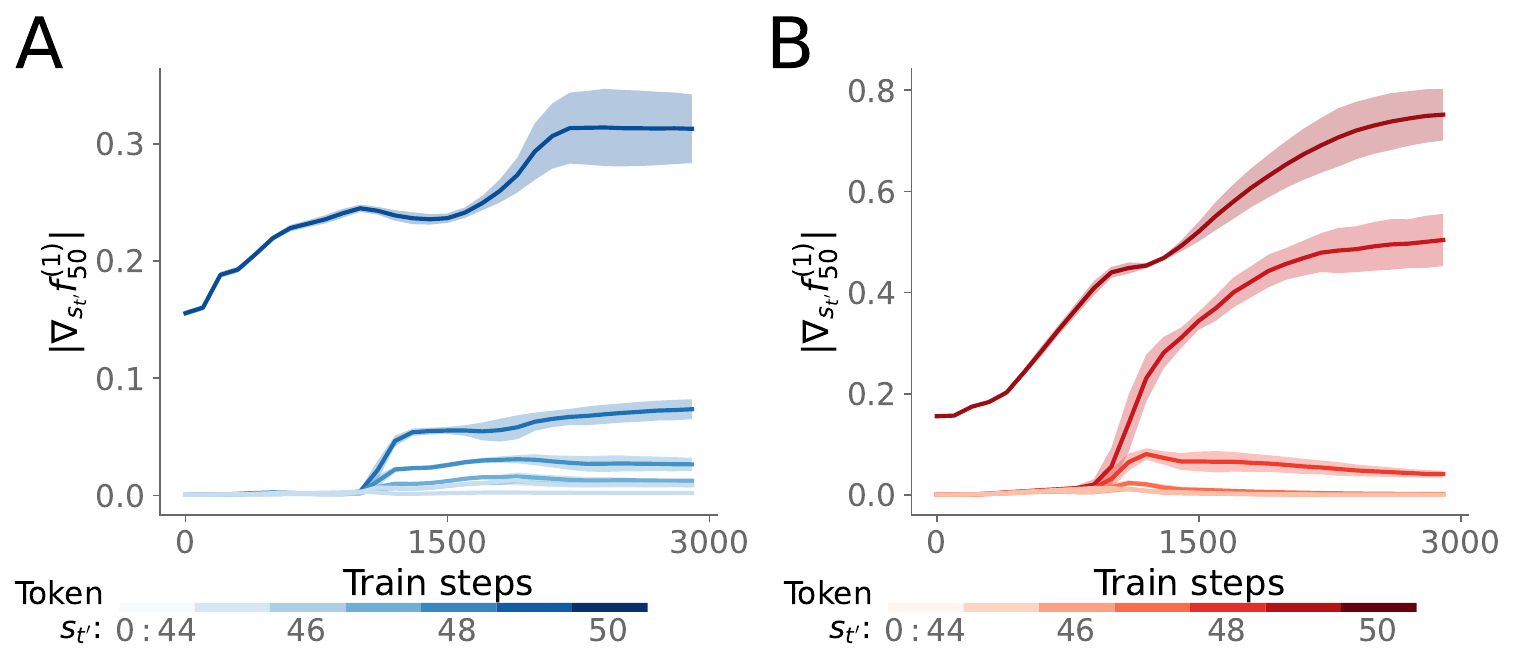}
    \caption{Gradient sensitivity analysis of activations after the first layer in full-softmax and hybrid-mesa Transformer models trained on fully-observable nonlinear dynamical systems over the course of training. The first softmax layer groups together the current and multiple previous tokens as predicted by our hypothesis. This can be seen in the high sensitivity to the current and previous tokens of the outputs of the first layer of the Transformer models.}
    \label{fig:sensitivities-nonlinear}
\end{figure}

\emph{Analysing copying behaviour in full-fledged Transformers:} As in the fully- and partially observable linear setting, we use both a token-probing and a gradient sensitivity experiment to test if trained Transformers learn a token binding mechanism in early layers. For the token probing, we report the performance of linear decoders trained to predict previous tokens from output. There, we linearly regress a the transformed token after the first layer for a batch of sequences at a single time-step against nonlinear transformed tokens from a range of previous time-steps and report the obtained MSE loss. Therefore, we employ the teacher used during training, $\text{MLP}^*$ Here, we also show further analyses where we compute the sensitivity norm $\| \nabla_{s_{t'}} f_{t}^{(1)}(s_{1:t},\theta)\|$ of the output of the first layer for all time steps $t' \leq t$. We report the results in Figure \ref{fig:sensitivities-nonlinear}.

\begin{figure}[htbp!]
    \centering
    \includegraphics[width=.6\textwidth]{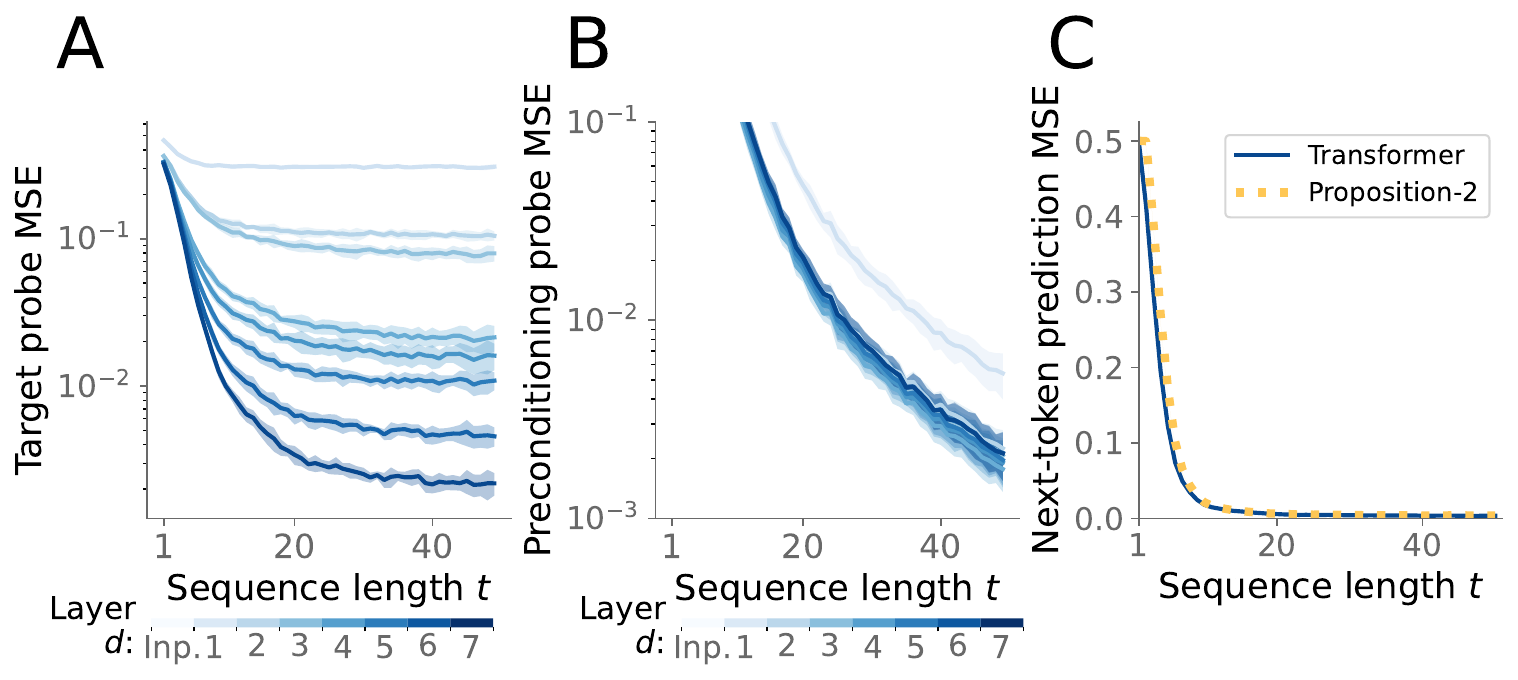}
    \caption{Evidence for mesa-optimization in standard (softmax) Transformers trained on fully observable nonlinear dynamical systems. \emph{(A)} Linear probes decode next-token target $s_{t+1}$ from internal Transformer activations, with decoding performance improving with depth (intensity color-coded) and context length, consistent with gradual optimization of an internal next-token prediction model. \emph({B}) Likewise for preconditioned input $(F_{t-1}F_{t-1}^\top + 1/\lambda I)^{-1}f_t$ probing, where $f_t$ are the nonlinearily transformed observations $f_t = \text{MLP}^*(s_t)$ using the teacher-MLP, consistent with the mesa-optimizer of Proposition $2$. \emph{(C)} Next-token prediction error of a 7-layer Transformer (blue line) decreases with context length in almost exactly the same way as 7 steps of Proposition $2$ (dashed yellow line), with hyperparameters of the latter set for best performance, not to match Transformer behavior.}
    \label{fig:layer-probing-nonlin}
\end{figure}

\emph{Analysing optimization algorithms in full-fledged Transformers trained on fully observable nonlinear dynamical systems:} We proceed by analysing later layers using the same method as in the linear settings (cf. \ref{fig:layer-probing-nonlin}). We compare the performance across fresh test sequences of the full-fledged model architectures and a hard-coded implementation of our proposed mesa-optimization that consists of six steps of preconditioning an internal optimization problem which is then solved in the last layer by an update step of gradient descent. Previously, we learn the parameters for the Chebyshev-iteration method for inverting matrices (as necessary for the proposed optimization procedure) by optimizing directly for solving fully observable nonlinear sequence models generated by the same teacher as used during training. Furthermore, we find strong evidence for mesa-optimization in various activation-probing experiments. We linearly regress activations separately per time-step against targets and preconditioned inputs as predicted by our Proposition $2$ for partially observable linear sequence models, $(F_{t-1}F_{t-1}^\top + 1/\lambda I)^{-1}f_t$ (here $f_t$ refers nonlinear transformed tokens MLP$^*(s_t)$ using the nonlinear teacher) and find gradually increasing performance over layers in both experiments.

\section{Language modeling}
\label{apx:language}

\begin{figure*}
\centering
\begin{minipage}{.245\textwidth}
    \includegraphics[width=1.\textwidth]{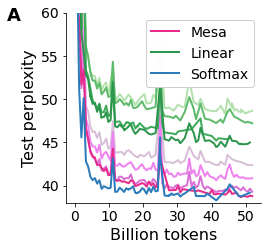}
\end{minipage}
\begin{minipage}{.245\textwidth}
    \includegraphics[width=1.\textwidth]{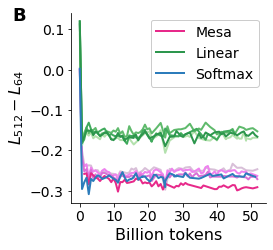}
\end{minipage}
  \caption{\looseness-1\textbf{Single-layer Transformers with key-shifts, the Pile.} We observe improved  (\textbf{A}) perplexity and (\textbf{B}) in-context learning scores when comparing one linear to one mesa layer with different DPFP sizes $\nu \in \{0,1,2,3\}$, corresponding inversely to color fade. Mesa layers consistently outperform linear layers, catching up with softmax.}
  \label{fig:training_trans_lin_key}
\end{figure*}

We present here first preliminary results on the performance of models which replace (some) softmax self-attention layer with the mesa-layer. 
Our hypothesis is that the mesa-layer will improve the in-context learning and working memory capabilities of a Transformer, in particular of the linear kind. We further hypothesize that this in turn translates to language modeling improvements, based on the high correlation between in-context learning and actual autoregressive loss reported by \citet{kaplan_scaling_2020}. We therefore quantify performance along two axes: the next-token prediction loss, the actual objective of base-optimization; and the ability to learn in-context,  measured as the difference in loss calculated over two timepoints within a sequence, as defined by \citet{kaplan_scaling_2020} and \citet{olsson_-context_2022}.

We train Transformers with various architectural configurations on the Pile \citep{gao_pile_2020}, a large compilation of various English text datasets including parts of Wikipedia, arXiv, and code.  We always model the first layer using softmax self-attention in all experiments. This decision is based on insights from our previous experiments, where base-optimization consistently attributed a mesa-objective creation role to this layer. We then compare pure softmax-only Transformers to two types of hybrid models, where the subsequent layers are either linear or mesa. We vary the depth of our models, from 2-layer attention-only to deeper 4-attention-layer models endowed with tokenwise MLPs which are present by default in standard Transformers. By transforming the data nonlinearly, MLP layers allow solving nonlinear regression problems by mesa-gradient descent. Following this reasoning, we further adopt in our hybrid-linear and hybrid-mesa Transformers the deterministic parameter-free projection (DPFP, size denoted by $\nu$) due to \citet{schlag_linear_2021}, a non-learned and simple to compute nonlinear transformation of keys and queries. We found that this significantly improved the performance of non-softmax attention layers. Finally, to represent discrete input symbols as real-valued vectors, we learn a vocabulary of real-valued vectors using the standard GPT-2 tokenizer. We note that all models have an (almost) identical number of parameters.

\begin{figure}[htbp!]
\hspace{-1cm}
    \includegraphics[width=6in]{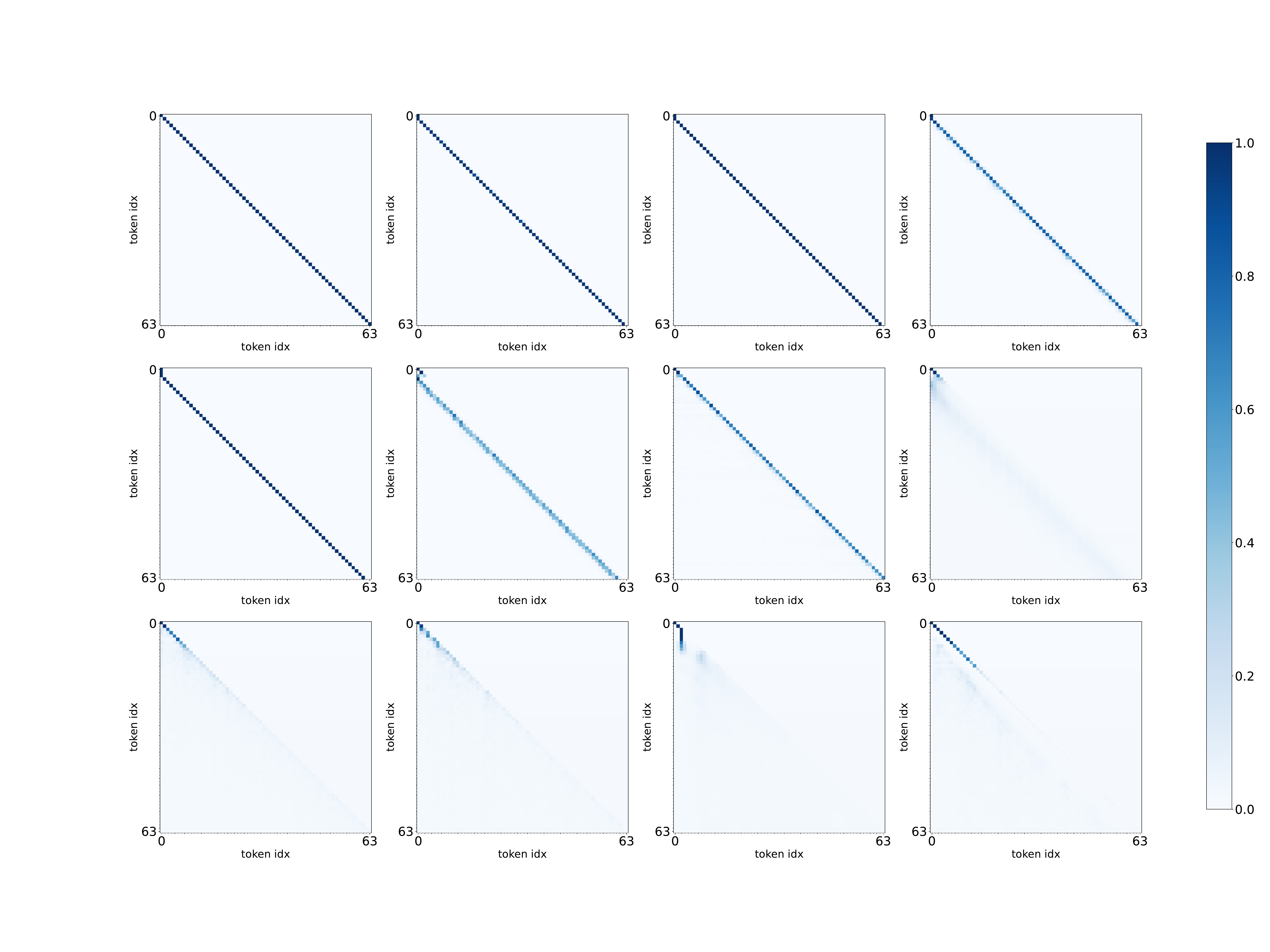}
    \vspace{-1.5cm}
    \caption{\textbf{Softmax attention maps of the 2-layer softmax-only Transformer trained on the Pile.} We average the attention maps of the first softmax-attention layer over a batch of size 256 and observe stable off diagonals with different offsets and widths indicating clean copying behavior based on positional encodings in multiple heads.}
    \label{apx:language-softmax-attn}
    \vspace{0.8cm}
\end{figure}

In line with our synthetic experiments, we observe stable learning across all model types of copying layers, indicated by the constant attention to tokens in direct or close proximity, as shown in Figure \ref{apx:language-softmax-attn}. We therefore reproduce the findings of \citet{olsson_-context_2022}, extending them to models that include other forms of attention. This phenomenon is predicted by the mesa-optimization theory presented here, where copy layers serve the purpose of constructing internal mesa-objective functions.  We note that, in contrast to our previous synthetic linear prediction tasks, the Pile is no longer Markovian of order 1. This is reflected in the more complicated attention maps, indicating more involved copying behavior. Additionally, we run an ablation where we compare to a single-layer control model whose first softmax layer is removed and replaced by a hardcoded one-step key-shift operator. Interestingly, such an operator can be found in previous work \citep{olsson_-context_2022,fu_hungry_2023}. Again, we verify the findings of \cite{olsson_-context_2022} and observe strong in-context learning 
scores, within a single layer, with the mesa-layer performing on-par with softmax, see Figure~\ref{fig:training_trans_lin_key}. As in \cite{schlag_linear_2021}, DPFP features substantially improve performance; we fix $\nu=3$ for the linear as well as the mesa layer for all other language modeling experiments.

\begin{figure*}
%\vspace{-0.3cm}
\centering
\begin{minipage}{.245\textwidth}
  \centering
  \begin{center}
    \includegraphics[width=1.\textwidth]{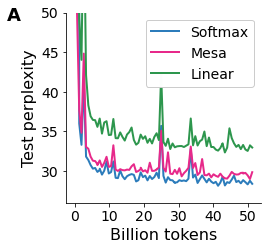}
  \end{center}
\end{minipage}
\begin{minipage}{.245\textwidth}
  \centering
  \begin{center}
    \includegraphics[width=1.\textwidth]{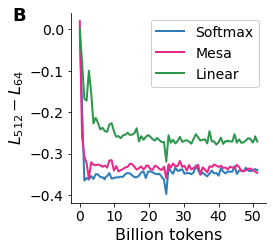}
  \end{center}
\end{minipage}
\begin{minipage}{.245\textwidth}
  \centering
  \begin{center}
    \includegraphics[width=1.\textwidth]{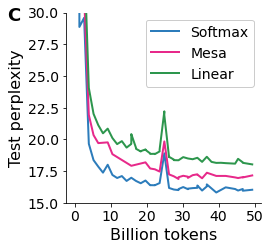}
  \end{center}
\end{minipage}
\begin{minipage}{.245\textwidth}
  \centering
  \begin{center}
    \includegraphics[width=1.\textwidth]{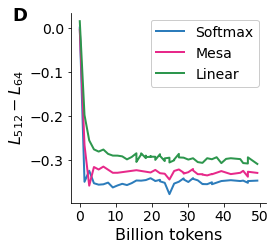}
  \end{center}
\end{minipage}
  \caption{\textbf{Language modeling experiments on the Pile.} We observe improved perplexity and in-context learning scores across all our language modeling experiments when switching from standard linear self-attention to the mesa-layer. As hypothesized, we confirm that in all models various copying heads can be found in the first softmax layer, see Figure \ref{apx:language-softmax-attn} for visualizations of the attention heads. (\textbf{A\&B}) 2-layer Transformers without MLPs and first layers softmax self-attention and second layer either softmax, mesa or linear. (\textbf{C\&D}) 4-layer Transformers with MLPs and first layers softmax self-attention and rest of the layers either all softmax, mesa or linear.}
  \label{fig:training_trans_lin}
  \vspace{-0.3cm}
\end{figure*}

We find that the hybrid-mesa Transformers dominate their hybrid-linear counterparts in terms of performance, across all configurations, essentially matching (for 2-layer models) or coming closer (for 4-layer models with MLPs) to pure-softmax Transformers, 
cf.~Figure~\ref{fig:training_trans_lin}. We leave for future work studying the mesa-layer equipped with forgetting factors, see Appendix \ref{apx:mesa-backward}, which could further improve upon our results here. This is reflected both in terms of perplexity and in-context learning scores. Strictly speaking, these results are not sufficient to make claims on whether mesa-optimization is occurring within standard Transformers. However, the high performance achieved by the hybrid-mesa models, which operate on mesa-optimization principles by design, suggests that mesa-optimization might be happening within conventional Transformers. More reverse-engineering work is needed to add weight to this conjecture.

We provide now additional details about the language modeling experiments. We use standard values found in the literature and the same hyperparameters, which we did not tune, across all experiments. We, if not stated otherwise, use the standard GPT-2 transformer architecture with LayerNorm \citep{ba_layer_2016}, MLPs between self-attention layer and skip-connection after every layer which we train on a standard (autoregressively) masked cross-entropy loss.
We do not use an input embedding layer but an output projection before computing the logits. To train enable stable training of the linear as well as the mesa-layer, we apply the proposed key and query normalization of schlag and simply devide them by their L2 norm. Intriguingly, this stabilizes training drastically also for the mesa-layer after which we did not observe any more instabilities. Note that this is very similar to using additional LayerNorm \citep{ba_layer_2016} on the keys and queries. Except from this normalization, all models are constructed and trained identically. See \ref{tab:hps-language} for an overview of all design decisions and hyperparameters. Also, we refer to the appendix of \cite{schlag_linear_2021} on how to compute the DPFP kernels to non-linearly alter the key and query features,we use $\nu =3$ if not stated otherwise.

\begin{table}[bt]
\begin{center}
\caption{Hyperparameters for language modelling experiments across all Transformer variants i.e. pure softmax, linear-hybrid and mesa-hybrid with/out MLPs.}

\label{tab:hps-language}
\small
\begin{tabular}{@{}ll@{}}
\toprule
Hyperparameter            & Value \\
\toprule
 Dataset & The pile \citep{gao_pile_2020} \\
 Tokenizer & GPT-2 tokenizer - we append a special "EOS" token between every sequence\\
 Context size & 1024 \\
 Vocabulary size & 50257 \\
 Vocabulary dim  & 756 \\
 Optimizer & Adam \citep{kingma_adam_2015} with $\epsilon= 1e^{-8}, \beta_1 =0.9, \beta_2=0.95$ \\
 Weight decay & 0.1  \\
  Batchsize &  256 \\
 Gradient clipping & Global norm of 1. \\
  Positional encodings &  We add standard positional encodings. \\
   Dropout &  We use embedding dropout of 0.1 right after adding positional encodings. \\
  Architecture details &  12 heads, key size 64, token size 756, no input- but output-embedding \\
 Weight init &  $W \sim \mathcal{N}(0, \sigma^2)$ with $\sigma= 0.02$ and bias parameter to zero. We scale all \\ & weight matrices before a skip connection with $\frac{1}{2\sqrt{N}}$ with $N$ the number of layers.\\
Learning rate scheduler &  Linear warm-up starting from $1e^{-6}$ to $3e^{-4}$ in the first 8000 training steps, \\ & cosine annealing to $2e-4$ for the next 300 billion tokens\\
MLP size & Widening factor 4 i.e. hidden dimension $4*756$ with ReLU \\
& non-linearities \citep{hahnloser_digital_2000}\\
Mesa regularization $\lambda$ & We initialize the learnable regularization parameter $\lambda$  for every mesa-head to 1. \\
\bottomrule
\end{tabular}
\end{center}
\end{table}

\section{Software}
The results reported in this paper were produced with open-source software. We used the Python programming language together with the Google JAX \citep{bradbury_jax_2018} framework, and the NumPy \citep{harris_array_2020}, Matplotlib \citep{hunter_matplotlib_2007}, Flax  \citep{hennigan_haiku_2020} and Optax \citep{babuschkin_deepmind_2020} packages.

\end{document}